\documentclass[ijoc,nonblindrev]{informs3}

\OneAndAHalfSpacedXI

\usepackage{hyperref}
\usepackage{mathtools}
\usepackage{subfig}
\usepackage[mathscr]{euscript}
\usepackage{enumitem}
\usepackage{bm}
\usepackage{booktabs}
\usepackage{microtype}

\def\E{\mathbb{E}}
\def\Var{{\rm Var}}
\def\Cov{{\rm Cov}}
\def\pr{\mathbb{P}}
\def\ud{\mathrm{d}}

\def\bx{{\bm x}}
\def\bX{{\bm X}}
\def\bY{{\bm Y}}
\def\bZ{{\bm Z}}

\def\cN{{\mathcal N}}
\def\cX{{\mathcal X}}
\def\cI{{\mathcal I}}

\def\PCS{\mathrm{PCS}}
\def\PCSE{\mathrm{PCS}_{\mathrm{E}}}
\def\APCSE{\widehat{\mathrm{PCS}}_{\mathrm{E}}}
\def\hhet{h_{\mathrm{Het}}}

\def\PCSmin{\mathrm{PCS}_{\mathrm{min}}}
\def\APCSmin{\widehat{\mathrm{PCS}}_{\mathrm{min}}}

\def\cZ{{\mathcal Z}}
\def\cS{{\mathcal{S}}}
\def\cA{{\mathcal A}}
\def\ba{{\bm a}}
\def\bv{{\bm v}}
\def\bz{{\bm z}}
\def\bs{{\bm s}}
\def\bone{{\bm 1}}

\def\cM{{\mathcal M}}
\def\tr{{\text{tr}}}

\makeatletter
\newcommand{\vast}{\bBigg@{4}}
\newcommand{\Vast}{\bBigg@{5}}   
\makeatother

{\theoremstyle{EXkey}\newtheorem{procedure}{Procedure}}

\newsavebox\CBox
\def\textBF#1{\sbox\CBox{#1}\resizebox{\wd\CBox}{\ht\CBox}{\textbf{#1}}}

\usepackage{multirow}

\usepackage{soul}

\usepackage{natbib}
 \bibpunct[, ]{(}{)}{,}{a}{}{,}%
 \def\bibfont{\small}%
 \def\BIBand{and}%

\TheoremsNumberedThrough     
\ECRepeatTheorems

\EquationsNumberedThrough    

\MANUSCRIPTNO{JOC-2018-08-OA-144.R3}

\begin{document}


\RUNAUTHOR{Shen, Hong, and Zhang}

\RUNTITLE{Ranking and Selection with Covariates for Personalized Decision Making.}

\TITLE{Ranking and Selection with Covariates for Personalized Decision Making}

\ARTICLEAUTHORS{%
%
 \AUTHOR{Haihui Shen}
 \AFF{Sino-US Global Logistics Institute, Antai College of Economics and Management, Shanghai Jiao Tong University,\\
 Shanghai, China, \EMAIL{shenhaihui@sjtu.edu.cn}}
 \AUTHOR{L. Jeff Hong\thanks{Corresponding Author}}
 \AFF{School of Management and School of Data Science, Fudan University, Shanghai, China, \EMAIL{hong\_liu@fudan.edu.cn}}
 \AUTHOR{Xiaowei Zhang}
 \AFF{Faculty of Business and Economics, University of Hong Kong, Pok Fu Lam, Hong Kong SAR, \EMAIL{xiaoweiz@hku.hk}}
} 

\ABSTRACT{
We consider a problem of ranking and selection via simulation in the context of personalized decision making, where the best alternative is not universal but varies as a function of some observable covariates.
The goal of ranking and selection with covariates (R\&S-C) is to use simulation samples to obtain a  selection policy that specifies the best alternative with certain statistical guarantee for subsequent individuals upon observing their covariates.
A linear model is proposed to capture the relationship between the mean performance of an alternative and the covariates.
Under the indifference-zone formulation, we develop two-stage procedures for both homoscedastic and heteroscedastic simulation errors, respectively,
and prove their statistical validity in terms of average probability of correct selection.
We also generalize the well-known slippage configuration, and prove that the generalized slippage configuration is the least favorable configuration for our procedures.
Extensive numerical experiments are conducted to investigate the performance of the proposed procedures, the experimental design issue, and the robustness to the linearity assumption.
Finally, we demonstrate  the usefulness of R\&S-C via a case study of selecting the best treatment regimen in the prevention of esophageal cancer.
We find that by leveraging disease-related personal information, R\&S-C can substantially improve patients' expected quality-adjusted life years by providing patient-specific treatment regimen.
}%



\pdfbookmark[1]{Title}{link-title}
\maketitle

%


\section{Introduction} \label{sec-introduction}

Ranking and selection (R\&S) is one of the most studied problems in the area of stochastic simulation.
It aims to select the one with the best mean performance from a set of alternatives through running simulation experiments; see \cite{kim2006} and \cite{ChenChickLeePujowidianto15} for reviews.
In the conventional R\&S setting, the mean performance of an alternative is considered as a constant.
In context of personalized decision making, however, such setting may be too rigid.
For instance, medical studies show that the effectiveness of a cancer chemotherapy treatment depends on the biometric characteristics of a patient such as tumor biomarker and gene expression \citep{yap2009,kim2011short}.
Therefore, for two patients with different characteristics, the best treatment regimen may be different.
Similar examples can also be found in marketing, where research shows that the effect of an online advertisement depends on customer purchasing preference \citep{arora2008},
and in training self-driving cars, where the best driving decision depends on the real-time ambient information collected by all the sensors \citep{katrakazas2015real}.
In all the above examples, it appears more reasonable to consider the mean performance of an alternative as a function of the \textit{covariates}, which include all of the additional contextual information,
such as the biometric characteristics in the cancer treatment example, the purchasing preference in the marketing example, and the ambient information in the self-driving car example.

One approach to solve the problem is to run conventional R\&S procedures once the covariates are observed.
However, this approach may be impractical in many situations for two reasons.
First, the decision maker may not have the access or the time to run the simulation model.
In the cancer treatment example, the simulation model often involves a complex Markov chain, and
one needs to simulate its steady state in order to estimate the treatment effects. It is well known that steady state simulation is computationally expensive as it may take a long time for the Markov chain to reach the steady state.
(The brute force simulation that is based on a simplified Markov chain model and a grid-based interpolation to compute the policy for assigning personalized treatment regimens took about 8 days on our desktop computer to finish; see Section \ref{sec-case} for details.)
In addition, the doctor may not have the access to the simulation model that needs to be run on sophisticated computer systems.
In the marketing example, the online retailer has to display the advertisement once the customer is logged in and thus has no time to run the simulation experiments.
In the self-driving car example, the time is more precious and the decisions have to be made in real time.
A second reason is about efficiency. Personalized decisions typically need to be made repeatedly for different people upon observing their covariates.
Then, running conventional R\&S procedures for each person is conceivably less efficient than developing a selection policy, which maps the covariates to the identity of the best alternative, and using it repeatedly for different people.

We note that we are interested in the kind of situations where the simulation model is expensive to run.
Unless the covariates are of a low dimension, it would be computationally prohibitive to discretize the domain space of the covariates into a grid and then simulate the alternatives at each grid point in a brute force manner.
Instead, an intelligent design is needed to allocate computational budget both to the alternatives and over the domain space of the covariates.
This is also a critical issue that differentiates the R\&S-C problem from the conventional R\&S problem,
which is only concerned with allocating computational budget to the alternatives.

In this paper we consider a new R\&S setting where the mean performances of all alternatives are functions of the covariates and, therefore, the identity of the best alternative is also a function of the covariates.
One may run simulation experiments to learn the mean performance functions of all alternatives and to use these learned functions to select the best alternative upon observing the covariates.
We call this problem \textit{ranking and selection with covariates} (R\&S-C).
Notice that, under this setting, the time-consuming component is the learning of the mean performance functions of all alternatives, and it requires a significant amount of simulation effort.
However, this can be done off-line.
Once the mean performance functions are learned, only these learned functions (which form the selection policy) need to be deployed.
Then, the selection of the best upon observing the covariates is basically computing the function values of all alternatives at the values of the covariates and it can be done on-line in real time, with negligible computation.
Notice that such ``off-line learning on-line application" approach allows the learned functions to be deployed to many users (e.g., doctors and self-driving cars) and used repeatedly with no additional cost.

\subsection{Main Contributions}

To tackle the R\&S-C problem, we first provide a general frequentist formulation.
We generalize important frequentist R\&S concepts, such as the indifference zone and the probability of correct selection (PCS), to the R\&S-C setting, and define the corresponding finite-sample statistical guarantee.
We also show that the R\&S-C formulation in general gives a better outcome than the R\&S formulation if one chooses to average the effects of the covariates.

Second, we consider a specific situation of the R\&S-C problem, where the mean performances of all alternatives are linear in the covariates
(
or linear in certain basis functions of the covariates
)
with unknown coefficients that may be estimated through linear regression,
and show that Stein's lemma \citep{stein1945}, which is a major cornerstone of the conventional frequentist R\&S, may be extended to linear regression contexts.
Despite its simplicity, linear models have distinct advantages in terms of their interpretability and robustness to model misspecification, and often show good performance in prediction \citep{james2013}.

Third, we propose two-stage procedures to solve R\&S-C problems with linear performance functions.
These procedures may be viewed as the extensions of the famous Rinott's procedure \citep{rinott1978} in the conventional frequentist R\&S, and they can handle both homoscedastic and heteroscedastic errors, respectively, in the linear models.
Based on the extended Stein's lemma that we develop, we prove that these procedures deliver the desired finite-sample statistical guarantee.
We also conduct numerical studies to assess the performances of the procedures and discuss their robustness to the linearity assumption and to the experimental design.

Lastly, we consider the personalized prevention regimen for esophageal cancer,
where the effectiveness of the prevention regimens are evaluated using a Markov simulation model developed and calibrated by domain experts in cancer research.
We compare the R\&S-C formulation with the conventional R\&S formulation,
and show that the R\&S-C formulation can significantly improve the expected quality adjusted life years of the patients who are diagnosed with Barrett's esophagus, a mild precursor to esophageal cancer.

\subsection{Related Literature}

R\&S has been studied extensively in the statistics and stochastic simulation literature.
In general, there are two streams  of procedures: frequentist and Bayesian.
Frequentist procedures typically aim to deliver the PCS under the indifference-zone formulation.
There are two-stage procedures \citep{rinott1978}, sequential procedures \citep{kim2001, hong2006}, and procedures designed to handle a large number of alternatives in parallel computing environments \citep{luo2015, ni2017efficient}.
These procedures are typically conservative and require more samples than necessary for average cases.
Bayesian procedures, on the other hand, often aim to allocate a finite computational budget to different alternatives either to maximize the posterior PCS or to minimize the expected opportunity cost.
There are a variety of approaches to developing Bayesian procedures, including value of information \citep{chick2001OR}, knowledge gradient \citep{frazier2008}, optimal computing budget allocation \citep{ChenChenDaiYucesan97}, and economics of selection procedures \citep{ChickGans09,chick2012}.
Bayesian procedures often require fewer samples than frequentist ones to achieve the same level of PCS.
However, they do not provide a (frequentist) statistical guarantee in general.
\cite{Frazier14} develops a Bayes-inspired procedure that includes many of the Bayesian features while still guaranteeing a frequentist PCS.

The R\&S-C problems have been tackled in the Bayesian framework.
\cite{hu2017sequential} propose to model the performance functions of alternatives as Gaussian random fields, and use the expected improvement criteria to develop Bayesian procedures.
\cite{pearce2017efficient} follow the same framework of \cite{hu2017sequential}, but focus on how to efficiently estimate the expected improvement over a continuous domain.
These Bayesian procedures for R\&S-C aim to adaptively allocate a given sampling budget to the alternatives and over the domain of covariates in an efficient way. However, a statistical guarantee on the performance of their solutions is yet to be proved.
In contrast to their approaches, we take a frequentist perspective in this paper, for the first time, to model and solve the R\&S-C problems
that aims to achieve certain finite-sample statistical guarantee on the performance of the solution.

Our research is also related to the literature on multi-arm bandit (MAB) with covariates.
MAB is an important class of sequential decision making problems in the fields of operations research, statistics and machine learning.
It was first proposed by \cite{robbins1952} and has been studied extensively since then; see, for instance, \cite{bubeck2012} for a comprehensive review of MAB.
In recent years, MAB with covariates (also known as contextual MAB) has drawn considerable attention as a tool for personalized decision making.
The mean performances in these problems are often modeled as linear functions of the covariates
\citep{auer2002b,rusmevichientong2010}.
In particular, \cite{GoldenshlugerZeevi13} consider a linear model whose coefficients are arm-dependent, which motivates our formulation of R\&S-C.
Nonparametric models have also been considered in the literature of MAB with covariates \citep{perchet2013,slivkins2014}, and this may be a direction of future study for R\&S-C.
However, a critical distinction between MAB with covariates and R\&S-C lies in the way that the covariates  are obtained. The values of the covariates arrive randomly in the former, whereas in the latter they can be chosen by the decision maker. The additional freedom may conceivably allow one to learn the relationship between the mean performance and the covariates more efficiently.

Linear models have also been considered in R\&S problems.
For example, \cite{negoescu2011} adopt a linear model when solving an R\&S problem in the context of drug discovery, where the mean performance of an alternative is a linear combination of attribute contributions.
However, the intention of introducing the linear model in \cite{negoescu2011} is quite different from ours.
Specifically, the linear model in their work forms a linear projection from the space of alternatives to the space of attributes, which dramatically reduce the computational complexity.
Their final goal is still to select the best alternative as a static decision, rather than the kind of decision policy that we seek.
Therefore, their R\&S problem is still in the conventional sense, which is different from the R\&S-C problems considered in this paper.

A preliminary version of this paper \citep{ShenHongZhang17} was published in the Proceedings of the 2017 Winter Simulation Conference.
This paper extends \cite{ShenHongZhang17} significantly by providing all the proofs for the statistical validity and for the least favorable configuration, discussing the experimental design and robustness to linearity assumptions, adding the discussions on non-normal simulation errors, and applying the proposed formulation and procedure to a case study on personalized medicine.

The remaining of the paper is organized as follows.
In \S \ref{sec-formulation} and \S\ref{sec-linear}, we formulate the R\&S-C problem and introduce the linear models.
In \S \ref{sec-hom}, we develop procedures for homoscedastic and heteroscedastic, and non-normal simulation errors .
The least favorable configuration is discussed in \S \ref{sec-LFC}, followed by numerical experiments in \S \ref{sec-numerical} and a robustness study in \S \ref{sec-robust}.
We demonstrate the practical value of R\&S-C in the context of personalized medicine in \ref{sec-case} and conclude in \S \ref{sec-conclusion}. Technical proofs are collected in the e-companion.

\section{Problem Formulation} \label{sec-formulation}

Suppose there are $k$ alternatives whose mean performances, denoted as $\mu_1(\bX),\ldots,\mu_k(\bX)$, are functions of $\bX=(X_1,\ldots,X_d)^\intercal$,
which is the vector of the observable random \textit{covariates} with support $\Theta \subset \mathbb{R}^d$.
Our goal is to develop a policy that selects the alternative with the largest mean performance upon observing the values of the covariates,
i.e., identifying $i^*(\bx) \coloneqq \argmax_{1\leq i\leq k} \left\{ \mu_i(\bX) |\bX=\bx \right\}$ for any $\bx=(x_1,\ldots,x_d)^\intercal \in\Theta$.
In the cancer treatment example considered in \S \ref{sec-introduction}, for instance, the alternatives are the different treatments,
the covariates are the biometric characteristics of a patient, the mean performances are the expected quality adjusted life years of the patient under different treatments,
and the goal is to identify a policy that selects the best treatment for the patient once the biometric characteristics of the patient are observed.

In this paper we suppose that there are simulation models that allow us to estimate $\mu_1(\bX),\ldots,\mu_k(\bX)$ once the values of $\bX$ are given.
The critical issue here is how to design off-line simulation experiments to learn $\mu_1(\bx),\ldots,\mu_k(\bx)$ accurately so that they
may be used to select the best alternative in real time with a predetermined level of precision upon observing the values of $\bX$
(e.g., PCS in a frequentist sense).
We call this problem \textit{ranking and selection with covariates} (R\&S-C) to emphasize that the decision is conditional on the covariates.

\begin{remark}
Throughout this paper, we assume that the value of $\bX$ is observable before making the selection decision.
This assumption is reasonable in many practical situations, including the three examples introduced in \S \ref{sec-introduction}.
Specifically, in the cancer treatment example, patients' characteristics such as tumor biomarkers and gene expressions can be identified through medical tests;
in the marketing example, customer preference can be inferred from the
demographic and behavioral information as well as the purchasing history (if available);
and in the self-driving car example, the ambient information is collected directly by the sensors.
\end{remark}

\subsection{Value of Covariates}

In the conventional R\&S problem, the goal may be viewed as selecting the unconditional best, i.e., to identify $i^\dagger \coloneqq \argmax_{1\leq i\leq k} \mu_i$, where $\mu_i \coloneqq \E[\mu_i(\bX)]$, $i=1,\ldots,k$,
and the expectation is taken with respect to the distribution of $\bX$.
In the cancer treatment example, for instance, the conventional R\&S selects the best treatment for the entire population instead of the best for an individual patient.
Notice that
$\mu_i(\bX)$ is a random variable.
Then, by Jensen's Inequality,
\begin{equation}\label{jensen1}
\E[\mu_{i^*(\bX)}(\bX)]
= \E \left[ \max_{1\leq i\leq k} \mu_i(\bX) \right] \geq \max_{1\leq i\leq k}  \E \left[ \mu_i(\bX) \right]
= \E[\mu_{i^\dagger}(\bX)].
\end{equation}
Therefore, the R\&S-C formulation typically outperforms the conventional R\&S formulation if the covariates are observable before the selection decision is made.
In the cancer treatment example, for instance, Equation (\ref{jensen1}) implies that the personalized-best treatment typically outperforms the population-best treatment.
This point will also be demonstrated in the cancer prevention example considered in \S \ref{sec-case}.

\begin{remark}
The distribution of $\bX$ is assumed to be known in this paper.
This is a common assumption in the conventional R\&S, where the distribution of $\bX$ needs to be known to evaluate $\E[\mu_i(\bX)]$ for all $i=1,\ldots,k$.
Here, the distribution can be discrete, continuous, or even a mixture of them.
Moreover, the elements of $\bX$ can be dependent on each other.
In practice, the distribution of $\bX$ is typically estimated through the  input modeling process \citep{law2000simulation}.
\end{remark}

\subsection{Indifference Zone}

The concept of indifference zone (IZ) plays a key role in the conventional frequentist R\&S \citep{Bechhofer54}.
It defines the smallest difference $\delta$ that the decision maker considers worth detecting.
Therefore, alternatives whose mean performances are within $\delta$ to the best are in the IZ and are considered ``indifferent" from the best.
In a frequentist setting, R\&S procedures need to deliver the PCS under any configurations of the means.
Without the IZ, the best and the other alternatives may be arbitrarily close in their means. One would then need infinitely many samples to identify the true best
even with PCS less than one.
In the presence of the IZ, the goal is to select one of the alternatives in the IZ and, therefore,
a finite number of samples are needed to ensure that the selected alternative is in the IZ with PCS less than one.

\begin{figure}
\begin{center}
\includegraphics[width=0.5\textwidth]{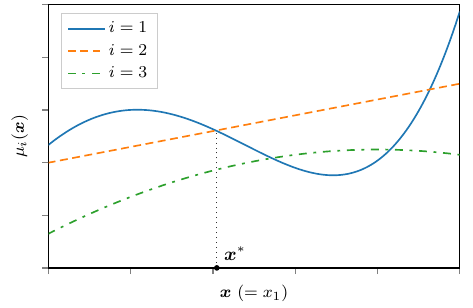}
\caption{Example of Mean Configuration with $d=1$ and $k=3$.} \label{fig-Conf}
\end{center}
\end{figure}

In the setting of R\&S-C, the configurations of the means depend on the values of the covariates.
They may be arbitrarily close if the mean surfaces $\mu_1(\bx),\ldots,\mu_k(\bx)$ intersect at some values of $\bx\in\Theta$ (see, for instance, Figure \ref{fig-Conf}).
Therefore, we also need IZ.
Given an IZ parameter, we define the event of correct selection (CS) given $\bX=\bx$ as
\[
\mbox{CS}(\bx)\coloneqq \left\{ \mu_{i^*(\bx)}(\bx) - \mu_{\widehat{i^*}(\bx)}(\bx)  < \delta \right\},
\]
where $\widehat{i^*}(\bx)$ denotes the selected best, and a CS event implies that the selected best is in the IZ.
Notice that our definition of a CS event is also known as a good selection (GS) event in some of the R\&S literature (see, for instance, \cite{ni2017efficient}),
where a CS event may be defined more restrictively as $\{i^*(\bx) = \widehat{i^*}(\bx)\}$ given that there are no alternatives in the IZ other than the best.
However, this more restrictive definition of CS often does not make sense in the context of R\&S-C.
For instance, when the support of the covariates cover the intersection points of the mean surfaces (see point $\bx^*$ in Figure \ref{fig-Conf}), no matter how small the IZ parameter $\delta$ is,
for certain values of the covariates that are in the neighborhood of the intersection points, there are always alternatives in the IZ other than the best, which makes the more restrictive definition of CS inapplicable.

\begin{remark}
Recently, \cite{fan2016} show that IZ may be unnecessary for the conventional frequentist R\&S if sequential procedures are used.
However, in order for their procedures to stop in finite time (with finite samples), the means of all alternatives have to be different.
In R\&S-C context, however, the mean values of some alternatives are the same at the intersection points (see Figure \ref{fig-Conf}).
Therefore, it is not clear how the procedures of \cite{fan2016} may be applied in the R\&S-C context.
\end{remark}

\subsection{Probability of Correct Selection}

We are now ready to define the PCS, which is the statistical guarantee that frequentist R\&S procedures typically deliver.
Let $\widehat{i^*}(\bx)$ denote the selection policy produced by an R\&S-C procedure.
Notice that the presence of the covariates complicates the definition of PCS, because one has to answer whether the PCS is defined for an individual or the population.
To address the issue, we first define the \textit{conditional} PCS, given $\bX=\bx$, as
\begin{equation} \label{eq-PCSX}
\text{PCS}(\bx) \coloneqq \pr \left\{ \mu_{i^*(\bX)}(\bX) - \mu_{\widehat{i^*}(\bX)}(\bX)  < \delta \big|\bX=\bx \right\},
\end{equation}
where the probability is taken with respect to the distribution of simulated samples that are used to estimate the mean functions $\mu_1(\bx),\ldots,\mu_k(\bx)$ and to derive the selection policy $\widehat{i^*}(\bx)$ for all $\bx\in\Theta$.

Notice that $\text{PCS}(\bx)$ may be viewed as the PCS for an individual whose covariates take the value $\bx$.
However, the covariates are random variables and, therefore, $\PCS(\bX)$ is also a random variable.
To use it as the statistical guarantee for R\&S-C procedures, one way is to consider some summary statistics of $\PCS(\bX)$.
To that end, we define the average PCS, denoted by $\PCSE$, as
\begin{equation} \label{eq-PCSE}
\PCSE \coloneqq \E \left[ \text{PCS}(\bX) \right].
\end{equation}
Notice that
\[\PCSE = \E\left[\pr \left\{ \mu_{i^*(\bX)}(\bX) - \mu_{\widehat{i^*}(\bX)}(\bX)  < \delta \big| \bX \right\}\right] = \pr \left\{ \mu_{i^*(\bX)}(\bX) - \mu_{\widehat{i^*}(\bX)}(\bX)  < \delta \right\}.\]
Therefore, $\PCSE$ is the \textit{unconditional} PCS and it is for the entire population.
If we set $\PCSE\ge 1-\alpha$, we are $(1-\alpha)$ confident that a random individual from the population will select his or her personalized best decision or a decision that is within the IZ.
We want to point out that other summary statistics of $\PCS(\bX)$ may also be used as decision criterion, for instance, one may define it to be certain quantile of the random variable $\PCS(\bX)$ or even $\min_{\bx\in\Theta} \PCS(\bx)$ to be more risk averse.
The $\min_{\bx\in\Theta} \PCS(\bx)$ statistic focuses on the worst $\PCS(\bx)$ over $\Theta$, a much more conservative criterion compared to $\PCSE$, while quantile of $\PCS(\bX)$ is a more flexible statistic, whose conservativeness can be adjusted by the quantile parameter.

Now we summarize our problem.
We want to design a selection procedure that samples each alternative offline to estimate the mean functions $\mu_1(\bx),\ldots,\mu_k(\bx)$ and then produce the selection policy $\widehat{i^*}(\bx)$ for all $\bx\in\Theta$.
This policy will be used to select the best alternative in real time upon observing the values of $\bX$, and it should reach the pre-specified target of $\PCSE$, say, $1-\alpha$.

\begin{remark}
We focus on $\PCSE$ in the main text of this paper.
Selection procedures can be developed and analyzed analogously if  the pre-specified target is $\PCSmin \coloneqq \min_{\bx\in\Theta} \PCS(\bx)\geq 1-\alpha$.
Detailed discussion including numerical experiments is provided in \S\hyperlink{EC.6}{EC.6}.
\end{remark}

\section{Linear Models and the Extended Stein's Lemma} \label{sec-linear}

Notice that the general formulation of R\&S-C problems, presented in \S \ref{sec-formulation}, allows the mean performance functions $\mu_1(\bx),\ldots,\mu_k(\bx)$ to take any forms.
To solve the problems, however, one needs to decide how to estimate these functions.
There are two approaches, parametric approach and nonparametric approach.
Both have pros and cons, and both are widely used in function estimation.
In this paper we take a parametric approach and assume that $\mu_1(\bx),\ldots,\mu_k(\bx)$ are linear functions of the covariates $\bx$ with unknown coefficients that need to be estimated through simulation experiments.
Let $Y_i(\bx)$ denote the random performance of alternative $i$ at the covariates $\bx$ for all $i=1,\ldots,k$ and $\bx\in\Theta$.
We make the following assumption on the forms of $\mu_1(\bx),\ldots,\mu_k(\bx)$ and distributions of $Y_1(\bx),\ldots,Y_k(\bx)$.

\begin{assumption} \label{ass-model}
For all $i=1,\ldots,k$,
\begin{align*}
\mu_i(\bx) &= \bx^\intercal \BFbeta_i,\\
Y_i(\bx) &= \mu_i(\bx) + \epsilon_i,
\end{align*}
where $\BFbeta_i =( \beta_{i1}, \ldots, \beta_{id})^\intercal \in \mathbb{R}^{d}$ is a vector of unknown coefficients and the simulation error $\epsilon_i$ follows a normal distribution with mean $0$ and variance $\sigma_i^2<\infty$.
In addition, the simulation errors are independent among different alternatives, different covariates, and different replications.
\end{assumption}

The linear model in Assumption \ref{ass-model} is crucial for analyzing the performance in terms of $\PCSE$ of the proposed selection procedures.
In particular, without the linear structure, it would be challenging to obtain a finite-sample performance guarantee,
even with the normality assumption on the simulation errors.
Under a nonparametric model such as kernel regression and tree-based methods, selection procedures as well as the analysis of their statistical performance would be drastically different from what is done in this paper.
We leave the investigation to future research.

Despite the simplicity, linear models usually have the advantages of high interpretability and robustness to model misspecification \citep{james2013}.
We will study in \S \ref{sec-robust} the robustness of the procedures developed in this paper when the linearity assumption does not hold.

\begin{remark} \label{remark:linear}
The linear model in Assumption \ref{ass-model} can be generalized to capture nonlinearity of $\mu_i(\bx)$ in $\bx$ by the use of basis functions.
That is, we may postulate $\mu_i(\bx) = \bm f(\bx)^\intercal \BFbeta_i$, where $\bm f(\bx)$ is a vector of
basis functions (e.g., polynomials or radial basis functions) that one selects carefully.
(Nevertheless, selecting a good set of basis functions is a nontrivial task and it is beyond the scope of this paper.)
Note that if we view $\bm f(\bx)$ as a new set of covariates through a change of variables, Assumption \ref{ass-model} and the analysis in the sequel still hold.
Moreover, we often set $X_1 \equiv 1$ to allow an intercept term in the linear model.
We may also include categorical variables in $\bx$ by the use of dummy variables.
\end{remark}

Notice that Assumption \ref{ass-model} basically requires all $Y_i(\bx)$'s to follow the standard linear regression assumption \citep{james2013} so that the unknown coefficient vectors $\BFbeta_i$'s may be estimated using standard ordinary least squares (OLS) method.
Furthermore, Assumption \ref{ass-model} is a natural extension of the normality assumption commonly used in the R\&S literature.
For instance, both \cite{rinott1978} and \cite{kim2001} assume that $Y_i=\mu_i+\epsilon_i$ for all $i=1,\ldots,k$.
We basically extend the mean $\mu_i$ to a linear function $\mu_i(\bx)=\bx^\intercal \BFbeta_i$ to add the effect of the covariates.
Moreover, we will see later in this section, the OLS estimators of the unknown parameters $\BFbeta_i$'s under Assumption \ref{ass-model} resemble the sample-mean estimators of the unknown means under the normality assumption.
This resemblance gives us great convenience to develop statistically valid R\&S-C procedures.

\subsection{Fixed Design}

Based on Assumption \ref{ass-model}, a critical issue in solving an R\&S-C problem is to obtain estimates of $\BFbeta_1,\ldots,\BFbeta_k$ that are accurate enough.
Therefore, we need to decide where to run simulation experiments (i.e., the \textit{design points}) and how many observations to run (i.e., the \textit{sample sizes}).
Here we want to emphasize again that estimation of $\BFbeta_i$'s are conducted offline based on simulation experiments at the chosen design points, instead of real experiments at randomly observed values of the covariates.
Hence, we are free in choosing the number of design points, their locations, and the number of samples to be taken at each design point.
This naturally becomes an experimental design problem, which could be formulated as an optimization problem with an objective of  minimizing certain metric of the error in estimating the policy $i^*(\cdot)=\argmax_{1\leq i\leq k} \mu_i(\cdot)$.
However, this problem is much more challenging to solve than the experimental design problem for linear regression \citep{silvey1980optimal}, primarily because the argmax operation is nonlinear in the unknown surfaces.
It is beyond the scope of this paper to find both the optimal design and the optimal sample size that jointly provide a guarantee on the PCS of the estimated policy $\hat i(\cdot)$.

In this paper, we choose to use a fixed set of design points to estimate $\BFbeta_i$ for all $i=1,\ldots,k$.
In particular, we select a set of $m$ design points, denoted as $\bx_1, \ldots, \bx_m \in \Theta$, with $m\ge d$, and conduct simulation experiments only at these design points for all alternatives.
Notice that the use of fixed design points eliminates the randomness in choosing design points.
It simplifies the analysis and makes statistically valid R\&S-C procedures significantly easier to develop.
When adopting a fixed design, the placement of the design points is an important issue.
We will discuss it in \S \ref{sec-robust}, under the premise that the number of design points is given.
As for now, we simply consider the situation where $m$ design points are chosen and they satisfy that \textit{$\cX^\intercal \cX$ is a nonsingular matrix}, where $\cX = (\bx_1, \ldots, \bx_m)^\intercal \in \mathbb{R}^{m\times d}$.
Notice that the nonsingularity of $\cX^\intercal \cX$ is a standard condition in linear regression \citep{james2013}.
It ensures that all $\BFbeta_i$'s may be estimated properly.

\subsection{Extended Stein's Lemma}

Conventional R\&S procedures often have a first stage to estimate the means and variances of all alternatives, and use them to determine the remaining sample sizes (or sampling policy).
For instance, the two-stage procedures of \cite{dudewicz1975} and \cite{rinott1978} and the sequential procedures of \cite{kim2001} and \cite{hong2006} use the sample variances,
and the OCBA procedure of \cite{ChenChenDaiYucesan97} use both the sample means and variances.
However, this may create a statistical issue because the overall sample size of an alternative now depends on its first-stage samples.
Then, \textit{what is the distribution of the overall sample mean?}

Stein's lemma \citep{stein1945} critically answers this question.
The lemma shows that, if $Y_1,Y_2,\ldots$, are independent and identically distributed (i.i.d.) normal random variables and $N$ depends on the first-stage samples \textit{only through the sample variance},
then the overall sample mean $(Y_1+\cdots+Y_N)/N$, conditionally on the first-stage sample variance, still has a normal distribution.
Consequently, this lemma became a cornerstone of the conventional frequentist R\&S procedures in proving finite-sample statistical guarantees with unknown variances;
see \cite{dudewicz1975} and \cite{rinott1978} for early use of this lemma in designing two-stage R\&S procedures, and Theorem 2 of \cite{kim2006} for a rephrased version of the lemma.

In R\&S-C we also face the problem of unknown variances, i.e., $\sigma_1^2,\ldots,\sigma_k^2$ are unknown in the linear models.
Moreover, we have to deal with the OLS estimators $\widehat\BFbeta_i$'s instead of only the sample means as in the conventional R\&S.
Suppose that we have $m\ge d$ design points with the design matrix $\cX$ and each sample includes an observation from every design point.
Then, we have the following extended Stein's lemma.
We defer its proof to \S\hyperlink{EC.1}{EC.1}, where a more general version is stated and proved,
but remark here that the assumption of the linear models is crucial.

\begin{lemma}[Extended Stein's Lemma] \label{lem-stein}
Let $\bY = \cX \BFbeta + \bm \epsilon$, where $\BFbeta \in \mathbb{R}^d$, $\cX \in \mathbb{R}^{m\times d}$, and $\bm \epsilon \sim \cN(\bm 0, \sigma^2 \cI)$ with $\bm 0$ denoting the zero vector in $\mathbb{R}^m$ and $\cI$ the identity matrix in $\mathbb{R}^{m\times m}$.
Assume that $\cX^\intercal \cX$ is nonsingular.
Let $T$ be a random variable independent  of $\sum_{\ell=1}^n \bY_\ell$ and of $\{\bY_\ell:\ell \geq n+1\}$, where $\bY_1, \bY_2, \ldots$, are independent samples of $\bY$.
Suppose that $N\geq n$ is an integer-valued function of $T$ and no other random variables.
Let $\widehat{\BFbeta} =N^{-1} (\cX^\intercal \cX)^{-1} \cX^\intercal \sum_{\ell=1}^{N} \bY_{\ell}$.
Then, for any $\bx \in \mathbb{R}^d$,
\begin{enumerate}[label=(\roman*)]
\item
$\bx^\intercal \widehat{\BFbeta} \big| T\; \sim\;  \cN \left(\bx^\intercal \BFbeta, \frac{\sigma^2}{N}\bx^\intercal (\cX^\intercal \cX)^{-1} \bx \right)$;
\item
\vspace{3pt}
$ \displaystyle \frac{\sqrt{N} (\bx^\intercal \widehat{\BFbeta} - \bx^\intercal \BFbeta)}{\sigma \sqrt{\bx^\intercal (\cX^\intercal \cX)^{-1} \bx}}$  is independent of $T$ and has the standard normal distribution.
\end{enumerate}
\end{lemma}

\vspace{6pt}
\begin{remark}
If we set $m=d=1$ and $\cX=1$, $\bY$ becomes a scalar and it follows $\cN(\beta_1,\sigma^2)$.
Then, Lemma \ref{lem-stein} becomes Stein's lemma \citep{stein1945}.
In this sense, Lemma \ref{lem-stein} is an extension of Stein's lemma to the linear regression context.
\end{remark}

\begin{remark} \label{remark:T}
In Lemma \ref{lem-stein}, if we let $T$ denote the OLS estimator of the variance $\sigma^2$ computed using samples $\bY_1,\ldots,\bY_n$.
Then, by \citet[Theorem 7.6b]{RencherSchaalje08}, $(nm-d)T/\sigma^2$ follows a chi-square distribution with $(nm-d)$ degrees of freedom and it is independent of $\sum_{\ell=1}^n \bY_\ell$ and of $\{\bY_\ell:\ell \geq n+1\}$.
Therefore, similar to the conventional frequentist R\&S, we may let the sample sizes of all alternatives depend on their first-stage OLS variance estimators and still keep the desired statistical properties.
\end{remark}

\section{Two-Stage Procedures} \label{sec-hom}

For conventional frequentist R\&S procedures, there are two-stage and sequential procedures.
Even though both types of procedures are designed based on the least favorable configuration of means, sequential procedures, such as those of \cite{kim2001} and \cite{hong2006},
take advantage of the information on means and allow the procedures to terminate earlier if the differences between the best and rest of the alternatives are significantly larger than the IZ parameter.
On the other hand, two-stage procedures, such as those of \cite{dudewicz1975} and \cite{rinott1978}, do not take the mean information into consideration and are thus often more conservative.

In R\&S-C problems, however, the configurations of the means depend on the realizations of the covariates.
For some realizations of the covariates, the differences may be larger than the IZ parameter;
while for other realizations, the differences may be much smaller, even close to zero (see, for instance, Figure \ref{fig-Conf}).
The procedures that we intend to design need to deliver a selection policy $\widehat {i^*}(\bx)$ for all $\bx\in\Theta$ before the covariates are realized,
and the policy may be used repeatedly for many realizations of the covariates.
Therefore, it is not clear whether sequential procedures may still be advantageous in the R\&S-C context.
In this paper we focus on designing two-stage procedures that deliver the desired finite-sample statistical guarantee.

\subsection{The Procedure and Statistical Validity}

We develop a two-stage procedure, called Procedure TS, for R\&S-C problems under Assumption \ref{ass-model}.
In the first stage, the procedure takes a small number of samples from all design points to estimate the total sample size required to deliver the desired statistical guarantee;
and in the second stage, it takes the additional samples and produces a selection policy based on all samples.
The structure of the procedure resembles many of the conventional two-stage R\&S procedures, including those of \cite{dudewicz1975} and \cite{rinott1978}.

\begin{procedure}[Procedure TS]

\begin{description}
\item[]
\item[Setup:]
Specify the target $\PCSE$ $1-\alpha$, the IZ parameter $\delta > 0$, the first-stage sample size $n_0\geq 2$, the number of design points $m\geq d$, and the design matrix $\mathcal{X}$ with a nonsingular $\cX^\intercal \cX$.
Let $h$ satisfy the following equation
\begin{equation} \label{eq-geth}
\E \left\{ \int_0^\infty \left[\int_0^\infty \Phi \left( \frac{h}{\sqrt{(n_0m-d) (t^{-1}+s^{-1})\bX^\intercal (\cX^\intercal \cX)^{-1} \bX}} \right) \eta(s)\ud s \right]^{k-1} \eta(t) \ud t \right\} = 1-\alpha,
\end{equation}
where $\Phi(\cdot)$ is the cumulative distribution function (cdf) of the standard normal distribution, $\eta(\cdot)$ is the probability density function (pdf) of the chi-square distribution with $(n_0m-d)$ degrees of freedom,
and the expectation is taken with respect to the distribution of $\bX$.

\item[First-stage Sampling:]
Take $n_0$ independent samples from each alternative $i$ at each design point $\bx_j$ through simulation, and denote them by $\bY_{i\ell}=(Y_{i\ell}(\bx_1),\ldots,Y_{i\ell}(\bx_m))^\intercal$,
$i=1,\ldots,k$, $\ell = 1,\ldots,n_0$.
For each $i=1,\ldots,k$, let
\begin{eqnarray*}
\widehat{\BFbeta}_{i0} &=& \frac{1}{n_0} \left(\cX^\intercal \cX\right)^{-1} \cX^\intercal \sum_{\ell=1}^{n_0} \bY_{i\ell},\\
S_i^2 &=& \frac{1}{n_0 m-d} \sum_{\ell=1}^{n_0}\left (\bY_{i\ell}-\cX \widehat{\BFbeta}_{i0} \right)^\intercal \left(\bY_{i\ell}-\cX \widehat{\BFbeta}_{i0} \right).
\end{eqnarray*}

\item[Second-stage Sampling:]
Compute the total sample size $N_i = \max \left\{\lceil h^2 S_i^2/\delta^2\rceil, n_0 \right\}$ for each $i$, where $\lceil a \rceil$ denotes the smallest integer no less than $a$.
Take $N_i-n_0$ additional independent samples from alternative $i$ at all design points through simulation, $\bY_{i,n_0+1}, \ldots, \bY_{iN_i}$, $i=1,\ldots,k$.
For each alternative $i$, let
\[\widehat{\BFbeta}_i = \frac{1}{N_i} (\cX^\intercal \cX)^{-1} \cX^\intercal \sum_{\ell=1}^{N_i} \bY_{i\ell}.\]

\item[Selection:]
Return $\widehat{i^*}(\bx) = \argmax_{1\leq i\leq k}\left \{ \bx^\intercal \widehat{\BFbeta}_i \right\}$ as the selection policy.

\end{description}
\end{procedure}

\vspace{6pt}
\begin{remark}
Similar to typical two-stage R\&S procedures, the first-stage sample size $n_0$ is chosen heuristically here.
If $n_0$ is too small, then $h$ calculated from \eqref{eq-geth} tends to be large, leading to excessive second-stage samples to compensate for the inaccurate variance estimator $S_i^2$ in the first stage. Taking $n_0 \geq 10$ is a common recommendation \citep{kim2006}.
\end{remark}

\begin{remark}
The constant $h$, defined in \eqref{eq-geth}, is computed numerically.
In our numerical experiments, the integrations and expectations are computed by the MATLAB built-in numerical integration function \texttt{integral}, and $h$ is solved by the MATLAB built-in root finding function \texttt{fzero}.
However, the numerical integration may suffer from the curse of dimensionality if the dimension of $\bX$ is large.
In such situations one may use the Monte Carlo method to approximate the expectation or apply the stochastic approximation method \citep{robbins1951stochastic} to find the root $h$.
See more discussion in \S\hyperlink{EC.2}{EC.2}.
\end{remark}

The following theorem states that Procedure TS is statistically valid under Assumption \ref{ass-model}.
We include the proof in \S\hyperlink{EC.3}{EC.3}, but remark here that the proof relies critically on the extended Stein's lemma (Lemma \ref{lem-stein}).

\begin{theorem} \label{thm-hom}
Suppose that Procedure TS is used to solve the R\&S-C problem and Assumption \ref{ass-model} is satisfied. Then,  $\PCSE \geq 1-\alpha$.
\end{theorem}

\subsection{Handling Heteroscedastic Errors} \label{sec-het}

In Assumption \ref{ass-model} we assume that the variance of simulated samples of an alternative does not change with respect to the values of the covariates.
This implies that the linear models all have homoscedastic simulation errors.
However, this assumption may not always hold.
In many practical situations, such as queueing and financial applications, simulation errors are often heteroscedastic.
In this subsection we present a two-stage R\&S-C procedure to take care of the heteroscedasticity in the linear models.

We first extend Assumption \ref{ass-model} to the following to allow heteroscedastic errors.
\begin{assumption} \label{ass-model2}
For all $i=1,\ldots,k$,
\begin{align*}
\mu_i(\bx) &= \bx^\intercal \BFbeta_i,\\
Y_i(\bx) &= \mu_i(\bx) + \epsilon_i(\bx),
\end{align*}
where $\BFbeta_i =( \beta_{i1}, \ldots, \beta_{id})^\intercal \in \mathbb{R}^{d}$ is a vector of unknown coefficients and the simulation error $\epsilon_i(\bx)$ follows a normal distribution with mean $0$ and variance $\sigma_i^2(\bx)<\infty$.
In addition, the simulation errors are independent among different alternatives, different covariates, and different replications.
\end{assumption}

When linear models have heteroscedastic errors, OLS estimators of $\BFbeta_i$'s are still consistent estimators.
However, subtle controls are needed to deliver the required $\PCSE$.
Because our experiments are controlled simulation experiments, we may run multiple simulation runs at each design point to calculate the sample variance at the design point.
We then use these sample variances to determine the (different) total sample sizes at different design points.
We call this new two-stage procedure Procedure TS$^+$.
One distinct feature of Procedure TS$^+$ is that it allows different design points to have different total sample sizes to handle heteroscedastic errors;
while Procedure TS always assigns the same total sample size to all design points.
The following is the procedure.
For simplicity, we use $\chi_\nu^2$ to denote the chi-square distribution with $\nu$ degrees of freedom.

\begin{procedure}[Procedure TS$^+$]
\item[]
\begin{description}
\item[Setup:]
Specify the target $\PCSE$ $1-\alpha$, the IZ parameter $\delta > 0$, the first-stage sample size $n_0\geq 2$, the number of design points $m\geq d$, and the design matrix $\mathcal{X}$ with a nonsingular $\cX^\intercal \cX$. Let $\hhet$ satisfy the following equation
\begin{equation} \label{eq-geth2}
\E \left\{ \int_0^\infty \left[ \int_0^\infty \Phi \left(  \frac{\hhet}{\sqrt{ (n_0-1) (t^{-1}+s^{-1})\bX^\intercal (\cX^\intercal \cX)^{-1} \bX }} \right) \gamma_{(1)}(s) \ud s \right]^{k-1} \gamma_{(1)}(t) \ud t \right\} = 1-\alpha,
\end{equation}
where $\gamma_{(1)}(\cdot)$ is the pdf of the smallest order statistic of $m$ i.i.d. $\chi_{n_0-1}^2$ random variables,  i.e.,
\begin{equation*} \label{eq-minf}
\gamma_{(1)}(t) = m \gamma(t) (1-\Gamma(t))^{m-1},
\end{equation*}
with $\gamma(\cdot)$ and $\Gamma(\cdot)$ denoting the pdf and cdf of the $\chi_{n_0-1}^2$ distribution, respectively, and the expectation is taken with respect to the distribution of $\bX$.

\item[First-stage Sampling:]
Take $n_0$ independent samples of each alternative $i$ from each design point $\bx_j$ through simulation, and denote them by $Y_{i\ell}(\bx_1),\ldots,Y_{i\ell}(\bx_m)$, $i=1,\ldots,k$, $\ell = 1,\ldots,n_0$.
For each $i$ and $j$, let
\[
\overline{Y}_{ij} =\frac{1}{n_0} \sum_{\ell=1}^{n_0} Y_{i\ell}(\bx_j)\quad \text{ and } \quad
S_{ij}^2 = \frac{1}{n_0-1} \sum_{\ell=1}^{n_0} \Big(Y_{i\ell}(\bx_j) - \overline{Y}_{ij} \Big)^2.
\]

\item[Second-stage Sampling:]
Compute the total sample size $N_{ij} = \max \left\{\lceil \hhet^2 S_{ij}^2/\delta^2\rceil, n_0 \right\}$ for each $i$ and $j$.
Take $N_{ij}-n_0$ additional independent samples from alternative $i$ at design point $\bx_j$ through simulation, $Y_{i,n_0+1}(\bx_j), \ldots, Y_{iN_{ij}}(\bx_j)$, $j=1,\ldots,m$, $i=1,\ldots,k$.
For each alternative $i$, let
\[\widehat{\BFbeta}_i = (\cX^\intercal \cX)^{-1} \cX^\intercal \widehat{\bY}_i,\]
where $\widehat{\bY}_i = (\widehat{Y}_{i1}, \ldots, \widehat{Y}_{im})^\intercal$ and
\[ \widehat{Y}_{ij} =\frac{1}{N_{ij}} \sum_{\ell=1}^{N_{ij}} Y_{i\ell}(\bx_j).\]

\item[Selection:]
Return $\widehat{i^*}(\bx) = \argmax_{1\leq i\leq k} \left\{  \bx^\intercal \widehat{\BFbeta}_i \right\}$ as the selection policy.

\end{description}
\end{procedure}

\begin{remark}
The smallest order statistics in Equation \eqref{eq-geth2} are introduced to make the computation of $\hhet$ feasible.
Without it, the equation for computing the constant $\hhet$ would involve $(2m)$-dimensional numerical integration, which becomes prohibitively difficult to solve for $m\geq 3$.
The price of using the smallest order statistic is that $\hhet$ is (slightly) larger then necessary, which introduces some  conservativeness in the procedure.
See Remark \ref{remark:h} in \S\hyperlink{EC.4}{EC.4} for more details.
\end{remark}

The following theorem states that Procedure TS$^+$ is statistically valid under Assumption \ref{ass-model2}.
Its proof, which is included in \S\hyperlink{EC.4}{EC.4}, is similar to that of Theorem \ref{thm-hom}, but technically more involved.
We remark here that the proof relies critically on a more generalized extension of Stein's lemma \citep{stein1945},
which is stated and proved as Lemma \ref{lem-stein2} in \S\hyperlink{EC.1}{EC.1}.

\begin{theorem} \label{thm-het}
Suppose that Procedure TS$^+$ is used to solve the R\&S-C problem and Assumption \ref{ass-model2} is satisfied. Then,  $\PCSE \geq 1-\alpha$.\end{theorem}

\begin{remark}
Let $N_{\text{TS}}$ and $N_{\text{TS}^+}$ denote the expected total sample sizes of Procedure TS of Procedure TS$^+$, respectively.
It can be shown that
$N_{\text{TS}}=\mathcal{O}(k^{1+\frac{2}{n_0m-d}})$ and
$N_{\text{TS}^+}= \mathcal{O}(k^{1+\frac{2}{n_0-1}})$
as $k\to\infty$;
meanwhile, $N_{\text{TS}}=\mathcal{O}(\alpha^{-\frac{2}{n_0m-d}})$ and $N_{\text{TS}^+}= \mathcal{O}(\alpha^{-\frac{2}{n_0-1}})$ as  $\alpha\to 0$.
The proofs are provided in  \S\hyperlink{EC.7}{EC.7}.
It turns out that two classical selection procedures for conventional R\&S problems -- the two-stage procedure in \cite{rinott1978} and the sequential procedure in \cite{kim2001} -- have an upper bound on their sample sizes that are similar to those of Procedure TS and Procedure TS$^+$ \cite[Lemma 4]{zhong2020}.
In this sense, Procedure TS and Procedure TS$^+$ can solve the R\&S-C problem, an extension of the conventional R\&S problem, without substantially increasing the sample size complexity.
Nevertheless, it must be stressed that this property relies heavily on the linear model assumption.
\end{remark}

\subsection{Comparison between Procedure TS and Procedure TS$^+$}
Clearly,  the  assumption of homoscedasticity yields more analytical and computational tractability than the assumption of heteroscedasticity.
However, if Procedure TS is used in the presence of heteroscedastic errors, it may fail to deliver the desired $\PCSE$ guarantee.
An intuitive explanation is that using a single variance estimate for all the design points may underestimate the variance at some design points, leading to insufficient sampling effort at those design points.

On the other hand, Procedure TS$^+$ may behave in an overly conservative manner when used in the case of homoscedastic errors.
This is because Procedure TS$^+$ requires estimation of the variances at all design points, which amounts to estimating the common variance repeatedly in the homoscedasticity setting, resulting in excessive sampling effort.
To be more specific, let us consider the estimators of the common variance $\sigma_i^2$ in Procedure TS and Procedure TS$^+$, which are $S_i^2$ and $S_{ij}^2$, respectively.
It is easy to see that they are both unbiased estimators of $\sigma_i^2$,
with the former having variance $2\sigma_i^4/(n_0m-d)$ and the latter $2\sigma_i^4/(n_0-1)$.
Since $n_0m-d \geq n_0d-d \geq n_0-1$, $S_{ij}^2$ has a larger variance.
This is not surprising as $S_{ij}^2$ just uses $n_0$ samples to estimate the variance whereas $S_i^2$ uses $n_0m$ samples.
Hence, Procedure TS$^+$ will require more second-stage samples to compensate for the less accurate variance estimator.
Furthermore, the use of the order statistic in Procedure TS$^+$ further loosens the lower bound of the $\PCSE$ and results in more excessive sample sizes.
These behaviors are revealed clearly through the numerical experiments in \S\ref{sec-numerical}.

The above discussion provides us a rule of thumb for choosing the procedures in practice.
Procedure TS may be preferred if either the problem has approximately homoscedastic errors, or the decision maker can tolerate some underachievement relative to the desired $\PCSE$.
On the other hand, Procedure TS$^+$ may be a better choice if the errors are notably heteroscedastic or if the decision maker is stringent on delivering the $\PCSE$ guarantee.

\subsection{Handling Non-Normal Errors}

We now discuss non-normal simulation errors.
We first consider the case of homoscedasticity and relax Assumption \ref{ass-model} to the following.

\begin{assumption}\label{ass-model3}
For all $i=1,\ldots,k$,
$\mu_i(\bx) = \bx^\intercal \BFbeta_i$ and
$Y_i(\bx) = \mu_i(\bx) + \epsilon_i$,
where $\BFbeta_i =( \beta_{i1}, \ldots, \beta_{id})^\intercal \in \mathbb{R}^{d}$ is a vector of unknown coefficients and the simulation error $\epsilon_i$ has mean $0$ and variance $\sigma_i^2<\infty$.
In addition, the simulation errors are independent among different alternatives, different covariates, and different replications.
\end{assumption}

In the absence of normality, the extended Stein's lemma (Lemma \ref{lem-stein}) does not hold.
As a consequence, Procedure TS does not provide finite-sample statistical validity in terms of $\PCSE$ under Assumption \ref{ass-model3}.
Instead, we establish its statistical validity in an asymptotic sense. In particular, we adopt the ``small $\delta$'' regime, that is, $\delta\to 0$.
This asymptotic regime is often used in R\&S literature; see, for example, \cite{kim2006} and \cite{luo2015}.

Note that as $\delta\to0$,  the smallest difference that the decision-maker deems worth detecting vanishes, and the R\&S-C problem becomes increasingly difficult, requiring infinitely many samples eventually.
Meanwhile, the central limit theorem suggests that the estimates of the linear coefficients are asymptotically normal, which would lead to the asymptotic validity of Procedure TS. The proof of the following theorem is given in \S\hyperlink{EC.5}{EC.5}.

\begin{theorem} \label{thm-hom-asy}
Suppose that Procedure TS is used to solve the R\&S-C problem and Assumption \ref{ass-model3} is satisfied. Then,
$\liminf_{\delta \to  0} \PCSE \geq 1-\alpha$.
\end{theorem}

Furthermore, we relax Assumption \ref{ass-model2} likewise and show that Procedure TS$^+$ is statistically valid asymptotically as $\delta \to 0$ as well. The proof is similar to that of Theorem \ref{thm-hom-asy}, so we omit the details.

\begin{assumption} \label{ass-model4}
For all $i=1,\ldots,k$,  $\mu_i(\bx) = \bx^\intercal \BFbeta_i$ and
$Y_i(\bx) = \mu_i(\bx) + \epsilon_i(\bx)$,
where $\BFbeta_i =( \beta_{i1}, \ldots, \beta_{id})^\intercal \in \mathbb{R}^{d}$ is a vector of unknown coefficients and the simulation error $\epsilon_i(\bx)$ has mean $0$ and variance $\sigma_i^2(\bx)<\infty$.
In addition, the simulation errors are independent among different alternatives, different covariates, and different replications.
\end{assumption}

\begin{theorem} \label{thm-het-asy}
Suppose that Procedure TS$^+$ is used to solve the R\&S-C problem and Assumption \ref{ass-model4} is satisfied. Then,
$\liminf_{\delta \to  0} \PCSE \geq 1-\alpha$.
\end{theorem}

\section{Least Favorable Configuration} \label{sec-LFC}

For conventional R\&S problems, the so-called \textit{least favorable configuration} (LFC) is an important concept, because it defines the most difficult configuration of the means of the alternatives for the selection procedures \citep{Bechhofer54}.
Indeed, many selection procedures are designed by analyzing the LFC.
If a selection procedure can meet the target PCS under its LFC, it can certainly meet the same target for all mean configurations.
It is well known that under the IZ formulation, the LFC for R\&S problems is the \textit{slippage configuration} (SC) for many procedures \citep{gupta1982}.
The SC is a configuration where there exists a unique best alternative and all other alternatives have equal means which differ from the best by exactly the IZ parameter.

To better understand our proposed procedures for R\&S-C, it is important to investigate their LFCs,
formally defined as follows.
For given $k$, distribution of $\bX$, and $\sigma_i^2(\bx)$, $i=1,\ldots,k$,
the LFC for a R\&S-C procedure is the value of $\BFbeta=(\BFbeta_i:1\leq i \leq k)$ that minimizes the $\PCSE$ of that procedure.
That is,
$$\text{LFC} \coloneqq \argmax_{\BFbeta=(\BFbeta_i:1\leq i \leq k)} \PCSE(\BFbeta),$$
where $\PCSE(\BFbeta)$ denotes the $\PCSE$ of the procedure under the configuration $\BFbeta$.
Note that this definition of LFC generalizes the same notion for the conventional R\&S problem.
If a selection procedure can meet the target $\PCSE$ under its LFC, it will meet the same target for any other configurations.

\begin{figure}
\begin{center}
\includegraphics[width=0.56\textwidth]{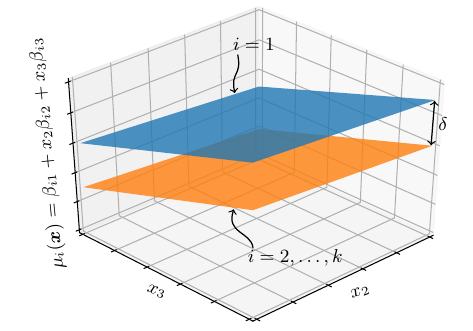}
\caption{Geometrical Illustration of the GSC of Linear Mean Surfaces for $d=3$.} \label{fig-GSC}
\end{center}
\begin{minipage}[t]{1\linewidth}
\vspace{-1.2em}
\SingleSpacedXI
\footnotesize{
\emph{Note.} \textsf{Coordinate $x_1$ is omitted since it is always 1.}
}
\end{minipage}
\end{figure}

We first generalize the SC in conventional R\&S problems to the R\&S-C setting and define the \textit{generalized slippage configuration} (GSC) as follows:
\begin{equation} \label{eq-GSC}
\mu_1(\bx) - \mu_i(\bx) = \delta,\quad \textnormal{for all}\ \bx\in\Theta\ \textnormal{and all}\ i=2,\ldots,k.
\end{equation}
Under the linearity assumption (Assumption \ref{ass-model} or \ref{ass-model2}), the GSC becomes
\begin{equation}\label{eq:GSC_equality}
 \bx^\intercal \BFbeta_1 - \bx^\intercal \BFbeta_i= \delta,  \quad \textnormal{for all}\ \bx\in\Theta\ \textnormal{and all}\ i=2,\ldots,k.
\end{equation}
Hence, under the GSC, the best alternative is the same for all $\bx\in\Theta$, and all other alternatives have equal mean performances.
It is worth mentioning that, the GSC of linear mean surfaces implies the existence of an intercept term (i.e., $X_1 \equiv 1$).
Geometrically, the GSC means that the hyperplanes formed by the mean performances of the inferior alternatives are identical and parallel to the hyperplane of the best alternative,
and the vertical distance between the two hyperplanes (i.e, the difference between the intercepts) is exactly $\delta$; see Figure \ref{fig-GSC} for an illustration for $d=3$.

It turns out that the GSC defined in \eqref{eq:GSC_equality} is the LFC for both Procedures TS and TS$^+$ under the IZ formulation.
We summarize this result in the following theorem.
A slightly more general result is provided and proved in \S\hyperlink{EC.8}{EC.8}.

\begin{theorem} \label{thm-GSC}
The GSC is the LFC for Procedures TS and TS$^+$.
\end{theorem}

\begin{remark}
Theorem \ref{thm-GSC} not only deepens our understanding of our procedures, but also helps us design numerical experiments to serve as a stress test for the proposed procedures.
\end{remark}

\section{Numerical Experiments} \label{sec-numerical}

In this section, we investigate numerically the statistical validity of the two proposed procedures.
We create a number of problem instances to test the procedures.
For each problem instance, we need to specify the number of alternatives $k$, the dimension of the covariates $d$, the design matrix $\cX$, the mean configuration parameters $\BFbeta_i$'s,
the variance configuration $\sigma_i^2(\cdot)$'s, and the distribution of $\bX$.
Instead of specifying the above aspects in a combinatorial fashion, which would result in an excessively large number of problem instances,
we first create a benchmark problem and then investigate the effect of a factor by varying it while keeping others unchanged.
All the numerical studies, including the numerical experiments in  this section, \S \ref{subsec:robust}, and \S \ref{sec-case}, are implemented in MATLAB  on a desktop computer with Windows 10 OS,  3.60 GHz CPU, and 16 GB RAM.
The source code is available at \url{https://shenhaihui.github.io/research}.

The benchmark  problem is formulated as follows.
Let $d=4$ and $k=5$.
Suppose that $\bX = (1,X_2,\ldots,X_d)^\intercal$, and $X_2,\ldots,X_d$ are i.i.d. \textsf{Uniform}$[0,1]$ random variables.
Here the first covariate is always 1, which is used to include the intercept terms for linear models.
We set each except the first entry of a $d$-dimensional design point to be 0 or 0.5, so there are $m=2^{d-1}$ design points in total.
We set the configuration of the means to be the GSC, i.e., $\beta_{11} - \delta = \beta_{i1} = 0, \beta_{1w} = \beta_{iw} = 1,$ for $i=2,\ldots,k$ and $w=2,\ldots,d$,
and set the simulation errors to be homoscedastic, particularly $\sigma_i(\bx)\equiv\sigma_i = 10$ for $i=1,\ldots,k$.

We then create 9 test problems below by varying one factor of the benchmark problem at a time, while keeping other factors the same.
\begin{enumerate}[label=\texttt{(\arabic*)}]
\item
$k=2$.
 \item
$k=8$.
\item
Randomly generated components of $\BFbeta_i$ from \textsf{Uniform}$[0,5]$, $i=1,\ldots,5$.
\item
Increasing variances (IV) configuration: $\sigma_1=5$, $\sigma_2=7.5$, $\sigma_3=10$, $\sigma_4=12.5$, $\sigma_5=15$.
\item
Decreasing variances (DV) configuration: $\sigma_1=15$, $\sigma_2=12.5$, $\sigma_3=10$, $\sigma_4=7.5$, $\sigma_5=5$.
\item
Heteroscedastic simulation errors: $\sigma_i(\bx)=10  \bx^\intercal \BFbeta_i $, $i=1,\ldots,5$.
\item
$d=2$.
\item
$d=6$.
\item
$X_i \sim \cN(0.5,1)$ truncated on $[0,1]$, and $\Cov(X_i, X_j)=0.5$, for $i,j=2,3,4$ and $i\neq j$.
\end{enumerate}
Compared to the benchmark problem, Problems \texttt{(1)} and \texttt{(2)} change the number of alternatives,
Problem \texttt{(3)} changes the configuration of the means so it is no longer the GSC,
Problems \texttt{(4)} and \texttt{(5)} change the configuration of the variances while retaining homoscedasticity,
Problem \texttt{(6)} considers heteroscedasticity, and Problems \texttt{(7)} and \texttt{(8)} change the dimensionality of the covariates,
and Problem \texttt{(9)} changes the distribution of the covariates.

We further create three large-scale problems.
\begin{enumerate}[label=\texttt{(\arabic*)}, resume]
  \item  $k=100$.
  \item $d=50$.
  \item $k=100$, $d=50$.
\end{enumerate}
Note that Problem \texttt{(12)} changes both $k$ and  $d$ relative to the the benchmark problem.
For these three problems, instead of taking $2^{d-1}$ design points as before, we use the Latin hypercube sampling with $m=2d$ design points.

In all the problem instances, we set $\alpha = 0.05$, $\delta = 1$, and $n_0 = 50$.
We conduct $R = 10^4$ macro-replications for each problem-procedure combination.
In each macro-replication $r=1,\ldots,R$, we apply Procedure TS and TS$^+$, respectively, to a problem to obtain a selection policy $\widehat{i^*_r}(\bx)$,
and then apply it to select the best alternative for each $\bx_t$, a realization of $\bX$ that is randomly generated from its distribution, for $t=1,\ldots,T$ with $T=10^5$.
We calculate the achieved $\PCSE$ as
\begin{equation}\label{eq-APCSE}
\APCSE\coloneqq  \frac{1}{R} \sum_{r=1}^R \frac{1}{T} \sum_{t=1}^T \mathbb{I} \left\{ \mu_{i^*(\bx_t)}(\bx_t) - \mu_{\widehat{i^*_r}(\bx_t)}(\bx_t)  < \delta \right\},
\end{equation}
where $\mathbb{I}{\{\cdot\}}$ denotes the indicator function.
We also report the average total sample size used by each procedure for producing the selection policy.

\begin{table}[!b]
\centering
\small
\caption{Results When the Target is $\PCSE \geq 95\%$.} \label{tab-PCSE}
    \begin{tabular}{lcccccccccc}
    \toprule
     & & &  \multicolumn{3}{c}{Procedure TS} & & & \multicolumn{3}{c}{Procedure TS$^+$} \\
     \cmidrule(lr){4-6} \cmidrule(lr){9-11}
     Problem & & &  $h$ & Sample Size&  $\APCSE$  & & & $\hhet$ & Sample Size &  $\APCSE$ \\
    \midrule
    \texttt{(0)} Benchmark & & & 3.423 & 46,865 & 0.9610 & &            & 4.034 & \phantom{1}65,138 & 0.9801 \\

    \texttt{(1)} $k=2$ & &     & 2.363 & \phantom{1}8,947 & 0.9501 & &  & 2.781 & \phantom{1}12,380 & 0.9702 \\
    \texttt{(2)} $k=8$ & &     & 3.822 & 93,542 & 0.9650 & &            & 4.510 & 130,200           & 0.9842 \\
    \texttt{(3)} Non-GSC & &   & 3.423 & 46,865 & 0.9987 & &            & 4.034 & \phantom{1}65,138 & 0.9994 \\
    \texttt{(4)} IV & &        & 3.423 & 52,698 & 0.9618 & &            & 4.034 & \phantom{1}73,265 & 0.9807 \\
    \texttt{(5)} DV & &        & 3.423 & 52,720 & 0.9614 & &            & 4.034 & \phantom{1}73,246 & 0.9806 \\
    \texttt{(6)} Het & &       & 3.423 & 58,626 & \framebox{0.9232}  & & & 4.034 & \phantom{1}81,555 & \textBF{0.9846} \\
    \texttt{(7)} $d=2$ & &     & 4.612 & 21,288 & 0.9593 & &            & 4.924 & \phantom{1}24,266 & 0.9662 \\
    \texttt{(8)} $d=6$ & &     & 2.141 & 73,428 & 0.9656 & &            & 2.710 & 117,626           & 0.9895 \\
    \texttt{(9)} Normal Dist & &     & 3.447 & 47,529 & 0.9626 & &            & 4.063 & \phantom{1}66,061           & 0.9821 \\

    \cmidrule{1-11}

    \texttt{(10)} $k=100$ & &             & 4.346 & \phantom{0}1,133,384 & 0.9758 & &            & 5.117 & \phantom{0}1,570,911           & 0.9918 \\
    \texttt{(11)} $d=50$ & &              & 3.222 & \phantom{00,}508,977 & 0.9583 & &            & 4.312 & \phantom{00,}911,326           & 0.9926 \\
    \texttt{(12)} $k=100$, $d=50$ & &     & 4.886 & 23,400,677 & 0.9765 & &            & 6.702 & 44,024,486           & 0.9991 \\
    \bottomrule
    \end{tabular}
\begin{minipage}[t]{1\linewidth}
\SingleSpacedXI
\vspace{0.6em}
\footnotesize{
\emph{Note.} \textsf{(i) In the presence of heteroscedasticity, the boxed number suggests that Procedure TS fails to deliver the target $\PCSE$, whereas the bold number suggests that Procedure TS$^+$ succeeds to do so.
{\fontseries{b}\selectfont
(ii) In these experiments, the sampling effort is negligible since it only involves generating normally distributed errors.
Thus, the run time for producing the selection policy by each procedure reflects the computational overhead of each procedure.
It is found that even for the relatively large-scale Problem (12), the run time is shorter than 1 second, which indicates negligible overhead.}}
}
\end{minipage}
\end{table}%

The numerical results are shown in Table \ref{tab-PCSE}, from which we have the following observations.
First, as expected, both procedures can deliver the target $\PCSE$ in their respective domains.
Procedure TS can deliver the designed $\PCSE$ if the simulation errors are homoscedastic,
while Procedure TS$^+$ can deliver the designed $\PCSE$ even when the simulation errors are heteroscedastic.
Moreover, the achieved $\PCSE$ is higher than the target in general; see, e.g., the column ``$\APCSE$'' under ``Procedure TS'' of Table \ref{tab-PCSE}, except the entry for Problem \texttt{(6)}.
This is especially the case if the configuration of the means is not the GSC, i.e. Problem \texttt{(3)}.
Overshooting the target $\PCSE$ suggests that the total sample size is larger than necessary for meeting the target $\PCSE$.
Such conservativeness is a well known issue for R\&S procedures under the IZ formulation; see \cite{fan2016} for an exposition on the issue.

Second,  if Procedure TS is applied to the instance of heteroscedasticity, (i.e., Problem \texttt{(6)}), the target $\PCSE$ cannot be met.
By contrast, if Procedure TS$^+$ is applied to the instances of homoscedasticity, (i.e., all problem instances except \texttt{(6)}), it becomes overly conservative compared to Procedure TS.
This is reflected by the achieved $\PCSE$ being substantially higher than the target and the sample size being substantially larger than that of Procedure TS.

Third, as the number of alternative $k$ increases, which corresponds to Problems \texttt{(1)}, \texttt{(0)}, and \texttt{(2)},
the sample size allocated to each alternative on average (measured by the ratio of  the total sample size to $k$) increases as well.
This is caused by the increase in the constant $h$ as $k$ increases.
Notice that the sample size required for alternative $i$ on one design point in Procedure TS is $N_i =\max\{\lceil h^2S_i^2/\delta^2\rceil, n_0\}$.
Thus, a larger $h$ means  a larger $N_i$. A similar argument holds for Procedure TS$^+$ as well.
This suggests that as $k$ increases, each alternative must be estimated more accurately in order to differentiate them.

Fourth, the numerical results of Problems \texttt{(4)} and \texttt{(5)} are almost identical.
In particular, the value of $h$ is identical for both problems,
because  the equations that determine $h$ (Equation \eqref{eq-geth}) and $\hhet$ (Equation \eqref{eq-geth2}) do not depend on the configuration of the variances.
Then, as the sum of the variances is the same for both problems, the total sample size that is approximately proportional to $h^2$ times the sum of the variances is almost the same for both problems.

Last, the results for Problems \texttt{(10)}--\texttt{(12)} are also as expected.
They show that both Procedure TS and Procedure TS$^+$ can be used to handle relatively large-scale problems.
Note that $h$ and $h_{\mathrm{Het}}$, the key quantities for determining the second-stage sample size of the two procedures, respectively, can be computed via Monte Carlo method or the stochastic approximation method; see further discussion in \S\hyperlink{EC.2}{EC.2}.
Therefore, the computational requirement for determining the two quantities, and thus the total sample size as well as the sample allocation is negligible relative to the expenses of running the simulation model.

\section{Experimental Design and Robustness to Linearity Assumptions}\label{sec-robust}

We have assumed so far that the designs points are given with the design matrix $\cX$ satisfying that $\cX^\intercal\cX$ is nonsingular.
In this section we discuss how to select the design points.
We show that the extreme design, i.e., locating the design points at the corners of the design region $\Theta$, is typically a good strategy under the linearity assumptions (e.g., Assumptions \ref{ass-model}-\ref{ass-model4}).
In practice, however, linearity assumptions are often satisfied only approximately.
Then, the selection of design points is critically related to the robustness of the linearity assumptions.
We show through numerical experiments that the extreme design may perform poorly when the linearity assumptions are violated mildly,
but distributing the design points evenly in the design region $\Theta$ appears to be quite robust to the linearity assumptions.

\subsection{Optimal Design under Linearity Assumptions} \label{subsec-optimal}

Experimental design is a classical problem in statistics.
In classical design for linear regression, the objective is often to choose a design that optimizes a certain criterion given a fixed total sample size.
Popularly used criteria include D-optimal design that minimizes the determinant of the covariance matrix of the OLS estimator of $\BFbeta$,
G-optimal design that minimizes the maximal variance of the fitted response over the design region, and many others; see \citet[Chapter 2]{silvey1980optimal} for more details on the subject.
Some of the optimal designs are equivalent under certain conditions.
For instance, \citet{kiefer1960equivalence} prove that the D-optimal design and G-optimal design are equivalent in the continuous case (also called the approximate case)
where the integer constraint on the sample size at each design point is relaxed; see \citet[Chapter 3]{silvey1980optimal} for a more careful and complete discussion on the general equivalence theory.

However, the optimal design in R\&S-C context is different from the classical ones. In our procedures, an optimal design is the design that minimizes the total sample size required by the procedures to deliver the predetermined $\PCSE$.
Using Procedure TS as an example, the total sample size is $\sum_{i=1}^k N_i m$, where $N_i$ is approximately $h^2S_i^2/\delta^2$.
Since the design matrix $\cX=(\bx_1,\ldots,\bx_m)^\intercal$, we may formulate the optimal design problem as the following optimization problem:
\begin{eqnarray*}
\underset{m,\bx_1,\ldots,\bx_m}{\textnormal{min}} && h^2 m \\[5pt]
\textnormal{s.t.} && \E \left\{ \int_0^\infty \left[\int_0^\infty \Phi \left( \frac{h}{\sqrt{(n_0m-d) (t^{-1}+s^{-1})\bX^\intercal (\cX^\intercal \cX)^{-1} \bX}} \right) \eta(s)\ud s \right]^{k-1} \eta(t) \ud t \right\} = 1-\alpha, \\
&& \textnormal{rank}\left(\cX^\intercal\cX\right) = d,\\
&& m \ge d,\ \textnormal{integer},\\
&& \bx_1,\ldots,\bx_m \in \Theta,
\end{eqnarray*}
where the first constraint is exactly \eqref{eq-geth}, and the second constraint ensures the nonsingularity of $\cX^\intercal\cX$.
The problem is in general a nonconvex integer programming problem, and it is difficult to solve.
Moreover, even without concerning the integer constraint, the optimal design is a function of the distribution of $\bX$ and is often difficult to characterize.
In this subsection, we derive an optimal design, which is invariant to the distribution of $\bX$, for a simplified case where $m$ is fixed and an additional constraint is imposed.

\begin{figure}[]
\centering
\includegraphics[width=0.5\textwidth]{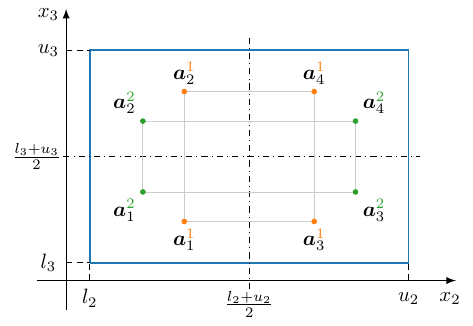}
\caption{Geometrical Illustration of the Symmetric Design for $d=3$ and $b=2$. }\label{fig-constraint}
\begin{minipage}[t]{1\linewidth}
\vspace{-1.2em}
\SingleSpacedXI
\footnotesize{
\emph{Note.} \textsf{Coordinate $x_1$ is omitted since it is always 1.}
}
\end{minipage}
\end{figure}

Specifically, we assume that
\begin{equation}\label{eq-Theta}
\Theta = \{1\}\times[l_2,u_2]\times\cdots\times[l_d,u_d], \ \text{ $d\geq 2$ and $l_w < u_w$ for all $w=2,\ldots,d$}.
\end{equation}
Here the first covariate is always 1, which is used to take care of the intercept terms in the linear models, and all other $d-1$ covariates are in an interval $[l_w,u_w]$ for $w=2,\ldots,d$.
Notice that all the following analysis can also apply to the linear models without intercept terms with similar arguments.
Suppose that we want to allocate $m = b(2^{d-1})$ design points in $\Theta$, where $b\geq 1$ is a fixed integer.
We denote these design points by $\ba_i^j \in \Theta$, $i=1,\ldots,2^{d-1}$ and $j=1,\ldots,b$.
Moreover, for any $j$, we let $\ba_1^j, \ldots, \ba_{2^{d-1}}^j$ be symmetric with respect to the Cartesian coordinate system located at the center of $\Theta$.
See Figure \ref{fig-constraint} for an illustration with $d=3$ and $b=2$.
Let $\cS^j \coloneqq \{\ba_1^j, \ldots, \ba_{2^{d-1}}^j\}$ for $j=1,\cdots,b$.
Then $\{\cS^1, \ldots, \cS^b\}$ denotes a set of $b(2^{d-1})$ design points (where duplicates are allowed) and we call it a \textit{symmetric design}.
Notice that the symmetric design ensures that $\textnormal{rank}\left(\cX^\intercal\cX\right) = d$.
The reason that we only consider symmetric designs is, without considering the distribution of the covariates $\bX$, symmetric designs are natural choices due to the symmetric nature of the design region $\Theta$.

Let $\cS^0$ denote the set of corner points of $\Theta$.
It is easy to see that $\cS^0$ has $2^{d-1}$ elements.
A simple design is to use all the points in $\cS^0$ for $b$ times, i.e., $\cS^1=\cdots= \cS^b = \cS^0$, and we call it the \textit{extreme design}.
Notice that the extreme design aims to spread out all the design points so that the OLS estimators of $\BFbeta_i$'s can have small variances.

The extreme design is also a symmetric design.
In the following theorem, we show that the extreme design is the best symmetric design regardless of the distribution of $\bX$.
The proof is included in \S\hyperlink{EC.9}{EC.9}.

\begin{theorem} \label{thm-design}
Suppose that Assumption \ref{ass-model} or \ref{ass-model3} holds, Procedure TS is used to solve an R\&S-C problem, and $m = b(2^{d-1})$ design points are allocated in $\Theta$ as assumed in \eqref{eq-Theta}.
Then, among all symmetric designs, the extreme design $\cS^1=\cdots =\cS^b = \cS^0$ minimizes the expected total sample size.
\end{theorem}

There is an interesting link between the optimal design in R\&S-C with that in classical linear regression setting.
That is, \emph{the extreme design is also both the D-optimal and G-optimal design in linear regression} when the total sample size is $b(2^{d-1})$, among all feasible designs (without the symmetry constraint).
This result is formally stated in Theorem \ref{thm-D-G-opt}, and its proof is included in \S\hyperlink{EC.10}{EC.10}, where the formal definitions of D-optimality and G-optimality are also given.
We want to emphasize that Theorem \ref{thm-D-G-opt} further justifies the consideration of the extreme designs for R\&S-C problems.

\begin{theorem} \label{thm-D-G-opt}
Consider the linear regression problem $Y(\bx) = \bx^\intercal \BFbeta + \epsilon$, where $\BFbeta, \bx \in \mathbb{R}^{d}$ and $\epsilon$ is random error with mean 0 and variance $\sigma^2$.
Let $d\geq 2$ and $x_1 \equiv 1$ so that the intercept term is included.
Suppose that $m = b(2^{d-1})$ design points are allocated in $\Theta$ as assumed in \eqref{eq-Theta}.
Then, among all feasible designs, the extreme design $\cS^1=\cdots =\cS^b = \cS^0$ is both D-optimal and G-optimal.
\end{theorem}

\begin{remark}
The total sample size of Procedure TS$^+$ depends on the variances of the design points, which are not known \textit{a priori}.
Therefore, we can only prove that, among all symmetric designs, the extreme design minimizes the constant $\hhet$ defined in Equation (\ref{eq-geth2}).
\end{remark}

\subsection{Robustness to Linearity Assumptions} \label{subsec:robust}

In practice, the linearity assumptions (i.e., Assumptions \ref{ass-model}--\ref{ass-model4}) often hold only approximately.
Notice that the linear models can be generalized to capture nonlinearity in $\mu_i(\bx)$ by the use of basis functions; see Remark \ref{remark:linear}.
Here, by saying that the linearity assumption does not hold, we actually mean that $\mu_i(\bx)$ is not linear in $\bx$ and we do not have a proper set of basis functions to perform a change of variables.

It is argued in \cite{james2013} that linear models are often robust to nonlinear behaviors and lead to good predictions.
However, the extreme design is in general not robust to nonlinearity, because it allocates no design points in the interior of the design region and leave the fitted models depending completely on the corner points.
To improve the robustness of the experimental design, one can allocate design points evenly in the design region.
One such design that is widely used is the so-called minimax design, which, roughly speaking, ensures that all points in the design region are not too far from the nearest design points.
It is shown by \cite{johnson1990minimax} that the minimax design is asymptotically equivalent to the Bayesian G-optimal design for general Gaussian processes,
which mimics the classical G-optimal design but is defined in a Bayesian framework.
In the rest of this subsection, we conduct numerical studies to compare the extreme design and the minimax design, and to understand their behaviors under different scenarios.

We consider the case where $\Theta = \{1\} \times [0,1]^{d-1}$,
and generate the true surfaces randomly from a $(d-1)$-dimensional second-order stationary Gaussian random field with mean $\bm 0$
and isotropic covariance $\Cov(\bz,\bz')= \exp\{-\lambda \| \bz-\bz' \|^2 \}$ for $\bz,\bz' \in \mathbb{R}^{d-1}$, where $\| \cdot \|$ represents the Euclidean norm.
Notice that parameter $\lambda$ controls the scale of the random field and larger $\lambda$ often leads to higher level of violation of the linearity assumption.
To keep the linearity assumptions approximately true, we discretize the surfaces with a step size $0.01$ for each coordinate, calculate the $R^2$ of the discretized observations, and only keep the surfaces whose $R^2$ is above 0.8.
We first obtain 50 such approximately linear random surfaces.
Then we randomly create 100 R\&S-C problems, each with 5 surfaces that are randomly drawn from those 50 surfaces.
We consider only the homoscedastic errors, and add normal noises with $\sigma_1^2=\cdots=\sigma_5^2=1$.

\begin{figure}[p]
\centering
\includegraphics[width=0.43\textwidth]{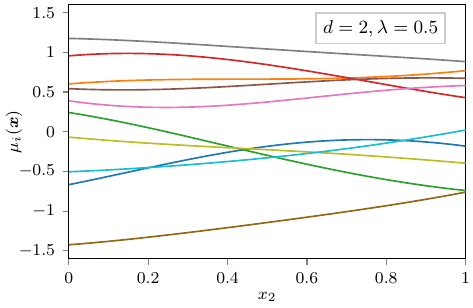} \hspace{0.025\textwidth}
\includegraphics[width=0.43\textwidth]{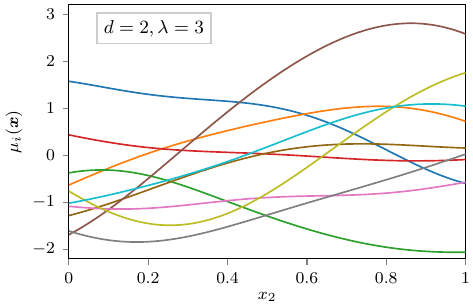}\\
\vspace{4pt}
\includegraphics[width=0.45\textwidth]{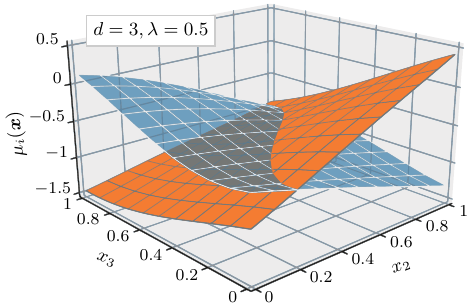} \hspace{0.01\textwidth}
\includegraphics[width=0.45\textwidth]{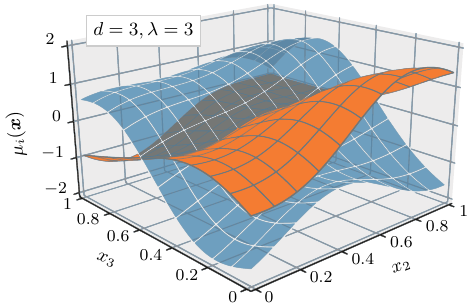}
\caption{Randomly Generated Surfaces with $R^2 \geq 0.8$.} \label{fig-shape}
\begin{minipage}[t]{1\linewidth}
\SingleSpacedXI
\vspace{-1.2em}
\footnotesize{
\emph{Note.} \textsf{In general case, nonlinear surfaces do not necessarily mean that the linearity assumption is violated.}
}
\end{minipage}
\end{figure}

We consider $\lambda=0.5$ and $\lambda=3$ to capture small and large violations of the linearity assumption.
We also consider $d=2$ and $d=3$.
Figure \ref{fig-shape} shows the typical shapes of these randomly generated surfaces.
For each R\&S-C problem, we let $X_2,\ldots,X_{d}$ be i.i.d. \textsf{Uniform}$[0,1]$ random variables, and set $\alpha=0.05$, $\delta=0.2$, $n_0=50$, $R=10^3$ and $T=10^4$.
We compare the extreme design and minimax design with $2(2^{d-1})$ points,
i.e., 4 design points when $d=2$ and 8 design points when $d=3$.
The design matrices are listed in Table \ref{tab-designs}.
We report the means and standard deviations (SD) of the average total sample size and the achieved $\PCSE$ (i.e., $\APCSE$) over 100 problems in Table \ref{tab-PCSE-sd}.
We also calculate the average \emph{regret} (also called the opportunity cost in Bayesian R\&S literature), which is defined as $\frac{1}{R} \sum_{r=1}^R \frac{1}{T} \sum_{t=1}^T \{\mu_{i^*(\bx_t)}(\bx_t) - \mu_{\widehat{i^*_r}(\bx_t)}(\bx_t)\}$.
The means and SD of the average regrets are also reported in Table \ref{tab-PCSE-sd}.

\begin{table}[p]
\caption{Extreme Designs and Minimax Designs for $d=2,3$.} \label{tab-designs}
\centering
\small
\begin{tabular}{cccccc}
\toprule
 \multicolumn{2}{c}{$d=2$} & & \multicolumn{2}{c}{$d=3$} \\
\cmidrule(lr){1-2} \cmidrule(lr){4-5}
 Extreme Design & Minimax Design & & Extreme Design & Minimax Design* \\ 
\midrule
$ \left( \begin{array}{cc} 1 & 0 \\ 1 & 1 \\ 1 & 0 \\ 1 & 1 \\ \end{array} \right)$
&
$ \left( \begin{array}{cc} 1 & 1/8 \\ 1 & 3/8 \\ 1 & 5/8 \\ 1 & 7/8 \\ \end{array} \right)$
& &
$ \left( \begin{array}{ccc} 1 & 0 & 0 \\ 1 & 0 & 1\\ 1 & 1 & 0\\ 1 & 1 & 1\\ 1 & 0 & 0 \\ 1 & 0 & 1\\ 1 & 1 & 0\\ 1 & 1 & 1\\ \end{array} \right)$
&
$ \left( \begin{array}{ccc} 1 & 0.1557 & 0.2086 \\ 1 & 0.1557 & 0.7914 \\ 1 & 0.8443 & 0.2086\\ 1 & 0.8443 & 0.7914\\
                            1 & 0.2468 & 0.5000 \\ 1 & 0.7532 & 0.5000 \\ 1 & 0.5000 & 0.1794\\ 1 & 0.5000 & 0.8206\\ \end{array} \right)$
\\
\bottomrule
\end{tabular}
\begin{minipage}[t]{0.73\linewidth}
\SingleSpacedXI
\vspace{0.6em}
\footnotesize{
\textsf{* See \cite{melissen1996improved}.}
}
\end{minipage}
\end{table}

\begin{table}[p]
\centering
\small
\caption{Means and SD (in brackets) over 100 Problems.} \label{tab-PCSE-sd}
\begin{tabular}{lccccccccccc}
\toprule
& & & & \multicolumn{3}{c}{Extreme Design} & & & \multicolumn{3}{c}{Minimax Design} \\
\cmidrule(lr){5-7} \cmidrule(lr){10-12}
Case & & $R^2$ & & Sample Size&  $\APCSE$ & Regret & & & Sample Size&  $\APCSE$ & Regret\\
\midrule
\multirow{2}{*}{$d=2, \lambda=0.5$} & & 0.965 & & 1,730 & 0.9948   & 0.007 & & & 2,869 & 0.9978  & 0.005  \\
                                   & & (0.002)& & (2)   & (0.0086) & (0.008)  & & & (5)   & (0.0037) & (0.005) \\
\addlinespace[1ex]
\multirow{2}{*}{$d=2, \lambda=3$}   & & 0.921 & & 1,730   & 0.8558  & 0.100    & & & 2,941 & 0.9799  & 0.013  \\
                                   & & (0.003) & & (2)   & (0.1394) & (0.109) & & & (50)   & (0.0297) & (0.016) \\
\addlinespace[1ex]
\multirow{2}{*}{$d=3, \lambda=0.5$} & & 0.917 & & 2,282   & 0.9528  & 0.024     & & & 4,659 & 0.9876  & 0.008  \\
                                   & & (0.003) & & (70)   & (0.0586)& (0.027)   & & & (16)   & (0.0118) & (0.007) \\
\addlinespace[1ex]
\multirow{2}{*}{$d=3, \lambda=3$}   & & 0.863 & & 2,425   & 0.7358  & 0.204     & & & 4,904  & 0.9133 & 0.047   \\
                                   & & (0.002) & &(120)   & (0.1306)& (0.139)   & & & (96)   & (0.0502) & (0.030) \\
\bottomrule
\end{tabular}
\end{table}%

From Table \ref{tab-PCSE-sd} we see that the extreme designs lead to significantly smaller total sample sizes than the minimax designs if the  linearity assumption is more or less satisfied (e.g., $\lambda=0.5$);
but their achieved $\PCSE$ and regrets are significantly poorer than those of the minimax designs if the  linearity assumption is more violated (e.g., $\lambda=3$).
Based on these observations, we have the following conclusions on experimental design and robustness on linearity assumptions.
\begin{itemize}
\item
The proposed procedures perform well when the surfaces are approximately linear, though the statistical guarantee may not always hold.
\item
When the true surfaces are exactly linear or only slightly nonlinear, the extreme design is preferred because it requires fewer samples to deliver the required $\PCSE$.
\item
When the true surfaces are relatively nonlinear, even designs, such as the minimax design, are preferred because they are more robust to nonlinearity.
\item
The above intuition suggests that when the design region $\Theta$ is of a general shape other than a hyperrectangle, it is better to allocate the design points afar from each other if the linearity is strong, while more evenly in the region if the linearity is weak.
\end{itemize}

\section{A Case Study: Personalized Treatment for Cancer Prevention} \label{sec-case}

\begin{figure}
\centering
\includegraphics[width=0.75\textwidth]{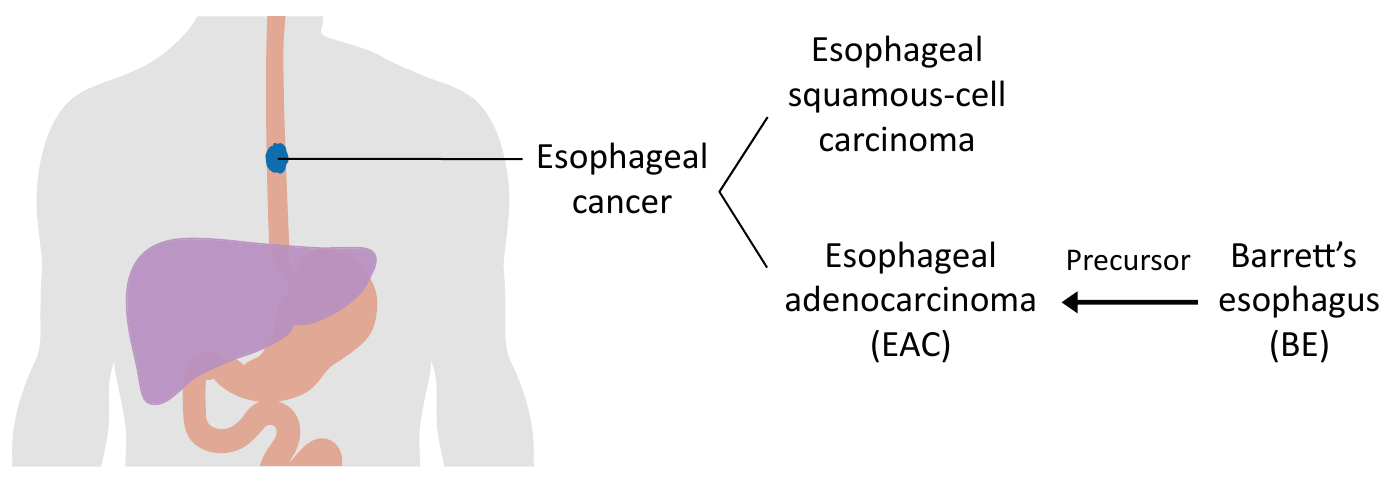}
\caption{Diagram of Esophageal Cancer.}  \label{fig-cancer}
\begin{minipage}[t]{1\linewidth}
\SingleSpacedXI
\vspace{-1.2em}
\footnotesize{
\emph{Note.} \textsf{Image source: Cancer Research UK / Wikimedia Commons, licensed under CC BY-SA 4.0.}
}
\end{minipage}
\end{figure}

Esophageal cancer (see Figure \ref{fig-cancer}) is the seventh-leading cause of cancer death among males (making up 4\%) in the United States, according to \textit{Cancer Facts \& Figures 2016} by American Cancer Society. Esophageal adenocarcinoma (EAC) is a main sub-type of esophageal cancer, and its incidence has increased by 500\% over the past 40 years \citep{hur2013,choi2014}.
Thus, the management of Barrett's esophagus (BE), a precursor to EAC, is an active topic in cancer research.
A common strategy for BE management is endoscopic surveillance, which attempts to prevent EAC through dysplasia treatment or to identify EAC before it becomes invasive.
Recently, chemoprevention has received substantial attention as a method to lower the progression of BE to EAC, and aspirin and statin are two particular drugs that are demonstrated to be effective \citep{kastelein2011}.
For each BE patient, the progression rate to cancer depends on a variety of factors including age, weight, lifestyle habits such as smoking and alcohol use, the grade of dysplasia, etc.
In addition, each patient may have a different response to drugs depending on his or her drug resistance and tolerance.
Hence, it is conceivable that the best treatment regimen for BE is patient-specific.

We formulate the problem of selecting the best treatment regimen for each BE patient as an R\&S-C problem.
There are three alternatives: endoscopic surveillance only ($i=1$), aspirin chemoprevention with endoscopic surveillance ($i=2$), and statin chemoprevention with endoscopic surveillance ($i=3$).
For simplicity, we consider only the starting age of a treatment regimen, risk (i.e., the annual progression rate of BE to EAC) and drug effects (i.e., the progression reduction effect of a drug) as patient characteristics that determine the effectiveness of a treatment regimen.
More specifically, the vector of covariates is $\bX=(1,X_1, X_2, X_3,X_4)^\intercal $, where $X_1$ is the starting age, $X_2$ is the risk, $X_3$ and $X_4$ are the drug effects of aspirin and statin, respectively.
We use the expected quality-adjusted life years (QALYs) as the performance measure to compare different alternatives.

\begin{figure}
\centering
\includegraphics[width=0.55\textwidth]{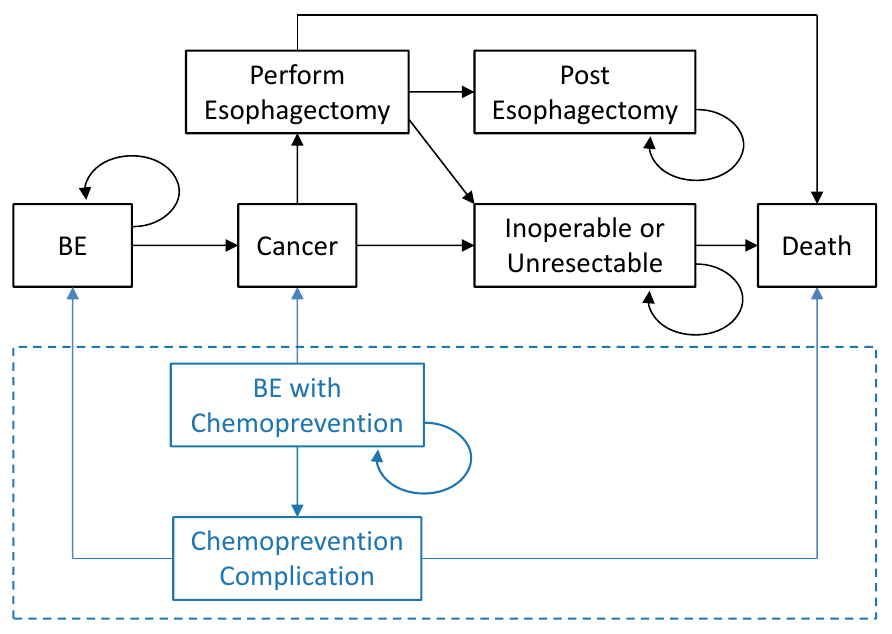}
\caption{Transition Diagram of the Markov Simulation Model.}  \label{fig-MC}
\begin{minipage}[t]{1\linewidth}
\SingleSpacedXI
\vspace{-1.2em}
\footnotesize{
\emph{Note.} \textsf{
(1) A person in each state may die from age-related all-cause mortality. (These transitions are omitted in the diagram.)
(2) The time duration between state transitions is one month.
(3) The details of the state transitions inside the dotted box depends on whether aspirin chemoprevention or  statin chemoprevention is used.
}
}
\end{minipage}
\end{figure}

To solve this R\&S-C problem, we need a model to simulate the QALYs of the treatment regimens for different patients.
Fortunately, a discrete-time Markov chain model developed by \cite{hur2004} and \cite{choi2014} may be used.
The model simulates the transitions among different health states of a BE patient until death, and the transition diagram of the model is shown in Figure \ref{fig-MC}.
The transition probability matrices are well calibrated so that the simulation outputs match the published results.
We adopt this model to simulate individual patients with specific characteristics which are defined by the covariates $\bX$ and assumed to be observable.
This Markov chain model is, of course, a highly simplified model compared to those having more detailed biological mechanisms (see for example, the MSCE-EAC model of \cite{curtius2015}).
However, as an illustrative purpose, we adopt this simple model due to its accessibility and relatively short running time,
because we need to run the model in a brute force way to obtain the mean performance surfaces of all alternative, i.e., $\mu_i(\bx)$ for $i=1,2,3$,
and use them as the true values to evaluate the performance of the proposed procedures.

In this case study, we assume that the distributions of the covariates are known because there are often ample historical data to calibrate these distributions in practice. Furthermore, we specify these distributions as follows: We assume $X_1\in[55, 80]$, as it is documented by \cite{naef1972} that there is a BE incidence peak for individuals with ages within this range.
We assume  $X_2\in[0,0.1]$ following the specification in \cite{hur2004}, and set $X_3\in[0,1]$ and $X_4\in[0,1]$ by definition.
Moreover, assume $\E[X_3]=0.53$ and $\E[X_4]=0.54$ following the study by \cite{kastelein2011}.
Nevertheless, due to lack of detailed data, we do not know the distribution of covariates exactly among the entire population of BE patients.
Instead, we suppose that $X_1,\ldots,X_4$ are independent and their distributions are specified in Table \ref{tab-dist}.
The design points are specified as follows.
We take $X_1$ from $\{61, 74\}$, $X_2$ from $\{0.1/4, 0.3/4\}$, $X_3$ from $\{1/4, 3/4\}$, and $X_4$ from $\{1/4, 3/4\}$, then combine them in a full factorial way.
Therefore, it is a relatively even design with 16 design points.

\begin{table}
\caption{Distributions of the Covariates.} \label{tab-dist}
\centering
\small
\begin{tabular}{cccc}
        \toprule
        Covariate   & Distribution & Support & Mean \\
        \midrule
        $X_1$ & \textsf{Discrete} (Figure \ref{fig-pmf}) & $\{55,\ldots,80\}$ & 64.78\\
        $X_2$ & \textsf{Uniform} $(0,0.1)$ & $[0,0.1]$ & 0.05\\
        $X_3$ & \textsf{Triangular} $(0,0.59,1)$ & $[0,1]$ & 0.53\\
        $X_4$ & \textsf{Triangular} $(0,0.62,1)$ & $[0,1]$ & 0.54\\
        \bottomrule
\end{tabular}
\end{table}

\begin{figure}
\centering
\includegraphics[width=0.5\textwidth]{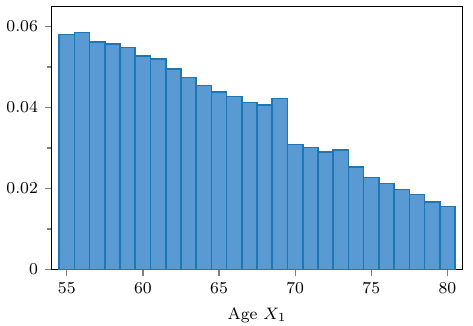}
\caption{Probability Mass Function of $X_1$ (Truncated).} \label{fig-pmf}
\begin{minipage}[t]{1\linewidth}
\SingleSpacedXI
\vspace{-1.2em}
\footnotesize{
\emph{Note.} \textsf{
Data source: U.S. 2016 population data, U.S. Census Bureau.
}
}
\end{minipage}
\end{figure}

Before carrying out the R\&S-C procedures, we conduct several trial runs of the simulation model, and we find that the linearity assumptions hold approximately and the simulation errors are clearly heteroscedastic. Therefore, Procedure TS$^+$ is used.
Notice that, to calculate the achieved $\PCSE$ (i.e., $\APCSE$) of our procedure, we need the true response surfaces $\mu_i(\bx)$, for all $\bx \in \Theta$ and $i=1,2,3$, to identify the true best selection policy $i^*(\bx)$.
To that end, we use extensive simulation to approximate the true response surfaces.
We discretize $X_2$ with a step size 0.01 and discretize $X_3$ and $X_4$ with a step size 0.1.
At each discretization point, we run the simulation model for $10^6$ replications so that the estimation error is negligible (e.g., the half-width of the 95\% confidence interval is less than 0.02 QALYs).
The response at any other $\bx$ is approximated via a linear interpolation.
To compute $\APCSE$, we conduct $R = 300$ macro-replications.
For each macro-replication $r$, we apply Procedure TS$^+$ to obtain the selection policy $\widehat{i^*_r}(\bx)$,
and then apply it to select the best treatment regimen for $T = 10^5$ simulated BE patients whose characteristics are randomly generated from the distribution of $\bX$.
Other parameters of Procedure TS$^+$ are specified as follows:
$\alpha = 0.05$, $\delta = 1/6$ (i.e., 2 months) and $n_0 = 100$.
Our case study shows that the $\APCSE = 99.5\%$, which is substantially higher than the target level $1-\alpha=95\%$.
This is because the configuration of the means of this problem is much more favorable than the GSC, and thus the selection procedure behaves in an overly conservative manner in this situation.
(Recall that Problem \texttt{(3)} in \S \ref{sec-numerical} has a similar behavior.)

\begin{remark}
In principle, one could compute the ``true'' response surfaces of this simulation model that correspond to the three treatment regimes in a brute force way subject to a discretization scheme, and then identify the best alternative for each individual patient. However, this would be very time consuming even for a coarse discretimization scheme and a moderate level of estimation accuracy as specified above. (It takes about 8 days on a desktop computer with Windows 10 OS, 3.60 GHz CPU, and 16 GB RAM to complete the above simulation implemented in MATLAB.)
By contrast, it only takes less than 1 minute for Procedure TS$^+$ to obtain a selection policy.
This demonstrates the practical value of our model and selection procedures in real-world applications.
\end{remark}

To demonstrate the usefulness of R\&S-C as a decision-making framework, we compare the personalized approach
with a more traditional approach, which selects the treatment regimen that is the best for the entire population, i.e., $i^\dagger = \argmax_{1\leq i\leq 3} \E[\mu_i(\bX)]$.
The latter corresponds to a conventional R\&S approach.
In this problem, we find $i^\dagger=3$, which indicates that alternative 3 is better than the others based on the population average.
Notice that choosing the population best, i.e., always selecting alternative 3, can also be regarded as a selection policy.
Based on our numerical study, we find that this policy correspond a $\APCSE$ of 75.8\%, i.e., alternative 3 is indeed the best or within the IZ for 75.8\% of the population.
In contrast, the personalized approach that we reported earlier has a $\APCSE$ of 99.5\%.
The 23.7\% difference in $\APCSE$  demonstrates clearly the advantage of the personalized approach.

\begin{figure}[]
 \includegraphics[width=0.49\textwidth]{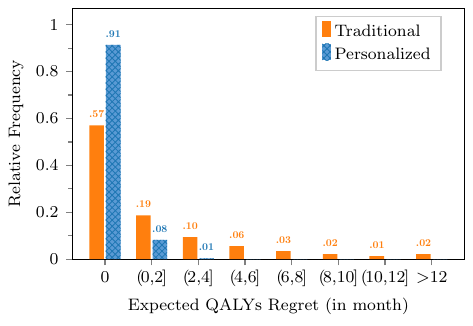}
\includegraphics[width=0.49\textwidth]{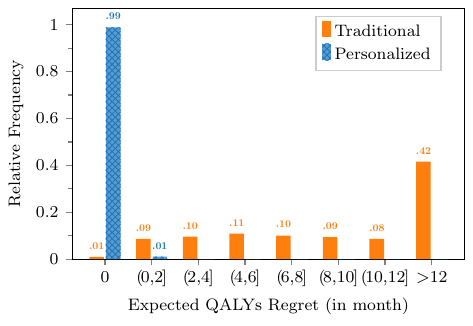}
\caption{Bar Charts of $\E[\mathrm{QALYs}|\bX]$ Regret under the Selected Treatment Regimen.}\label{fig-QALY-regret}
\begin{minipage}[t]{1\linewidth}
\SingleSpacedXI
\vspace{-1.2em}
\footnotesize{
\emph{Note.} \textsf{
Left: Entire population, $\bX \in \Theta$. Right: Specific population, $\bX=(1,X_1,X_2,0.9,0.2)^\intercal$.
}}
\end{minipage}
\end{figure}

In addition to $\PCSE$, we consider QALYs regret as another criterion to compare the two approaches.
More specifically, we consider the expected QALYs regret,
which is the expected difference between the QALYs under the true optimal treatment regimen and the selected one by each approach.
Conditionally on $\bX=\bx$, the expected regret is $\mu_{i^*(\bx)}(\bx) - \mu_{3}(\bx)$ for the traditional approach and $\mu_{i^*(\bx)}(\bx) - \mu_{\widehat{i^*}(\bx)}(\bx)$ for the personalized approach,
where $\widehat{i^*}(\bx)$ comes from one macro-replication of Procedure TS$^+$.
The results are plotted in Figure \ref{fig-QALY-regret}, where the left panel shows the distribution of regret for the entire BE population (i.e., $\bX \in \Theta$),
and the right panel shows the distribution of regret for a specific group of patients (i.e., $X_3=0.9$ and $X_4=0.2$).

From these results, we see that, using the personalized approach (i.e., the R\&S-C approach), the BE patients have much lower expected QALYs regret than using the traditional approach (i.e., the conventional R\&S approach).
Among the entire BE population (left panel of Figure \ref{fig-QALY-regret}),
when the personalized approach is used, over 99\% of the patients have either no regret or a regret that is less than or equal to 2 months (i.e., the IZ parameter).
However, when the traditional approach is used, close to a quarter (i.e., 24\%) of the patients have a regret that  is more than 2 months and 2\% of them have a regret that is above 12 months.

If we look at the specific group of patients, e.g., the group as considered in the right panel of Figure \ref{fig-QALY-regret},
we see that the reduction of the regret using the personalized approach is even more substantial, which demonstrates the key point of personalized medicine,
that is, a universal treatment, even it seems fairly good for the entire population, may perform quite poorly for certain groups of patients, where we can do much better with the help of personalized medicine.

\section{Conclusions} \label{sec-conclusion}

Ranking and selection is a long-standing research problem in simulation literature.
The emerging popularity of personalized decision making leads us to consider this classical problem in a new environment where the performance of an alternative depends on some observable random covariates.
A critical feature in the new setting is that the goal is not to seek a single alternative having a superior performance, but a selection policy as a function of the covariates.
Albeit computed offline via simulation model, the selection policy can be applied online to specify the best alternative for the subsequent individuals after observing their covariates.
Therefore, R\&S-C reflects a shift in perspective regarding the role of simulation:  a tool for system control instead of system design.
In particular, we demonstrate the practical value of R\&S-C via a case study of personalized medicine for selecting the best treatment regimen in prevention of esophageal cancer.

This paper uses a linear model to capture the relationship between the response of an alternative and the covariates, and develops two-stage selection procedures accordingly under the IZ formulation.
However, the presence of covariates complicates the concept of PCS, since the best alternative varies as a function of the covariates.
We define statistical validity of a procedure in terms of average PCS, while other forms of unconditional PCS are also possible.
This paper is a first step towards understanding R\&S-C problems under a frequentist perspective.
There are many potential directions for future work such as nonparametric models and sequential selection procedures.


%
%
%




\vspace{1cm}

\pdfbookmark[1]{References}{link8}
\bibliographystyle{informs2014} 
\bibliography{RSCov} 

\begin{thebibliography}{48}
\providecommand{\natexlab}[1]{#1}
\providecommand{\url}[1]{\texttt{#1}}
\providecommand{\urlprefix}{URL }

\bibitem[{Arora et~al.(2008)Arora, Dreze, Ghose, Hess et~al.}]{arora2008}
Arora N, Dreze X, Ghose A, Hess JD, et~al. (2008) Putting one-to-one marketing
  to work: Personalization, customization, and choice. \emph{Market. Lett.}
  19:305--321.

\bibitem[{Auer(2002)}]{auer2002b}
Auer P (2002) Using confidence bounds for exploitation-exploration trade-offs.
  \emph{J. Mach. Learn. Res.} 3:397--422.

\bibitem[{Bechhofer(1954)}]{Bechhofer54}
Bechhofer RE (1954) A single-sample multiple decision procedure for ranking
  means of normal populations with known variances. \emph{Ann. Math. Stat.}
  25(1):16--39.

\bibitem[{Bubeck \protect\BIBand{} Cesa-Bianchi(2012)}]{bubeck2012}
Bubeck S, Cesa-Bianchi N (2012) Regret analysis of stochastic and nonstochastic
  multi-armed bandit problems. \emph{Found. Trends Mach. Learn.} 5(1):1--122.

\bibitem[{Chen et~al.(2015)Chen, Chick, Lee, \protect\BIBand{}
  Pujowidianto}]{ChenChickLeePujowidianto15}
Chen CH, Chick SE, Lee LH, Pujowidianto NA (2015) Ranking and selection:
  Efficient simulation budget allocation. Fu MC, ed., \emph{Handbook of
  Simulation Optimization}, 45--80 (Springer).

\bibitem[{Chen et~al.(1997)Chen, Chen, Dai, \protect\BIBand{}
  Y\"{u}cesan}]{ChenChenDaiYucesan97}
Chen HC, Chen CH, Dai L, Y\"{u}cesan E (1997) New development of optimal
  computing budget allocation for discrete event simulation. \emph{Proc. 1997
  Winter Simul. Conf.}, 334--341.

\bibitem[{Chick \protect\BIBand{} Frazier(2012)}]{chick2012}
Chick SE, Frazier PI (2012) Sequential sampling with economics of selection
  procedures. \emph{Manag. Sci.} 58(3):550--569.

\bibitem[{Chick \protect\BIBand{} Gans(2009)}]{ChickGans09}
Chick SE, Gans N (2009) Economic analysis of simulation selection problems.
  \emph{Manag. Sci.} 55(3):421--437.

\bibitem[{Chick \protect\BIBand{} Inoue(2001)}]{chick2001OR}
Chick SE, Inoue K (2001) New two-stage and sequential procedures for selecting
  the best simulated system. \emph{Oper. Res.} 49(5):732--743.

\bibitem[{Choi et~al.(2014)Choi, Perzan, Tramontano, Kong, \protect\BIBand{}
  Hur}]{choi2014}
Choi SE, Perzan KE, Tramontano AC, Kong CY, Hur C (2014) Statins and aspirin
  for chemoprevention in {B}arrett's esophagus: Results of a cost-effectiveness
  analysis. \emph{Canc. Prev. Res.} 7(3):341--350.

\bibitem[{Curtius et~al.(2015)Curtius, Hazelton, Jeon, \protect\BIBand{}
  Luebeck}]{curtius2015}
Curtius K, Hazelton WD, Jeon J, Luebeck EG (2015) A multiscale model evaluates
  screening for neoplasia in {B}arrett's esophagus. \emph{PLoS Comput. Biol.}
  11(5):e1004272.

\bibitem[{Dudewicz \protect\BIBand{} Dalal(1975)}]{dudewicz1975}
Dudewicz EJ, Dalal SR (1975) Allocation of observations in ranking and
  selection with unequal variances. \emph{Sankhy{\=a} B} 28--78.

\bibitem[{Fan et~al.(2016)Fan, Hong, \protect\BIBand{} Nelson}]{fan2016}
Fan W, Hong LJ, Nelson BL (2016) Indifference-zone-free selection of the best.
  \emph{Oper. Res.} 64(6):1499--1514.

\bibitem[{Frazier(2014)}]{Frazier14}
Frazier PI (2014) A fully sequential elimination procedure for
  indifference-zone ranking and selection with tight bounds on probability of
  correct selection. \emph{Oper. Res.} 62(4):926--942.

\bibitem[{Frazier et~al.(2008)Frazier, Powell, \protect\BIBand{}
  Dayanik}]{frazier2008}
Frazier PI, Powell WB, Dayanik S (2008) A knowledge-gradient policy for
  sequential information collection. \emph{SIAM J. Control Optim.}
  47(5):2410--2439.

\bibitem[{Goldenshluger \protect\BIBand{} Zeevi(2013)}]{GoldenshlugerZeevi13}
Goldenshluger A, Zeevi A (2013) A linear response bandit problem. \emph{Stoch.
  Syst.} 3(1):230--261.

\bibitem[{Gupta \protect\BIBand{} Miescke(1982)}]{gupta1982}
Gupta SS, Miescke KJ (1982) On the least favorable configurations in certain
  two-stage selection procedures. Kallianpur G, Krishnaiah PR, Ghosh JK, eds.,
  \emph{Statistics and Probability: Essays in Honor of C. R. Rao}, 295--305
  (North Holland).

\bibitem[{Hong(2006)}]{hong2006}
Hong LJ (2006) Fully sequential indifference-zone selection procedures with
  variance-dependent sampling. \emph{Naval Res. Logist.} 53(5):464--476.

\bibitem[{Hu \protect\BIBand{} Ludkovski(2017)}]{hu2017sequential}
Hu R, Ludkovski M (2017) Sequential design for ranking response surfaces.
  \emph{SIAM/ASA J. Uncertain. Quantif.} 5(1):212--239.

\bibitem[{Hur et~al.(2013)Hur, Miller, Kong, Dowling, Nattinger, Dunn,
  \protect\BIBand{} Feuer}]{hur2013}
Hur C, Miller M, Kong CY, Dowling EC, Nattinger KJ, Dunn M, Feuer EJ (2013)
  Trends in esophageal adenocarcinoma incidence and mortality. \emph{Cancer}
  119(6):1149--1158.

\bibitem[{Hur et~al.(2004)Hur, Nishioka, \protect\BIBand{} Gazelle}]{hur2004}
Hur C, Nishioka NS, Gazelle GS (2004) Cost-effectiveness of aspirin
  chemoprevention for {B}arrett's esophagus. \emph{J. Natl. Canc. Inst.}
  96(4):316--325.

\bibitem[{James et~al.(2013)James, Witten, Hastie, \protect\BIBand{}
  Tibshirani}]{james2013}
James G, Witten D, Hastie T, Tibshirani R (2013) \emph{An Introduction to
  Statistical Learning}, volume 112 (Springer).

\bibitem[{Johnson et~al.(1990)Johnson, Moore, \protect\BIBand{}
  Ylvisaker}]{johnson1990minimax}
Johnson ME, Moore LM, Ylvisaker D (1990) Minimax and maximin distance designs.
  \emph{J. Statist. Plann. Inference} 26(2):131--148.

\bibitem[{Kastelein et~al.(2011)Kastelein, Spaander, Biermann
  et~al.}]{kastelein2011}
Kastelein F, Spaander MCW, Biermann K, et~al. (2011) Nonsteroidal
  anti-inflammatory drugs and statins have chemopreventative effects in
  patients with {Barrett}'s esophagus. \emph{Gastroenterology}
  141(6):2000--2008.

\bibitem[{Katrakazas et~al.(2015)Katrakazas, Quddus, Chen, \protect\BIBand{}
  Deka}]{katrakazas2015real}
Katrakazas C, Quddus M, Chen WH, Deka L (2015) Real-time motion planning
  methods for autonomous on-road driving: State-of-the-art and future research
  directions. \emph{Transp. Res. C: Emerging Technol.} 60:416--442.

\bibitem[{Kiefer \protect\BIBand{} Wolfowitz(1960)}]{kiefer1960equivalence}
Kiefer J, Wolfowitz J (1960) The equivalence of two extremum problems.
  \emph{Canad. J. Math.} 12:363--366.

\bibitem[{Kim et~al.(2011)Kim, Herbst, Wistuba, Lee et~al.}]{kim2011short}
Kim ES, Herbst RS, Wistuba II, Lee JJ, et~al. (2011) The {BATTLE} trial:
  Personalizing therapy for lung cancer. \emph{Canc. Discov.} 1(1):44--53.

\bibitem[{Kim \protect\BIBand{} Nelson(2001)}]{kim2001}
Kim SH, Nelson BL (2001) A fully sequential procedure for indifference-zone
  selection in simulation. \emph{ACM Trans. Model. Comput. Simul.}
  11(3):251--273.

\bibitem[{Kim \protect\BIBand{} Nelson(2006)}]{kim2006}
Kim SH, Nelson BL (2006) Selecting the best system. Henderson SG, Nelson BL,
  eds., \emph{Handbooks in Operations Research and Management Science},
  volume~13, 501--534 (Elsevier).

\bibitem[{Law \protect\BIBand{} Kelton(2000)}]{law2000simulation}
Law AM, Kelton WD (2000) \emph{Simulation Modeling and Analysis} (McGraw-Hill
  New York), 3rd edition.

\bibitem[{Luo et~al.(2015)Luo, Hong, Nelson, \protect\BIBand{} Wu}]{luo2015}
Luo J, Hong LJ, Nelson BL, Wu Y (2015) Fully sequential procedures for
  large-scale ranking-and-selection problems in parallel computing
  environments. \emph{Oper. Res.} 63(5):1177--1194.

\bibitem[{Melissen \protect\BIBand{} Schuur(1996)}]{melissen1996improved}
Melissen JBM, Schuur PC (1996) Improved coverings of a square with six and
  eight equal circles. \emph{Electron. J. Combin.} 3(1):R32,10 {pp.}
  (electronic).

\bibitem[{Naef \protect\BIBand{} Savary(1972)}]{naef1972}
Naef A, Savary M (1972) Conservative operations for peptic esophagitis with
  stenosis in columnar-lined lower esophagus. \emph{Ann. Thorac. Surg.}
  13(6):543--551.

\bibitem[{Negoescu et~al.(2011)Negoescu, Frazier, \protect\BIBand{}
  Powell}]{negoescu2011}
Negoescu DM, Frazier PI, Powell WB (2011) The knowledge-gradient algorithm for
  sequencing experiments in drug discovery. \emph{INFORMS J. Comput.}
  23(3):346--363.

\bibitem[{Ni et~al.(2017)Ni, Ciocan, Henderson, \protect\BIBand{}
  Hunter}]{ni2017efficient}
Ni EC, Ciocan DF, Henderson SG, Hunter SR (2017) Efficient ranking and
  selection in parallel computing environments. \emph{Oper. Res.}
  65(3):821--836.

\bibitem[{Pearce \protect\BIBand{} Branke(2017)}]{pearce2017efficient}
Pearce M, Branke J (2017) Efficient expected improvement estimation for
  continuous multiple ranking and selection. \emph{Proc. 2017 Winter Simul.
  Conf.}, 2161--2172.

\bibitem[{Perchet \protect\BIBand{} Rigollet(2013)}]{perchet2013}
Perchet V, Rigollet P (2013) The multi-armed bandit problem with covariates.
  \emph{Ann. Stat.} 41(2):693--721.

\bibitem[{Rencher \protect\BIBand{} Schaalje(2008)}]{RencherSchaalje08}
Rencher AC, Schaalje GB (2008) \emph{Linear Models in Statistics} (John Wiley
  \& Sons, Inc.), 2nd edition.

\bibitem[{Rinott(1978)}]{rinott1978}
Rinott Y (1978) On two-stage selection procedures and related
  probability-inequalities. \emph{Comm. Stat. Theor. Meth.} 7(8):799--811.

\bibitem[{Robbins(1952)}]{robbins1952}
Robbins H (1952) Some aspects of the sequential design of experiments.
  \emph{Bull. Am. Math. Soc.} 58(5):527--535.

\bibitem[{Robbins \protect\BIBand{} Monro(1951)}]{robbins1951stochastic}
Robbins H, Monro S (1951) A stochastic approximation method. \emph{Ann. Math.
  Stat.} 22(3):400--407.

\bibitem[{Rusmevichientong \protect\BIBand{}
  Tsitsiklis(2010)}]{rusmevichientong2010}
Rusmevichientong P, Tsitsiklis JN (2010) Linearly parameterized bandits.
  \emph{Math. Oper. Res.} 35(2):395--411.

\bibitem[{Shen et~al.(2017)Shen, Hong, \protect\BIBand{}
  Zhang}]{ShenHongZhang17}
Shen H, Hong LJ, Zhang X (2017) Ranking and selection with covariates.
  \emph{Proc. 2017 Winter Simul. Conf.}, 2137--2148.

\bibitem[{Silvey(1980)}]{silvey1980optimal}
Silvey SD (1980) \emph{Optimal Design: An Introduction to the Theory for
  Parameter Estimation} (Chapman and Hall).

\bibitem[{Slivkins(2014)}]{slivkins2014}
Slivkins A (2014) Contextual bandits with similarity information. \emph{J.
  Mach. Learn. Res.} 15(1):2533--2568.

\bibitem[{Stein(1945)}]{stein1945}
Stein C (1945) A two-sample test for a linear hypothesis whose power is
  independent of the variance. \emph{Ann. Math. Stat.} 16(3):243--258.

\bibitem[{Yap et~al.(2009)Yap, Carden, \protect\BIBand{} Kaye}]{yap2009}
Yap TA, Carden CP, Kaye SB (2009) Beyond chemotherapy: Targeted therapies in
  ovarian cancer. \emph{Nat. Rev. Canc.} 9(3):167--181.

\bibitem[{Zhong \protect\BIBand{} Hong(2020)}]{zhong2020}
Zhong Y, Hong LJ (2020) Knockout-tournament procedures for large-scale ranking
  and selection in parallel computing environments. \emph{Working Paper.}
  \urlprefix\url{https://www.researchgate.net/publication/339200267}.

\end{thebibliography}


\begin{thebibliography}{}

  \bibitem[{Horn \protect\BIBand{} Johnson(2013)}]{horn1990matrix_ec}
Horn RA, Johnson CR (2013) \emph{Matrix Analysis} (Cambridge University Press), 2nd edition.

  \bibitem[{Kiefer \protect\BIBand{} Wolfowitz(1960)}]{kiefer1960equivalence_ec}
Kiefer J, Wolfowitz J (1960) The equivalence of two extremum problems. \emph{Canad. J. Math.} 12:363--366.

  \bibitem[{Rencher \protect\BIBand{} Schaalje(2008)}]{RencherSchaalje08_ec}
Rencher AC, Schaalje GB (2008) \emph{Linear Models in Statistics} (John Wiley \& Sons, Inc.), 2nd edition.

  \bibitem[{Robbins \protect\BIBand{} Monro(1951)}]{robbins1951stochastic_ec}
Robbins H, Monro S (1951) A stochastic approximation method. \emph{Ann. Math. Stat.} 22(3):400--407.

  \bibitem[Rockafellar (1970)]{rockafellar1970_ec}
Rockafellar RT (1970) \emph{Convex Analysis} (Princeton University Press).

  \bibitem[{Silvey(1980)}]{silvey1980optimal_ec}
Silvey SD (1980) \emph{Optimal Design: An Introduction to the Theory for Parameter Estimation} (Chapman and Hall).

  \bibitem[{Slepian(1962)}]{slepian1962_ec}
Slepian D (1962) The one-sided barrier problem for {Gaussian} noise. \emph{Bell System Tech. J.} 41(2):463--501.

  \bibitem[Zhong \protect\BIBand{} Hong(2020)]{zhong2020_ec}
Zhong Y, Hong LJ (2020) Knockout-tournament procedures for large-scale ranking and selection in parallel computing environments.
\emph{Working Paper.} URL \url{https://www.researchgate.net/publication/339200267}.

\end{thebibliography}

\ECSwitch


\pdfbookmark[1]{E-Companion}{link-e0}
\ECHead{Technical Proofs and Additional Discussions}

\pdfbookmark[2]{Proof of Lemma \ref{lem-stein}}{link-e1}
\hypertarget{EC.1}{
\section*{EC.1. \hspace{5pt} Proof of Lemma \ref{lem-stein}}
}

The following Lemma \ref{lem-stein2} is a more general version of Lemma \ref{lem-stein} in \S \ref{sec-linear}.
Therefore we only provide the proof of Lemma \ref{lem-stein2} and remark that Lemma \ref{lem-stein} is a special case.
Also note that Lemma \ref{lem-stein2} is used directly in the proof of Theorem \ref{thm-het}.

\begin{lemma} \phantomsection \label{lem-stein2}
For each $j=1,\ldots,m$, let $Y(\bx_j) = \bx_j^\intercal \BFbeta + \epsilon(\bx_j)$, where $\BFbeta, \bx_j \in \mathbb{R}^{d}$ and  $\epsilon(\bx_j) \sim \cN(0,\sigma_{j}^2)$.
Suppose that $\epsilon(\bx_1), \ldots, \epsilon(\bx_m)$ are independent. Let $Y_{1}(\bx_j), Y_{2}(\bx_j), \ldots$ be independent samples of $Y(\bx_j)$.
Let $T$ be a set of random variables independent of $\sum_{\ell=1}^n  Y_{\ell}(\bx_j)$ and of $\{Y_\ell(\bx_j):\ell\geq n+1\}$, for all $j=1,\ldots,m$.
Suppose $N_j\geq n$ is  an integer-valued function of $T$ and no other random variables.
Let $\widehat{Y}_j =N_j^{-1} \sum_{\ell=1}^{N_j} Y_{\ell}(\bx_j)$, $\widehat{\bY} = (\widehat{Y}_1, \ldots, \widehat{Y}_m)^\intercal$, $\cX=(\bx_1, \ldots, \bx_m)^\intercal$,
$\widehat{\BFbeta} = (\cX^\intercal \cX)^{-1} \cX^\intercal \widehat{\bY}$, and $\Sigma = \mathrm{Diag}( \sigma_{1}^2/N_1, \ldots, \sigma_{m}^2/N_m )$.
Then, for any $\bx \in \mathbb{R}^d$,
\begin{enumerate}[label=(\roman*)]
\item
$\bx^\intercal \widehat{\BFbeta} \big| T \; \sim \; \cN(\bx^\intercal \BFbeta, \bx^\intercal (\cX^\intercal \cX)^{-1} \cX^\intercal \Sigma \cX (\cX^\intercal \cX)^{-1} \bx)$;
\item
\vspace{3pt}
$ \displaystyle \frac{\bx^\intercal \widehat{\BFbeta} - \bx^\intercal \BFbeta}{\sqrt{\bx^\intercal (\cX^\intercal \cX)^{-1} \cX^\intercal \Sigma \cX (\cX^\intercal \cX)^{-1} \bx}}$
is independent of $T$  and has the standard normal distribution.
\end{enumerate}
\end{lemma}

\proof{Proof.}
For part (i), by the definition of $\widehat{\BFbeta}$, it suffices to show that $\widehat{\bY} \big| T \sim \cN(\cX \BFbeta, \Sigma)$.
We first notice that $Y(\bx_j) \sim \cN(\bx_j^\intercal \BFbeta, \sigma_j^2)$. Since $T$ is independent of $\sum_{\ell=1}^n Y_{\ell}(\bx_j)$,
\[\sum_{\ell=1}^n Y_{\ell}(\bx_j) \Big|T\;\sim\; \cN(n \bx_j^\intercal \BFbeta, n \sigma_j^2).\]
On the other hand, since $T$ is independent of  $\{Y_\ell(\bx_j):\ell\geq n+1\}$ and $N_j$ is a function only of $T$,
\[\sum_{\ell=n+1}^{N_j} Y_{\ell}(\bx_j) \Big|T \;\sim\; \cN((N_j-n) \bx_j^\intercal \BFbeta, (N_j-n)\sigma_j^2).\]
Since $\sum_{\ell=1}^n Y_{\ell}(\bx_j)$ and $\sum_{\ell=n+1}^{N_j} Y_{\ell}(\bx_j)$ are independent,
\[\widehat{Y}_j \Big| T = \frac{1}{N_j} \left( \sum_{\ell=1}^{n} Y_{\ell}(\bx_j) + \sum_{\ell=n+1}^{N_j} Y_{\ell}(\bx_j) \right) \Big| T
\;\sim\; \cN \left(\bx_j^\intercal \BFbeta,  \sigma_j^2 /N_j\right).\]
Notice that  $\widehat{Y}_1, \ldots, \widehat{Y}_m$ are independent conditionally on $T$, so $\widehat{\bY}  \big| T \sim N(\cX \BFbeta, \Sigma)$.

For part (ii),  let
\[V=\frac{\bx^\intercal \widehat{\BFbeta} - \bx^\intercal \BFbeta}{\sqrt{\bx^\intercal (\cX^\intercal \cX)^{-1} \cX^\intercal \Sigma \cX (\cX^\intercal \cX)^{-1} \bx}},\]
then $V|T \sim \cN(0,1)$ by part (i). Notice that $\pr (V<v|T)=\Phi(v)$ is not a function of $T$ for any $v$, so $V$ is independent of $T$.
 \Halmos
\endproof

\begin{remark}
It is easy to see that Lemma \ref{lem-stein} in \S \ref{sec-linear} is a special case of Lemma \ref{lem-stein2} with $\sigma_1 = \cdots = \sigma_m = \sigma$ and $N_1 = \cdots = N_m = N$.
\end{remark}

\pdfbookmark[2]{Computing $h$ in High Dimensions}{link-e2}
\hypertarget{EC.2}{
\section*{EC.2. \hspace{5pt} Computing $h$ in High Dimensions}
}

If $\bX$ is high-dimensional, the numerical integration in \eqref{eq-geth} for computing $h$ suffers from the curse of dimensionality. For instance, the error in the trapezoidal rule for $d$-dimensional numerical integration is $\mathcal{O}(n^{-2/d})$ in general. One solution is the Monte Carlo method. Let
$$f(\bx,h) \coloneqq \int_0^\infty \left[\int_0^\infty \Phi \left( \frac{h}{\sqrt{(n_0m-d) (t^{-1}+s^{-1})\bx^\intercal (\cX^\intercal \cX)^{-1} \bx}} \right) \eta(s)\ud s \right]^{k-1} \eta(t) \ud t,$$
and generate $n$ i.i.d. samples of $\bX$, $\bx_1,\ldots,\bx_n$.
Then, $\E[f(\bX,h)]$, the left-hand side of \eqref{eq-geth}, can be approximated by $n^{-1}\sum_{i=1}^n f(\bx_i,h)$ with error $\mathcal{O}(n^{-1/2})$. So the Monte Carlo method is more efficient when $d>4$. We can then solve $n^{-1} \sum_{i=1}^n f(\bx_i,h) = 1-\alpha$ for $h$ by using the MATLAB built-in root finding function \texttt{fzero}.

Another approach to computing $h$ in high dimensions is the stochastic approximation method \citep{robbins1951stochastic_ec}.
Given an initial value $h_0\geq 0$, define
$h_{n+1} = \Pi \left\{ h_n - a_n (f(\bx_n,h_n) - (1-\alpha)) \right\}$, where $\Pi\left\{\cdot\right\}$ denotes a projection that maps a point outside $[0,\infty)$ to $[0,\infty)$ (e.g., $\Pi\{\cdot\}=\|\cdot\|$ or $\Pi\{\cdot\}=\max\{0,\cdot\}$),
$\bx_n$ is an independent realization of $\bX$, and $\{a_n\}$ is a sequence of constants satisfying $\sum_{n=0}^\infty a_n = \infty$ and $\sum_{n=0}^\infty a_n^2 < \infty$.
A common choice of $\{a_n\}$ is $a_n = a/n$, for some $a>0$. It can be shown that $h_n$ converges to $h$ at a rate of $\mathcal{O}(n^{-1/2})$.

\pdfbookmark[2]{Proof of Theorem \ref{thm-hom}}{link-e3}
\hypertarget{EC.3}{
\section*{EC.3. \hspace{5pt} Proof of Theorem \ref{thm-hom}}
}

The proof of Theorem \ref{thm-hom} critically relies on the extended Stein's lemma (Lemma \ref{lem-stein}).
It also needs the following lemma, often known as Slepian's Inequality \citep{slepian1962_ec}.

\begin{lemma}[Slepian's Inequality]\label{lem-slepian}
Suppose that $(Z_1,\ldots,Z_k)^\intercal$ has a multivariate normal distribution. If $\Cov(Z_i, Z_j)\geq 0$ for all $1\leq i,j\leq k$, then, for any  constants $c_i$, $i=1,\ldots,k$,
\[\pr\left(\bigcap_{i=1}^k\{Z_i\geq c_i\}\right)\geq \prod_{i=1}^k \pr (Z_i\geq c_i).\]
\end{lemma}

\begin{proof}{Proof of Theorem \ref{thm-hom}.}
Notice that $N_i$  is an integer-valued function only of $S_i^2$, which is the OLS estimator of $\sigma^2$.
Under Assumption \ref{ass-model}, by Lemma \ref{lem-stein} and Remark \ref{remark:T},
\begin{equation}\label{eq:normal}
\bX^\intercal \widehat{\BFbeta}_i \Big | \left(\bX, S_i^2\right) \; \sim \; \cN \left(\bX^\intercal \BFbeta_i,\frac{\sigma_i^2}{N_i}\bX^\intercal (\cX^\intercal \cX)^{-1} \bX \right ),\quad i=1,\ldots,k.
\end{equation}
Moreover, let $\xi_i = (n_0m-d)S_i^2/\sigma_i^2$ for all $i=1,\ldots,k$.
Then, $\xi_i$ has the chi-square distribution with $(n_0m-d)$ degrees of freedom, for $i=1,\ldots,k$ (see Remark \ref{remark:T}).

For notational simplicity, we let $V(\bX)\coloneqq \bX^\intercal (\cX^\intercal \cX)^{-1} \bX$ and temporarily write $i^*=i^*(\bX)$ to suppress the dependence on $\bX$.
Let $\Omega(\bx) \coloneqq \{i: \bX^\intercal \BFbeta_{i^*} - \bX^\intercal \BFbeta_i  \geq \delta | \bX=\bx\}$ be the set of alternatives outside the IZ given $\bX=\bx$.
For each $i \in \Omega(\bX)$, $\bX^\intercal \widehat{\BFbeta}_{i^*}$ is independent of $\bX^\intercal \widehat{\BFbeta}_i$ given $\bX$.
It then follows from \eqref{eq:normal} that
\begin{equation} \label{eq-normal}
\bX^\intercal \widehat{\BFbeta}_{i^*} - \bX^\intercal \widehat{\BFbeta}_i  \Big| \left(\bX, S_{i^*}^2, S_i^2\right)\; \sim\; \cN\left(\bX^\intercal \BFbeta_{i^*} - \bX^\intercal \BFbeta_i, ( \sigma_{i^*}^2 / N_{i^*}+\sigma_i^2/N_i) V(\bX) \right).
\end{equation}
Hence, letting  $Z$ denote a standard normal random variable, for each $i\in \Omega(\bX)$, we have
\begin{align}
 \pr \left(\bX^\intercal \widehat{\BFbeta}_{i^*} - \bX^\intercal \widehat{\BFbeta}_i  > 0 \Big| \bX,S_{i^*}^2,S_i^2 \right)
&= \pr \left(Z
>  \frac{ -(\bX^\intercal \BFbeta_{i^*} - \bX^\intercal \BFbeta_i )}{\sqrt{ (\sigma_{i^*}^2/N_{i^*}+\sigma_i^2/N_i) V(\bX)}} \Bigg| \bX,S_{i^*}^2,S_i^2 \right)\nonumber \\
&\geq \pr \left( Z
>  \frac{-\delta}{\sqrt{[\sigma_{i^*}^2  \delta^2 /(h^2 S_{i^*}^2) +\sigma_i^2  \delta^2 /(h^2 S_i^2)] V(\bX)}} \Bigg| \bX,S_{i^*}^2,S_i^2 \right)\nonumber \\
& =  \Phi\left( \frac{h}{\sqrt{(n_0m-d)(\xi_{i^*}^{-1} + \xi_i^{-1})V(\bX)}}  \right),  \label{eq-normalP}
\end{align}
where the inequality follows the definitions of $\Omega(\bX)$ and $N_i$, and the last equality follows the definition of $\xi_i$.

Then, conditionally on $\bX$, by the definition \eqref{eq-PCSX}, the CS event must occur if alternative $i^*$ eliminates all alternatives in $\Omega(\bX)$.
Thus,
\begin{align}\label{eq:lb_pcs}
\mathrm{PCS}(\bX) &\geq \pr \left( \bigcap_{i \in \Omega(\bX)} \left\{ \bX^\intercal \widehat{\BFbeta}_{i^*} - \bX^\intercal \widehat{\BFbeta}_i  > 0 \right\} \bigg| \bX \right) \nonumber \\
&= \E \left[ \pr \left( \bigcap_{i \in \Omega(\bX)} \left\{ \bX^\intercal \widehat{\BFbeta}_{i^*} - \bX^\intercal \widehat{\BFbeta}_i  > 0 \right\} \bigg| \bX, S_{i^*}^2, \left\{S_i^2: i \in \Omega(\bX) \right\} \right) \Bigg| \bX \right],
\end{align}
where the equality is due to the tower law of conditional expectation.
Notice that  conditionally on $\{\bX, S_{i^*}^2, \{S_i^2: i\in\Omega(\bX) \} \}$, $\{\bX^\intercal \widehat{\BFbeta}_{i^*} - \bX^\intercal \widehat{\BFbeta}_i: i\in\Omega(\bX)\}$ is multivariate normal by \eqref{eq-normal}.
Moreover, for $i,i' \in \Omega(\bX)$ and $i\neq i'$, due to the conditional independence between $\bX^\intercal \widehat{\BFbeta}_i$  and  $\bX^\intercal \widehat{\BFbeta}_{i'}$,
\[\Cov\left(\bX^\intercal \widehat{\BFbeta}_{i^*} - \bX^\intercal \widehat{\BFbeta}_i, \bX^\intercal \widehat{\BFbeta}_{i^*} - \bX^\intercal \widehat{\BFbeta}_{i'} \Big| \bX, S_{i^*}^2, \left\{ S_i^2: i\in\Omega(\bX) \right\}\right)= \Var\left(\bX^\intercal \widehat{\BFbeta}_{i^*} \Big| \bX, S_{i^*}^2\right)>0.\]
Therefore, applying  \eqref{eq:lb_pcs} and Lemma \ref{lem-slepian}, we have
\begin{align}
\mathrm{PCS}(\bX) &\geq \E \left[ \prod_{i\in\Omega(\bX)} \pr \left( \bX^\intercal \widehat{\BFbeta}_{i^*} - \bX^\intercal \widehat{\BFbeta}_i  > 0 \Big| \bX, S_{i^*}^2, S_i^2\right)  \Bigg| \bX \right]  \nonumber\\
&\geq \E \left[ \prod_{i\in\Omega(\bX)} \Phi\left( \frac{h}{\sqrt{(n_0m-d)(\xi_{i^*}^{-1} + \xi_i^{-1})V(\bX)}}  \right)\Bigg| \bX \right] \nonumber \\
&= \int_0^\infty \left[\int_0^\infty  \Phi\left( \frac{h}{\sqrt{(n_0m-d)(t^{-1} + s^{-1})V(\bX)}}  \right) \eta(s)\ud s \right]^{|\Omega(\bX)|} \eta(t) \ud t,
\label{eq:lb_pcs2}
\end{align}
where the second inequality follows from \eqref{eq-normalP}, and $|\Omega(\bX)|$ denotes the cardinality of $\Omega(\bX)$.
Since $0\leq  \Phi(\cdot) \leq 1$ and $\eta(\cdot)$ is a pdf, the integral inside the square brackets in \eqref{eq:lb_pcs2} is no greater than 1.
Moreover, since $|\Omega(\bX)|\leq k-1$, hence,
\[\mathrm{PCS}(\bX) \geq  \int_0^\infty \left[\int_0^\infty\Phi\left( \frac{h}{\sqrt{(n_0m-d)(t^{-1} + s^{-1})V(\bX)}}  \right) \eta(s)\ud s \right]^{k-1} \eta(t) \ud t. \]
Then, it follows immediately from the definition of $h$ in \eqref{eq-geth} that $\PCSE=\E[\PCS(\bX)]\ge 1-\alpha$.
\Halmos
\end{proof}

\pdfbookmark[2]{Proof of Theorem \ref{thm-het}}{link-e4}
\hypertarget{EC.4}{
\section*{EC.4. \hspace{5pt} Proof of Theorem \ref{thm-het}}
}

The proof of Theorem \ref{thm-het} critically relies on Lemma \ref{lem-stein2}.

\proof{Proof of Theorem \ref{thm-het}.}

Under Assumption \ref{ass-model2}, for $i=1,\ldots,k,j=1,\ldots,m$, $\overline{Y}_{ij}$ is independent of $ S_{ij}^2$;
moreover, let $\sigma_{ij}= \sigma_i(\bx_j)$, then $\xi_{ij}\coloneqq (n_0-1)S_{ij}^2/\sigma_{ij}^2 \sim \chi_{n_0-1}^2$; see, e.g., Examples 5.6a and 5 in \citet{RencherSchaalje08_ec}.
Let $\mathcal{S}_i \coloneqq  \{S_{i1}^2, \ldots, S_{im}^2 \}$, for $i=1,\ldots,k$.
Then, $\mathcal{S}_i$ is independent of $\sum_{\ell=1}^{n_0} Y_{i\ell}(\bx_j)$ and  of $\{Y_{i\ell}(\bx_j):\ell \geq n_0+1\}$.
Since $N_{i1}, \ldots, N_{im}$  are integer-valued functions only of $\mathcal{S}_i$, by Lemma \ref{lem-stein2}, for $i=1,\ldots,k$,
\[\bX^\intercal \widehat{\BFbeta}_i \Big| \left( \bX, \mathcal{S}_i \right) \;\sim\;  \mathcal N\left(\bX^\intercal \BFbeta_i, \bX^\intercal (\cX^\intercal \cX)^{-1} \cX^\intercal \Sigma_i \cX (\cX^\intercal \cX)^{-1} \bX \right),\]
where $ \Sigma_i = \mathrm{Diag}(\sigma_{i1}^2/N_{i1},\ldots, \sigma_{im}^2/N_{im})$.

For notational simplicity, let $\bm a \coloneqq (a_1,\ldots,a_m)^\intercal \coloneqq \cX (\cX^\intercal \cX)^{-1} \bX$ and write $i^*=i^*(\bX)$ to suppress the dependence on $\bX$. Then,
\begin{equation}\label{eq:normal_het}
\bX^\intercal \widehat{\BFbeta}_i  \Big| \left(\bX, \mathcal{S}_i\right) \;\sim\;  N\left(\bX^\intercal \BFbeta_i,  \sum_{j=1}^m a_j^2 \sigma_{ij}^2 / N_{ij}\right).
\end{equation}
Let $\Omega(\bx) \coloneqq \{i: \bX^\intercal \BFbeta_{i^*} - \bX^\intercal \BFbeta_i  \geq \delta | \bX=\bx\}$ be the set of alternatives outside the IZ given $\bX=\bx$.
For each $i \in \Omega(\bX)$, $\bX^\intercal \widehat{\BFbeta}_{i^*}$ is independent of $\bX^\intercal \widehat{\BFbeta}_i$ given $\bX$. It then follows from \eqref{eq:normal_het} that
\begin{equation} \label{eq-normal2}
\bX^\intercal \widehat{\BFbeta}_{i^*} - \bX^\intercal \widehat{\BFbeta}_i  \Big| \left(\bX, \mathcal{S}_{i^*}, \mathcal{S}_i \right) \; \sim\; \cN\left(\bX^\intercal \BFbeta_{i^*} - \bX^\intercal \BFbeta_i,  \sum_{j=1}^m a_j^2 (\sigma_{i^*j}^2 / N_{i^*j}+\sigma_{ij}^2 / N_{ij}) \right).
\end{equation}
Hence, letting $Z$ denote a standard normal random variable, for each $i\in \Omega(\bX)$, we have
\begin{align}
 \pr \left(\bX^\intercal \widehat{\BFbeta}_{i^*} - \bX^\intercal \widehat{\BFbeta}_i  > 0 \Big| \bX,\mathcal{S}_{i^*}, \mathcal{S}_i \right)
&= \pr \left( Z > \frac{-( \bX^\intercal \BFbeta_{i^*} - \bX^\intercal \BFbeta_i)} { \sqrt{\sum_{j=1}^m a_j^2\left( \sigma_{i^*j}^2 / N_{i^*j}+\sigma_{ij}^2 / N_{ij} \right)} } \Bigg| \bX,\mathcal{S}_{i^*},\mathcal{S}_i \right)\nonumber \\
&\geq \pr \left( Z > \frac{-\delta} { \sqrt{\delta^2 \hhet^{-2} \sum_{j=1}^m a_j^2\left( \sigma_{i^*j}^2 / S_{i^*j}^2 + \sigma_{ij}^2 / S_{ij}^2 \right)} } \Bigg| \bX,\mathcal{S}_{i^*},\mathcal{S}_i \right)\nonumber \\
&= \Phi \left(\frac{\hhet} { \sqrt{ (n_0-1) \sum_{j=1}^m a_j^2 \left( 1/\xi_{i^*j} + 1/\xi_{ij} \right)} } \right), \label{eq-normalP20}
\end{align}
where  the inequality follows the definition of $\Omega(\bX)$ and $N_{ij}$, and the last equality from that of $\xi_{ij}$.

Then, conditionally on $\bX$, by the definition \eqref{eq-PCSX}, the CS event must occur if alternative $i^*$ eliminates all alternatives in $\Omega(\bX)$.
Thus,
\begin{align}\label{eq:lb_pcs_het}
\mathrm{PCS}(\bX) & \geq \pr \left( \bigcap_{i\in\Omega(\bX)} \left\{ \bX^\intercal \widehat{\BFbeta}_{i^*} - \bX^\intercal \widehat{\BFbeta}_i  > 0 \right\} \bigg| \bX \right) \nonumber\\
& = \E \left[ \pr \left( \bigcap_{i\in\Omega(\bX)} \left\{ \bX^\intercal \widehat{\BFbeta}_{i^*} - \bX^\intercal \widehat{\BFbeta}_i  > 0 \right\} \bigg| \bX, \mathcal{S}_{i^*}, \left\{\mathcal{S}_i: i \in \Omega(\bX) \right\} \right) \Bigg| \bX \right],
\end{align}
where the equality is due to the tower law of conditional expectation.
Notice that  conditionally on $\{\bX, \mathcal{S}_{i^*}, \{ \mathcal{S}_i: i\in\Omega(\bX)\} \}$, $\{ \bX^\intercal \widehat{\BFbeta}_{i^*} - \bX^\intercal \widehat{\BFbeta}_i: i\in\Omega(\bX) \}$ is multivariate normal by \eqref{eq-normal2}.
Moreover, for $i,i' \in \Omega(\bX)$ and $i \neq i'$, due to the conditional independence between $\bX^\intercal \widehat{\BFbeta}_i$  and  $\bX^\intercal \widehat{\BFbeta}_{i'}$,
\[\Cov\left(\bX^\intercal \widehat{\BFbeta}_{i^*} - \bX^\intercal \widehat{\BFbeta}_i, \bX^\intercal \widehat{\BFbeta}_{i^*} - \bX^\intercal \widehat{\BFbeta}_{i'} \Big| \bX,\mathcal S_{i^*}, \{ \mathcal{S}_i: i\in\Omega(\bX)\} \right)= \Var\left(\bX^\intercal \widehat{\BFbeta}_{i^*} \Big| \bX, \mathcal S_{i^*}\right)>0.\]
Therefore, applying  \eqref{eq:lb_pcs_het} and Lemma \ref{lem-slepian},
\begin{align}
\mathrm{PCS}(\bX) &\geq \E \left[ \prod_{i\in\Omega(\bX)} \pr \left( \bX^\intercal \widehat{\BFbeta}_{i^*} - \bX^\intercal \widehat{\BFbeta}_i  > 0 \Big| \bX, \mathcal S_{i^*}, \mathcal{S}_i\right)  \Bigg| \bX \right]  \nonumber\\
&\geq \E \left[ \prod_{i\in\Omega(\bX)}  \Phi \left(\frac{\hhet} { \sqrt{ (n_0-1) \sum_{j=1}^m a_j^2 \left( 1/\xi_{i^*j} + 1/\xi_{ij} \right)} } \right) \Bigg| \bX \right], \label{eq:lb_pcs_het2}
\end{align}
where the second inequality follows from \eqref{eq-normalP20}.

Notice that $\xi_{ij}$'s are i.i.d. $\chi_{n_0-1}^2$ random variables. Let  $\xi_i^{(1)} = \min \left\{ \xi_{i1}, \ldots, \xi_{im}\right\}$ be their smallest  order statistic.
Then for each $i \in \Omega(\bX)$,
\begin{align}
\sum_{j=1}^m a_j^2 \left(1/\xi_{i^*j} + 1/\xi_{ij} \right) &\leq \sum_{j=1}^m a_j^2\left(1/\xi_{i^*}^{(1)} + 1/\xi_i^{(1)}\right) = \left(1/\xi_{i^*}^{(1)} + 1/\xi_i^{(1)}\right) \bm a^\intercal \bm a. \label{eq-order}
\end{align}
It then follows from \eqref{eq:lb_pcs_het2} and  \eqref{eq-order} that
\begin{align}
\mathrm{PCS}(\bX) &\geq \E \left[ \prod_{i\in\Omega(\bX)}  \Phi \left(\frac{\hhet} { \sqrt{ (n_0-1) (1/\xi_{i^*}^{(1)} + 1/\xi_i^{(1)} ) \bm a^\intercal \bm a} } \right) \Bigg| \bX \right] \nonumber \\
&= \int_0^\infty \left[ \int_0^\infty \Phi \left( \frac{\hhet}{\sqrt{ (n_0-1) (t^{-1}+s^{-1}) \bm a^\intercal \bm a }}  \right) \gamma_{(1)}(s) \ud s \right]^{|\Omega(\bX)|} \gamma_{(1)}(t) \ud t. \label{eq:lb_pcs2_het}
\end{align}
Since $0\leq \Phi(\cdot) \leq 1$ and $\gamma_{(1)}(\cdot)$ is a pdf, the integral inside the square brackets in \eqref{eq:lb_pcs2_het} is no greater than 1.
Moreover, since $|\Omega(\bX)|\leq k-1$, hence,
\begin{align*}
\mathrm{PCS}(\bX) &\geq \int_0^\infty \left[ \int_0^\infty \Phi \left( \frac{\hhet}{\sqrt{ (n_0-1) (t^{-1}+s^{-1}) \bm a^\intercal \bm a }}  \right) \gamma_{(1)}(s) \ud s \right]^{k-1} \gamma_{(1)}(t) \ud t \\
&= \int_0^\infty \left[ \int_0^\infty \Phi \left( \frac{\hhet}{\sqrt{ (n_0-1) (t^{-1}+s^{-1}) \bX^\intercal (\cX^\intercal \cX)^{-1} \bX }}  \right) \gamma_{(1)}(s) \ud s \right]^{k-1} \gamma_{(1)}(t) \ud t,
\end{align*}
where the equality holds because
\[\bm a^\intercal \bm a = (\cX (\cX^\intercal \cX)^{-1} \bX)^\intercal \cX (\cX^\intercal \cX)^{-1} \bX  = \bX^\intercal (\cX^\intercal \cX)^{-1} \bX. \]
It follows immediately from the definition of $\hhet$ in \eqref{eq-geth2} that $\PCSE=\E[\PCS(\bX)]\ge 1-\alpha$.
\Halmos
\endproof

\begin{remark} \phantomsection \label{remark:h}
We have introduced the smallest order statistics in \eqref{eq-order} for computational feasibility.
Without it, Procedure TS$^+$ would still be valid provided that we can compute the constant $\hhet$ from the following equation,
\begin{equation*}\label{eq-geth3}
\E \left\{ \int_{\mathbb{R}_+^m} \left[  \int_{\mathbb{R}_+^m} g(\bX,\hhet)  \prod_{j=1}^m\gamma(s_j)\ud s_1\cdots \ud s_m\right]^{k-1} \prod_{j=1}^m \gamma(t_j)  \ud t_1 \cdots \ud t_m \right\} = 1-\alpha,
\end{equation*}
where
\[g(\bX,\hhet) \coloneqq \Phi \left( \frac{\hhet}{\sqrt{ (n_0-1) \sum_{j=1}^m a_j^2(t_j^{-1} +s_j^{-1}) }} \right).\]
However, it is prohibitively challenging to solve the above two equations numerically for $m\geq 3$.
By introducing the smallest order statistic, we can instead solve \eqref{eq-geth2} for $\hhet$, which is much easier computationally,
while the price is $\hhet$ will be a little larger then necessary as the lower bound of the $\PCSE$ is further loosened.
Also, in analogy to the discussion in \S\hyperlink{EC.2}{EC.2}, when $\bX$ is high-dimensional, $\hhet$ in \eqref{eq-geth2} can also be solved via Monte Carlo method or stochastic approximation method.
\end{remark}

\pdfbookmark[2]{Proof of Theorem \ref{thm-hom-asy}}{link-e5}
\hypertarget{EC.5}{
\section*{EC.5. \hspace{5pt} Proof of Theorem \ref{thm-hom-asy}}
}

\proof{Proof of Theorem \ref{thm-hom-asy}.}
First notice that, conditionally on $S_i^2$, $N_i \to  \infty$ as $\delta \to  0$.
Recall that $\widehat{\BFbeta}_i = \frac{1}{N_i} (\cX^\intercal \cX)^{-1} \cX^\intercal \sum_{\ell=1}^{N_i} \bY_{i\ell}$,
and $\bY_{i\ell}=(Y_{i\ell}(\bx_1),\ldots,Y_{i\ell}(\bx_m))^\intercal$, $i=1,\ldots,k$.
Under Assumption \ref{ass-model3}, $Y_{i\ell}(\bx)$ is independent of $Y_{i'\ell'}(\bx')$ for any $(i,\ell,\bx)\neq (i',\ell',\bx')$; moreover,  $Y_{i\ell}(\bx)$ and $Y_{i\ell'}(\bx)$ are identically distributed for $\ell=\ell'$.
Recall that $N_i = \max \left\{\lceil h^2 S_i^2/\delta^2\rceil, n_0 \right\}$,
and for small enough $\delta$, $N_i = \lceil h^2 S_i^2/\delta^2\rceil$.
We first establish the following convergence result by the central limit theorem, for each $i=1,\ldots,k$.
\begin{equation}\label{eq-convergence}
\frac{\sqrt{N_i}}{\sigma_i} \left(\frac{1}{N_i}\sum_{\ell=1}^{N_i} Y_{i\ell}(\bx) - \bx^\intercal \BFbeta_i \right) \Bigg| S_i^2 \Rightarrow  Z,
\end{equation}
as $\delta \to  0$, where ``$\Rightarrow $'' denotes convergence in distribution, and $Z$ is a standard normal random variable.
To see \eqref{eq-convergence}, we split the left-hand side of  \eqref{eq-convergence} as follows.
\begin{align}
&\frac{\sqrt{N_i}}{\sigma_i} \left(\frac{1}{N_i}\sum_{\ell=1}^{N_i} Y_{i\ell}(\bx) - \bx^\intercal \BFbeta_i \right) \nonumber\\
= &\frac{\sqrt{N_i}}{\sigma_i} \left\{\frac{n_0}{N_i} \left(\frac{1}{n_0}\sum_{\ell=1}^{n_0} Y_{i\ell}(\bx) - \bx^\intercal \BFbeta_i\right)  + \frac{N_i-n_0}{N_i} \left( \frac{1}{N_i-n_0}\sum_{\ell=n_0+1}^{N_i} Y_{i\ell}(\bx) - \bx^\intercal \BFbeta_i \right) \right\} \nonumber\\
= & \frac{n_0}{\sigma_i \sqrt{N_i}} \left(\frac{1}{n_0}\sum_{\ell=1}^{n_0} Y_{i\ell}(\bx) - \bx^\intercal \BFbeta_i\right)  + \frac{\sqrt{N_i-n_0}}{\sqrt{N_i}} \frac{\sqrt{N_i-n_0}}{\sigma_i} \left( \frac{1}{N_i-n_0}\sum_{\ell=n_0+1}^{N_i} Y_{i\ell}(\bx) - \bx^\intercal \BFbeta_i \right). \label{eq-split}
\end{align}
Conditionally on $S_i^2$, as $\delta \to 0$, $N_i \to \infty$, which implies that
$\frac{n_0}{\sigma_i \sqrt{N_i}} \left(\frac{1}{n_0}\sum_{\ell=1}^{n_0} Y_{i\ell}(\bx) - \bx^\intercal \BFbeta_i\right) \to 0$ almost surely,
$\frac{\sqrt{N_i-n_0}}{\sqrt{N_i}} \to 1$, and
$$\frac{\sqrt{N_i-n_0}}{\sigma_i} \left( \frac{1}{N_i-n_0}\sum_{\ell=n_0+1}^{N_i} Y_{i\ell}(\bx) - \bx^\intercal \BFbeta_i \right) \Rightarrow  Z,$$
given by the central limit theorem.
These three convergence results together with \eqref{eq-split} establish \eqref{eq-convergence}.

It is then easy to see
\[\frac{\sqrt{N_i}}{\sigma_i} \left(\frac{1}{N_i}\sum_{\ell=1}^{N_i} \bY_{i\ell}(\bx) - \cX \BFbeta_i \right) \Big| S_i^2 \Rightarrow  \bZ,\]
as $\delta\to 0$,
where $\bZ \sim \cN(\bm 0, \cI)$ is a  standard $m$-variate normal random vector.
Hence,
\begin{equation}\label{eq-converge-single}
\frac{\sqrt{N_i}}{\sqrt{V(\bX)}} \left(\bX^\intercal \widehat{\BFbeta}_i  - \bX^\intercal \BFbeta_i \right) \Big| \{\bX, S_i^2\} \Rightarrow  \sigma_i Z,
\end{equation}
as $\delta\to 0$, where $V(\bX)\coloneqq \bX^\intercal (\cX^\intercal \cX)^{-1} \bX$.

To simplify notation, we write $i^*=i^*(\BFX)$ to temporarily suppress the dependence on $\BFX$.
Let $\Omega(\bx) \coloneqq \{i: \bX^\intercal \BFbeta_{i^*} - \bX^\intercal \BFbeta_i  \geq \delta | \bX=\bx\}$ be the set of alternatives outside the IZ given $\bX=\bx$.
Let $U_{i^*} \coloneqq \frac{\sqrt{N_{i^*}}}{\sqrt{V(\bX)}} \left(\bX^\intercal \widehat{\BFbeta}_{i^*}  - \bX^\intercal \BFbeta_{i^*} \right)$.
Then, $U_{i^*} | \{\bX, S_{i^*}^2\} \Rightarrow  \sigma_{i^*} Z$, as $\delta\to 0$, by \eqref{eq-converge-single}.
For $i \in \Omega(\bX)$, let
$$U_i \coloneqq \frac{\sqrt{N_{i^*}}}{\sqrt{V(\bX)}} \left(\bX^\intercal \widehat{\BFbeta}_i  - \bX^\intercal \BFbeta_i \right) = \frac{\sqrt{N_{i^*}}}{\sqrt{N_i}} \frac{\sqrt{N_i}}{\sqrt{V(\bX)}} \left(\bX^\intercal \widehat{\BFbeta}_i  - \bX^\intercal \BFbeta_i \right).$$
Then, $U_i | \{\bX, S_{i^*}^2, S_i^2\} \Rightarrow  \frac{S_{i^*}}{S_i}\sigma_i Z$, as $\delta\to 0$, due to \eqref{eq-converge-single} and that $\sqrt{N_{i^*}}/\sqrt{N_i} \to S_{i^*}/S_i$ as $\delta \to 0$.

For notational simplicity, we temporarily let $s$ denote the cardinality of $\Omega(\bX)$, and refer to the $s$ alternatives in $\Omega(\bX)$ as alternatives $1,\ldots,s$, without loss of generality.
As $U_{i^*},U_1,\ldots,U_s$ are independent of each other given $\{\bX, S_{i^*}^2, S_1^2, \ldots, S_k^2\}$,
as $\delta\to 0$,
$$
(U_{i^*},U_1,\ldots,U_s)^\intercal \big| \{\bX, S_{i^*}^2, S_1^2, \ldots, S_s^2\} \Rightarrow \left(\sigma_{i^*} Z_0,\frac{S_{i^*}}{S_1}\sigma_1 Z_1,\ldots,\frac{S_{i^*}}{S_s}\sigma_s Z_s\right)^\intercal,
$$
where $Z_0,Z_1,\ldots,Z_s$ are independent standard normal random variables.
Hence, by the continuous mapping theorem, as $\delta\to 0$,
\begin{equation}\label{eq-asy-normal}
(U_{i^*}-U_1,\ldots,U_{i^*}-U_s)^\intercal \big| \{\bX, S_{i^*}^2, S_1^2, \ldots, S_s^2\} \Rightarrow
\left(\sigma_{i^*} Z_0 - \frac{S_{i^*}}{S_1}\sigma_1 Z_1,\ldots,\sigma_{i^*} Z_0 - \frac{S_{i^*}}{S_s}\sigma_s Z_s\right)^\intercal,
\end{equation}
where the limit is multivariate normal, and for $i,j\in \{1,\ldots,s\}$ and $i\neq j$,
\begin{equation}\label{eq-asy-cov}
\Cov\left(\sigma_{i^*} Z_0 - \frac{S_{i^*}}{S_i}\sigma_i Z_i,\ \sigma_{i^*} Z_0 - \frac{S_{i^*}}{S_j}\sigma_j Z_j \Big| \bX, S_{i^*}^2, S_1^2, \ldots, S_s^2\right)= \sigma_{i^*}^2 >0.
\end{equation}

Now we have
\begin{align}
&\liminf_{\delta\to 0} \mathrm{PCS}(\bX) \nonumber \\
\geq& \liminf_{\delta\to 0} \pr \left( \bigcap_{i \in \Omega(\bX)} \left\{ \bX^\intercal \widehat{\BFbeta}_{i^*} - \bX^\intercal \widehat{\BFbeta}_i  > 0 \right\} \bigg| \bX \right) \label{eq-asy-1}\\
=& \liminf_{\delta\to 0} \E \left[ \pr \left( \bigcap_{i \in \Omega(\bX)} \left\{ \bX^\intercal \widehat{\BFbeta}_{i^*} - \bX^\intercal \widehat{\BFbeta}_i  > 0 \right\} \bigg| \bX, S_{i^*}^2, \left\{S_i^2: i \in \Omega(\bX) \right\} \right) \Bigg| \bX \right] \label{eq-asy-1-1}\\
\geq & \ \E \left[\liminf_{\delta\to 0} \pr \left( \bigcap_{i \in \Omega(\bX)} \left\{\bX^\intercal \widehat{\BFbeta}_{i^*} - \bX^\intercal \widehat{\BFbeta}_i  > 0 \right\} \bigg| \bX, S_{i^*}^2, \left\{S_i^2: i \in \Omega(\bX) \right\} \right) \Bigg| \bX \right] \label{eq-asy-2}\\
= & \ \E \left[\liminf_{\delta\to 0} \pr \left( \bigcap_{i \in \Omega(\bX)} \left\{U_{i^*}- U_i  > \frac{-(\bX^\intercal {\BFbeta}_{i^*} - \bX^\intercal {\BFbeta}_i)}{\sqrt{V(\bX)/N_{i^*}}} \right\} \bigg| \bX, S_{i^*}^2, \left\{S_i^2: i \in \Omega(\bX) \right\} \right) \Bigg| \bX \right] \label{eq-asy-3}\\
= & \ \E \left[ \pr \left( \bigcap_{i \in \Omega(\bX)} \left\{\sigma_{i^*} Z_0 - \frac{S_{i^*}}{S_i}\sigma_i Z_i > \frac{-(\bX^\intercal {\BFbeta}_{i^*} - \bX^\intercal {\BFbeta}_i)}{\sqrt{V(\bX)/N_{i^*}}} \right\} \bigg| \bX, S_{i^*}^2, \left\{S_i^2: i \in \Omega(\bX) \right\} \right) \Bigg| \bX \right] \label{eq-asy-4}\\
\geq & \ \E \left[\prod_{i\in\Omega(\bX)} \pr \left(\sigma_{i^*} Z_0 - \frac{S_{i^*}}{S_i}\sigma_i Z_i > \frac{-(\bX^\intercal {\BFbeta}_{i^*} - \bX^\intercal {\BFbeta}_i)}{\sqrt{V(\bX)/N_{i^*}}} \bigg| \bX, S_{i^*}^2, S_i^2\right)  \Bigg| \bX \right] \label{eq-asy-5}\\
\geq & \ \E \left[\prod_{i\in\Omega(\bX)} \pr \left( \left(\sigma_{i^*}^2 +\frac{S_{i^*}^2}{S_i^2}\sigma_i^2\right)^{1/2}Z  > \frac{ -\delta }{\sqrt{V(\bX)\delta^2/(h^2S_{i^*}^2)}} \bigg| \bX, S_{i^*}^2, S_i^2\right)  \Bigg| \bX \right] \label{eq-asy-6}\\
= & \ \E \left[\prod_{i\in\Omega(\bX)} \pr \left( Z > \frac{-h}{\sqrt{[\sigma_{i^*}^2/S_{i^*}^2 + \sigma_i^2/S_i^2]V(\BFX)}} \right)  \Bigg| \bX \right] \nonumber\\
= & \ \E \left[ \prod_{i\in\Omega(\bX)} \Phi\left( \frac{h}{\sqrt{(n_0m-d)(\xi_{i^*}^{-1} + \xi_i^{-1})V(\bX)}}  \right)\Bigg| \bX \right], \nonumber
\end{align}
where \eqref{eq-asy-1} holds because the CS event must occur if alternative $i^*$ eliminates all alternatives in $\Omega(\bX)$,
\eqref{eq-asy-1-1} is due to the tower law of conditional expectation,
\eqref{eq-asy-2} is due to Fatou's Lemma,
\eqref{eq-asy-3} is by the definitions of $U_{i^*}$ and $U_i$,
\eqref{eq-asy-4} is by \eqref{eq-asy-normal},
\eqref{eq-asy-5} is obtained by Lemma \ref{lem-slepian} together with \eqref{eq-asy-cov},
and \eqref{eq-asy-6} follows from the definitions of $\Omega(\bX)$ and $N_{i^*}$.

The rest of the proof follows the same argument as that in the proof of Theorem \ref{thm-hom}.
\Halmos
\endproof

\pdfbookmark[2]{Choosing PCSmin as Target}{link-e5-2}
\hypertarget{EC.6}{
\section*{EC.6. \hspace{5pt} Choosing $\PCSmin$ as Target}
}

\vspace{10pt}
\subsection*{EC.6.1. \hspace{5pt} Two-Stage Procedures}
If we use $\PCSmin = \min_{\bx\in\Theta} \PCS(\bx)$ to measure correct selection across the population, and set the pre-specified target as $\PCSmin \geq 1-\alpha$, instead of $\PCSE \geq 1-\alpha$, we are in a more conservative case
wherein we require  the selection policy produced by the selection procedure to make correct selection with probability at least $1-\alpha$ for \emph{all} values of the covariates.
In this case, both Procedure TS and Procedure TS$^+$ can be revised slightly to retain statistical validity under the new criterion.
In particular, we only need change the definition of the constant $h$ (resp., $\hhet$) in Procedure TS (resp., Procedure TS$^+$), while keeping the other parts of the procedure the same.
The following results are parallel to those for $\PCSE$, that is, Theorems \ref{thm-hom}--\ref{thm-het-asy}.
The proofs are essentially the same and thus we omit the details.

\begin{theorem}
Suppose that Procedure TS is used to solve the R\&S-C problem with the constant $h$ in the procedure being solved from
\begin{equation} \label{eq-geth-min}
\min_{\bx\in\Theta} \left\{ \int_0^\infty \left[\int_0^\infty \Phi \left( \frac{h}{\sqrt{(n_0m-d) (t^{-1}+s^{-1})\bx^\intercal (\cX^\intercal \cX)^{-1} \bx}} \right) \eta(s)\ud s \right]^{k-1} \eta(t) \ud t \right\} = 1-\alpha.
\end{equation}
\begin{itemize}
    \item If Assumption \ref{ass-model} is satisfied, then $\PCSmin \geq 1-\alpha$.
    \item If Assumption \ref{ass-model3} is satisfied, then $\liminf_{\delta \to  0} \PCSmin \geq 1-\alpha$.
\end{itemize}
\end{theorem}

\begin{theorem}
Suppose that Procedure TS$^+$ is used to solve the R\&S-C problem with the constant $\hhet$ in the procedure being solved from
\begin{equation} \label{eq-geth2-min}
\min_{\bx\in\Theta} \left\{ \int_0^\infty \left[ \int_0^\infty \Phi \left(  \frac{\hhet}{\sqrt{ (n_0-1) (t^{-1}+s^{-1})\bx^\intercal (\cX^\intercal \cX)^{-1} \bx }} \right) \gamma_{(1)}(s) \ud s \right]^{k-1} \gamma_{(1)}(t) \ud t \right\} = 1-\alpha.
\end{equation}
\begin{itemize}
    \item If Assumption \ref{ass-model2} is satisfied, then $\PCSmin \geq 1-\alpha$.
    \item If Assumption \ref{ass-model4} is satisfied, then $\liminf_{\delta \to  0} \PCSmin \geq 1-\alpha$.
\end{itemize}
\end{theorem}

\begin{remark}
It is computationally easier to solve for $h$  from \eqref{eq-geth-min}  than from \eqref{eq-geth}.
First, there is no need to compute the expectation with respect to the distribution of $\BFx$  in \eqref{eq-geth-min}, which amounts to
multidimensional numerical integration.
Second, note that the minimizer of the left-hand side of \eqref{eq-geth-min} is the same as the maximizer of $\bx^\intercal (\cX^\intercal \cX)^{-1} \bx$, since the function $\Phi(\cdot)$ is increasing.
Since $\cX^\intercal \cX$ is nonsingular, it is easy to see that $\bx^\intercal (\cX^\intercal \cX)^{-1} \bx$ is convex in $\bx$.
Thus, if $\Theta$ is a bounded closed set, the maximizer must lie in the set of all extreme points of the convex hull of $\Theta$; see, for example, Theorem 32.2 and Corollary 32.3.3 in \cite{rockafellar1970_ec}.
A similar argument can be made for the computation of $\hhet$ by comparing \eqref{eq-geth2-min} with \eqref{eq-geth2}.
\end{remark}

\vspace{5pt}
\subsection*{EC.6.2. \hspace{5pt} Numerical Results}

Define the achieved $\PCSmin$ as
\[\APCSmin \coloneqq  \min_{\bx\in\{\bx_1,\ldots,\bx_T\}} \frac{1}{R} \sum_{r=1}^R \mathbb{I} \left\{ \mu_{i^*(\bx)}(\bx) - \mu_{\widehat{i^*_r}(\bx)}(\bx)  < \delta \right\}.\]
Table \ref{tab-PCSmin} reports $\APCSE$ and $\APCSmin$ when the target is $\PCSmin \geq 95\%$, while Table \ref{tab-PCSE-2} reports the case when the target is $\PCSE \geq 95\%$.

\begin{table}[!b]
\centering
{
\small
\caption{Results When the Target is $\PCSmin \geq 95\%$.} \label{tab-PCSmin}
    \begin{tabular}{lcccccccccccc}
    \toprule
     & & &  \multicolumn{4}{c}{Procedure TS (using $h$ in \eqref{eq-geth-min})} & & & \multicolumn{4}{c}{Procedure TS$^+$ (using $h$ in \eqref{eq-geth2-min})} \\
     \cmidrule(lr){4-7} \cmidrule(lr){10-13}
     Problem & & &  $h$ & Sample &  $\APCSE$ & $\APCSmin$ & & & $\hhet$ & Sample &  $\APCSE$ & $\APCSmin$\\
    \midrule
    \texttt{(0)} Benchmark & & & 5.927 & 140,543 & 0.9989 & 0.9609 & & & 6.990 & 195,337 & 0.9997 & 0.9840 \\
    \texttt{(1)} $k=2$ & & & 4.362 & \phantom{1}30,447 & 0.9958 & 0.9481 & & & 5.132 & \phantom{1}42,164 & 0.9987 & 0.9709 \\
    \texttt{(2)} $k=8$ & & & 6.481 & 268,749 & 0.9993 & 0.9657 & & & 7.651 & 374,716 & 0.9999 & 0.9852 \\
    \texttt{(3)} Non-GSC & & & 5.927 & 140,542 & 1.0000 & 0.9958 & & & 6.990 & 195,337 & 1.0000 & 0.9980 \\
    \texttt{(4)} IV & & & 5.927 & 158,139 & 0.9989 & 0.9590 & & & 6.990 & 219,871 & 0.9998 & 0.9869 \\
    \texttt{(5)} DV & & & 5.927 & 158,100 & 0.9990 & 0.9628 & & & 6.990 & 219,741 & 0.9998 & 0.9837 \\
    \texttt{(6)} Het & & & 5.927 & 175,698 & 0.9952 & \framebox{0.9032} & & & 6.990 & 244,488 & 0.9999 & \textBF{0.9904} \\
    \texttt{(7)} $d=2$ & & & 7.155 & \phantom{1}51,161 & 0.9954 & 0.9600 & & & 7.648 & \phantom{1}58,493 & 0.9971 & 0.9708 \\
    \texttt{(8)} $d=6$ & & & 3.792 & 230,221 & 0.9994 & 0.9667 & & & 4.804 & 369,307 & 1.0000 & 0.9944 \\

    \texttt{(9)} Normal Dist & &     & 5.927 & 140,550 & 0.9990 & 0.9623 & &    & 6.990 & 195,404           & 0.9997 & 0.9851\\

    \cmidrule{1-13}

    \texttt{(10)} $k=100$ & &     & \phantom{1}7.385 & 3,272,127 & 0.9999 & 0.9754 & &      & \phantom{1}8.678 & 4,518,029        & 1.0000 & 0.9941 \\
    \texttt{(11)} $d=50$ & &      & \phantom{1}9.444 & 4,370,569 & 1.0000 & 0.9998 & &      & 12.631 & 7,818,201        & 1.0000 & 1.0000\\
    \texttt{(12)} $k=100$, $d=50$ & &     & 13.875 & $1.89\times10^8$ & 1.0000 & 1.0000 & &            & 18.970 & $3.53\times10^8$           & 1.0000 & 1.0000\\

    \bottomrule
    \end{tabular}
\begin{minipage}[t]{1\linewidth}
\SingleSpacedXI
\vspace{0.6em}
\footnotesize{
\emph{Note.} \textsf{In the presence of heteroscedasticity, the boxed number suggests that Procedure TS fails to deliver the target $\PCSmin$, whereas the bold number suggests that Procedure TS$^+$ succeeds to do so.}
}
\end{minipage}
}
\end{table}

\begin{table}[!b]
\centering
{
\small
\caption{Results When the Target is $\PCSE \geq 95\%$.} \label{tab-PCSE-2}
    \begin{tabular}{lcccccccccccc}
    \toprule
     & & &  \multicolumn{4}{c}{Procedure TS (using $h$ in \eqref{eq-geth})} & & & \multicolumn{4}{c}{Procedure TS$^+$ (using $h$ in \eqref{eq-geth2})} \\
     \cmidrule(lr){4-7} \cmidrule(lr){10-13}
     Problem & & &  $h$ & Sample &  $\APCSE$ & $\APCSmin$ & & & $\hhet$ & Sample &  $\APCSE$ & $\APCSmin$\\
    \midrule
    \texttt{(0)} Benchmark & & & 3.423 & 46,865 & 0.9610 & 0.7476 & & & 4.034 & \phantom{1}65,138 & 0.9801 & 0.8120 \\

    \texttt{(1)} $k=2$ & & & 2.363 & \phantom{1}8,947 & 0.9501 & 0.8094 & & & 2.781 & \phantom{1}12,380 & 0.9702 & 0.8541 \\
    \texttt{(2)} $k=8$ & & & 3.822 & 93,542 & 0.9650 & 0.7290 & & & 4.510 & 130,200 & 0.9842 & 0.8098 \\
    \texttt{(3)} Non-GSC & & & 3.423 & 46,865 & 0.9987 & 0.9400 & & & 4.034 & \phantom{1}65,138 & 0.9994 & 0.9599 \\
    \texttt{(4)} IV & & & 3.423 & 52,698 & 0.9618 & 0.7589 & & & 4.034 & \phantom{1}73,265 & 0.9807 & 0.8184 \\
    \texttt{(5)} DV & & & 3.423 & 52,720 & 0.9614 & 0.7544 & & & 4.034 & \phantom{1}73,246 & 0.9806 & 0.8143 \\
    \texttt{(6)} Het & & & 3.423 & 58,626 & \framebox{0.9232} & 0.6368 & & & 4.034 & \phantom{1}81,555 & \textBF{0.9846} & 0.8625 \\
    \texttt{(7)} $d=2$ & & & 4.612 & 21,288 & 0.9593 & 0.7941 & & & 4.924 & \phantom{1}24,266 & 0.9662 & 0.8223 \\
    \texttt{(8)} $d=6$ & & & 2.141 & 73,428 & 0.9656 & 0.7662 & & & 2.710 & 117,626 & 0.9895 & 0.8589 \\
    \texttt{(9)} Normal Dist & &     & 3.447 & 47,529 & 0.9626 & 0.7579 & &            & 4.063 & \phantom{1}66,061           & 0.9821 & 0.8230\\

    \cmidrule{1-13}

    \texttt{(10)} $k=100$ & &             & 4.346 & \phantom{0}1,133,384 & 0.9758 & 0.5952 & &            & 5.117 & \phantom{0}1,570,911           & 0.9918 & 0.7218 \\
    \texttt{(11)} $d=50$ & &              & 3.222 & \phantom{00,}508,977 & 0.9583 & 0.7522 & &            & 4.312 & \phantom{00,}911,326           & 0.9926 & 0.8749\\
    \texttt{(12)} $k=100$, $d=50$ & &     & 4.886 & 23,400,677 & 0.9765 & 0.6189 & &            & 6.702 & 44,024,486           & 0.9991 & 0.8854\\
    \bottomrule
    \end{tabular}
\begin{minipage}[t]{1\linewidth}
\SingleSpacedXI
\vspace{0.6em}
\footnotesize{
\emph{Note.} \textsf{In the presence of heteroscedasticity, the boxed number suggests that Procedure TS fails to deliver the target $\PCSE$, whereas the bold number suggests that Procedure TS$^+$ succeeds to do so.}
}
\end{minipage}
}
\end{table}%

First, results in Table \ref{tab-PCSmin} show that Procedure TS and Procedure TS$^+$ with $h$ and $\hhet$ computed from \eqref{eq-geth-min} and \eqref{eq-geth2-min}, respectively,
can deliver the target $\PCSmin$ in their respective domains.
In particular,
Procedure TS using $h$ in \eqref{eq-geth-min} can deliver the target $\PCSmin$ if the simulation errors are homoscedastic,
while Procedure TS$^+$ using $h$ in \eqref{eq-geth2-min} can do the same even when the simulation errors are heteroscedastic.
Moreover, the achieved $\PCSmin$ is higher than the target in general; see, e.g., the column ``$\APCSmin$'' under ``Procedure TS'' of Table \ref{tab-PCSmin}, except the entry for Problem \texttt{(6)}.
This kind of conservativeness is also observed in Table \ref{tab-PCSE-2},
where Procedure TS using $h$ in \eqref{eq-geth} and Procedure TS$^+$ using $h$ in \eqref{eq-geth2} are used when the objective is to meet the target $\PCSE$.

Second, the numerical results show that $\PCSmin$ is a much more conservative criterion than $\PCSE$.
In particular, if the target is $\PCSE\geq 1-\alpha$, then $\APCSmin$ is significantly lower than $1-\alpha$, except for Problem \texttt{(3)}, in which the non-GSC amplifies the procedures' conservativeness stemming from the IZ formulation and provides the ``extra'' sample size needed for making $\APCSmin$ reach the target; see Table \ref{tab-PCSE-2}.
By contrast, if the target is $\PCSmin\geq 1-\alpha$, then $\APCSE$ is virtually 1 for each problem-procedure combination; see Table \ref{tab-PCSmin}.
Another indication of the conservativeness of $\PCSmin$ is that in each problem-procedure combination, the sample size when using $\PCSmin$ as the criterion is about three times larger than that when using $\PCSE$.
For example, in Table \ref{tab-PCSE-2} the sample size for Problem \texttt{(0)} with Procedure TS is 46,865, whereas the corresponding sample size in Table \ref{tab-PCSmin} is 140,543.

\pdfbookmark[2]{Asymptotic Sample Size Analysis}{link-e5-3}
\hypertarget{EC.7}{
\section*{EC.7. \hspace{5pt} Asymptotic Sample Size Analysis}
}

For ease of presentation, we relax the integrality constraint of the sample size, but it has no essential impact on the asymptotic analysis of the sample size.

\vspace{5pt}
\subsection*{EC.7.1. \hspace{5pt} Procedure TS}
The expected total sample size of Procedure TS is
\begin{equation}\label{expected-sample-size-TS}
    N_{\text{TS}} = \E \bigg[\sum_{i=1}^k mN_i \bigg] = m h^2 \sum_{i=1}^k \E \big[\max \left\{ S_i^2/\delta^2, n_0 / h^2 \right\}\big],
\end{equation}
where $h$ is solved from \eqref{eq-geth} if $\PCSE$ is used as the criterion, whereas from \eqref{eq-geth-min} if $\PCSmin$ is used.
We will provide an asymptotic upper bound on $N_{\text{TS}}$ in two asymptotic regimes, i.e., as $k\to\infty$ and as $\alpha\to 0$.
Note that the expression of $N_{\text{TS}}$ involves both $h^2$ and $1/h^2$, which depend on $k$ and $\alpha$.
We first establish in Lemma \ref{lem-h-bound} both a lower bound and a upper bound on $h$.
Its proof is deferred to \S\hyperlink{EC.7.3}{EC.7.3}.

\begin{lemma} \phantomsection \label{lem-h-bound}
Let $h$ be the constant solved from either \eqref{eq-geth} or \eqref{eq-geth-min}, and let $\underline{\alpha}\in(0, 1/2)$ be a constant.
Then,
$$0 < \underline{h} \leq h \leq \left\{2(n_0m-d)\Big[\Big(\frac{2(k-1)}{\alpha}\Big)^{\frac{2}{n_0m-d}}-1\Big] \times \max_{\bx\in\Theta} \bx^\intercal (\cX^\intercal \cX)^{-1} \bx \right\}^{1/2},$$
for all $k\geq 2$ and $\alpha\leq \underline{\alpha}$,
where $\underline{h}$ is a solution of \eqref{eq-geth} for $k=2$ and $\alpha=\underline{\alpha}$.
\end{lemma}

Then, it follows immediately from \eqref{expected-sample-size-TS} and  Lemma \ref{lem-h-bound} that for any $k$ and small $\alpha$,
\begin{align*}
N_{\text{TS}} \leq{}& 2m(n_0m-d)\Big[\Big(\frac{2(k-1)}{\alpha}\Big)^{\frac{2}{n_0m-d}}-1\Big] \times \max_{\bx\in\Theta} \bx^\intercal (\cX^\intercal \cX)^{-1} \bx \times \sum_{i=1}^k \E \big[\max \left\{ S_i^2/\delta^2, n_0 / \underline{h}^2 \right\}\big]\\
\leq{}& C_{\text{TS}} \times k\Big(\frac{k}{\alpha}\Big)^{\frac{2}{n_0m-d}},
\end{align*}
where $C_{\text{TS}}$ is a constant independent of $k$ and $\alpha$, given by
\[
C_{\text{TS}} = 2m(n_0m-d)\times 2^{\frac{2}{n_0m-d}} \times \max_{\bx\in\Theta} \bx^\intercal (\cX^\intercal \cX)^{-1} \bx \times \max_{1\leq i\leq k} \E \big[\max \left\{ S_i^2/\delta^2, n_0 / \underline{h}^2 \right\}\big].
\]
Hence, we conclude that $N_{\text{TS}} = \mathcal{O}(k^{1+\frac{2}{n_0m-d}})$ as $k\to\infty$, and $N_{\text{TS}} = \mathcal{O}(\alpha^{-\frac{2}{n_0m-d}})$ as $\alpha\to 0$.

\vspace{5pt}
\subsection*{EC.7.2. \hspace{5pt} Procedure TS$^+$}
The expected total sample size of Procedure TS$^+$ is
\begin{equation}\label{expected-sample-size-TS+}
    N_{\text{TS}^+} = \E \bigg[\sum_{i=1}^k \sum_{j=1}^m N_{ij} \bigg] = \hhet^2 \sum_{i=1}^k \sum_{j=1}^m \E \big[ \max \left\{ S_{ij}^2/\delta^2, n_0/\hhet^2 \right\} \big],
\end{equation}
where $\hhet$ is solved from \eqref{eq-geth2} if $\PCSE$ is used as the criterion, whereas from \eqref{eq-geth2-min} if $\PCSmin$ is used.
Similar to the analysis in \S EC.7.1, we first give bounds on $\hhet$ in Lemma \ref{lem-hhet-bound}. The proof is presented in \S\hyperlink{EC.7.4}{EC.7.4}.

\begin{lemma} \phantomsection \label{lem-hhet-bound}
Let $\hhet$ be the constant solved either from \eqref{eq-geth2} or \eqref{eq-geth2-min}, and let $\underline{\alpha}\in(0,1/2)$ be a constant.
Then,
$$0 < \underline{\hhet} \leq \hhet \leq \left\{2(n_0-1)\Big[\Big(\frac{2m(k-1)}{\alpha}\Big)^{\frac{2}{n_0-1}}-1\Big] \times \max_{\bx\in\Theta} \bx^\intercal (\cX^\intercal \cX)^{-1} \bx \right\}^{1/2},$$
for all $k\geq 2$ and $\alpha\leq \underline{\alpha}$,
where $\underline{\hhet}$ is a solution of \eqref{eq-geth2} when $k=2$ and $\alpha=\underline{\alpha}$.
\end{lemma}

Then, it follows from \eqref{expected-sample-size-TS} and Lemma \ref{lem-hhet-bound} that
for any $k$ and small $\alpha$,
\begin{align*}
N_{\text{TS}^+}  \leq{}& 2(n_0-1)\Big[\Big(\frac{2m(k-1)}{\alpha}\Big)^{\frac{2}{n_0-1}}-1\Big] \times \max_{\bx\in\Theta} \bx^\intercal (\cX^\intercal \cX)^{-1} \bx \times \sum_{i=1}^k \sum_{j=1}^m \E \big[ \max \left\{ S_{ij}^2/\delta^2, n_0/\underline{\hhet}^2 \right\} \big], \\
\leq{}& C_{\text{TS}^+} \times k\Big(\frac{k}{\alpha}\Big)^{\frac{2}{n_0-1}},
\end{align*}
where $C_{\text{TS}^+}$ is a constant independent of $k$ and $\alpha$, given by
\[
C_{\text{TS}^+} = 2(n_0-1)\times (2m)^{\frac{2}{n_0-1}} \times \max_{\bx\in\Theta} \bx^\intercal (\cX^\intercal \cX)^{-1} \bx \times\max_{1\leq i\leq k} \sum_{j=1}^m \E \big[ \max \left\{ S_{ij}^2/\delta^2, n_0/\underline{\hhet}^2 \right\} \big].
\]
Hence, we conclude that $N_{\text{TS}^+} = \mathcal{O}(k^{1+\frac{2}{n_0-1}})$ as $k\to\infty$, and $N_{\text{TS}} = \mathcal{O}(\alpha^{-\frac{2}{n_0-1}})$ as $\alpha\to 0$.

\begin{remark}
It is straightforward to see that with the same design matrix $\cX$ and  initial sample size $n_0$,
$N_{\text{TS}^+}=\mathcal{O}(k^{1+\frac{2}{n_0-1}}\alpha^{-\frac{2}{n_0-1}})$ has a higher order of magnitude than $N_{\text{TS}}=\mathcal{O}(k^{1+\frac{2}{n_0m-d}}\alpha^{-\frac{2}{n_0m-d}})$ as $k\to\infty$ or $\alpha\to 0$, since $n_0m-d\geq n_0-1$ for all $m\geq d \geq 1$.
\end{remark}

\begin{remark}
For Procedure TS$^+$, $m$ is not involved in the order of magnitude of $N_{\text{TS}^+}$ as $k\to\infty$ or $\alpha\to 0$, but only takes effect in the leading constant $C_{\text{TS}^+}$.
By contrast,
a larger value of $m$ leads to a lower order of magnitude of $N_{\text{TS}}$ for Procedure TS.
An intuitive explanation for the above difference is that, increasing $m$ will result in a more accurate estimation of the common variance $\sigma_i^2$ in Procedure TS, while it does not affect estimation of the variances in Procedure TS$^+$ since they are estimated separately.
This suggests that Procedure TS$^+$ for the linear models will favor the minimal $m$, that is, $m=d$.
\end{remark}

\vspace{5pt}
\hypertarget{EC.7.3}{
\subsection*{EC.7.3. \hspace{5pt} Proof of Lemma \ref{lem-h-bound}}
}

We first prove the lower bound.
Let
$$f(\bx,h) \coloneqq \int_0^\infty \left[\int_0^\infty \Phi \left( \frac{h}{\sqrt{(n_0m-d) (t^{-1}+s^{-1})\bx^\intercal (\cX^\intercal \cX)^{-1} \bx}} \right) \eta(s)\ud s \right]^{k-1} \eta(t) \ud t.$$
Then, $h$ solved from \eqref{eq-geth} satisfies $\E[f(\bX,h)]=1-\alpha$, whereas $h$ solved from \eqref{eq-geth-min} satisfies $\min_{\bx\in\Theta} f(\bx,h)=1-\alpha$.
Note that both $\E[f(\bX,h)]$ and $\min_{\bx\in\Theta} f(\bx,h)$ are increasing functions in $h$, so
a smaller $\alpha$ will yield a larger $h$.
It is also clear that a larger $k$ will yield a larger $h$.
Hence, $\underline{h}$ defined in Lemma \ref{lem-h-bound} is smaller than $h$ solved from \eqref{eq-geth} for all $k\geq 2$ and $\alpha\leq \underline{\alpha}$.
Note that $h$ solved from \eqref{eq-geth} is smaller than $h$ solved from \eqref{eq-geth-min} with everything else the same.
The lower bound is then established.

The proof for the upper bound is similar to the proof of Lemma 4 in \cite{zhong2020_ec}.
Specifically, let
\[
h^* \coloneqq \left\{2(n_0m-d)\Big[\big(\frac{2(k-1)}{\alpha}\big)^{\frac{2}{n_0m-d}}-1\Big] \times \max_{\bx\in\Theta} \bx^\intercal (\cX^\intercal \cX)^{-1} \bx \right\}^{1/2}.
\]
To show that $h$, which is  solved from either \eqref{eq-geth} or \eqref{eq-geth-min}, is no larger than $h^*$, it suffices to show that
$\E[f(\bX,h^*)]\geq1-\alpha$ and $\min_{\bx\in\Theta} f(\bx,h^*)\geq1-\alpha$, which is clearly true if we can show that $f(\bx,h^*)\geq1-\alpha$ for any $\bx\in\Theta$.

Let $Z_1,\ldots,Z_{k-1}$ be $(k-1)$ independent standard normal random variables. Let $\xi_1,\ldots,\xi_k$ be $k$  independent chi-square random variables with $(n_0m-d)$ degrees of freedom.
Moreover, assume that $Z_i$ is independent of $\xi_{i'}$, for $1\leq i \leq k-1$, $1\leq i' \leq k$.
Then, for any $\bx\in\Theta$,
\begin{align}
f(\bx,h^*) &= \E \left[ \pr \left( \bigcap_{i=1}^{k-1} \left\{Z_i \leq \frac{h^*}{\sqrt{(n_0m-d) (\xi_k^{-1}+\xi_i^{-1})\bx^\intercal (\cX^\intercal \cX)^{-1} \bx}} \right\} \Bigg| \xi_k \right) \right] \nonumber\\
&\geq \E \left[ 1 - \sum_{i=1}^{k-1} \pr \left(Z_i > \frac{h^*}{\sqrt{(n_0m-d) (\xi_k^{-1}+\xi_i^{-1})\bx^\intercal (\cX^\intercal \cX)^{-1} \bx}} \Bigg| \xi_k \right) \right] \nonumber\\
&= 1 - (k-1) \times \pr \left(Z_1 > \frac{h^*}{\sqrt{(n_0m-d) (\xi_k^{-1}+\xi_1^{-1})\bx^\intercal (\cX^\intercal \cX)^{-1} \bx}} \right). \label{eq:f_x_h}
\end{align}
By the Chernoff bound, $\pr(Z_1>a)\leq \exp\big\{-\frac{a^2}{2}\big\}$ for all $a>0$.
Hence,
\begin{align}
&\pr \left(Z_1 > \frac{h^*}{\sqrt{(n_0m-d) (\xi_k^{-1}+\xi_1^{-1})\bx^\intercal (\cX^\intercal \cX)^{-1} \bx}} \right)\nonumber\\
\leq{}& \E \left[\exp\left\{-\frac{(h^*)^2}{2(n_0m-d) (\xi_k^{-1}+\xi_1^{-1})\bx^\intercal (\cX^\intercal \cX)^{-1} \bx}\right\} \right] \nonumber\\
\leq{}& \E \Bigg[\exp\Bigg\{-\frac{\big(\frac{2(k-1)}{\alpha}\big)^{\frac{2}{n_0m-d}}-1}{\xi_k^{-1}+\xi_1^{-1}}\Bigg\} \Bigg] \nonumber\\
\leq{}& \E \Bigg[\exp\Bigg\{-\frac{\big(\frac{2(k-1)}{\alpha}\big)^{\frac{2}{n_0m-d}}-1}{2} \xi_{(1)}\Bigg\} \Bigg] \label{eq:pr_Z_h},
\end{align}
where $\xi_{(1)} \coloneqq \min\{\xi_k,\xi_1\}$ is the smallest order statistic of two independent chi-square random variables with $(n_0m-d)$ degrees of freedom.
Here,
the second inequality holds by the definition of $h^*$.
Let $f_{n_0m-d}(\cdot)$ and $F_{n_0m-d}(\cdot)$ denote the pdf and cdf of the chi-square random variables with $(n_0m-d)$ degrees of freedom, respectively.
Then the pdf of $\xi_{(1)}$ is known as $2f_{n_0m-d}(t)(1-F_{n_0m-d}(t))$.
Hence, following \eqref{eq:pr_Z_h}, we have
\begin{align}
&\pr \left(Z_1 > \frac{h^*}{\sqrt{(n_0m-d) (\xi_k^{-1}+\xi_1^{-1})\bx^\intercal (\cX^\intercal \cX)^{-1} \bx}} \right) \nonumber\\
\leq & \int_0^\infty \exp\Bigg\{-\frac{\big(\frac{2(k-1)}{\alpha}\big)^{\frac{2}{n_0m-d}}-1}{2} t\Bigg\} \times 2f_{n_0m-d}(t)(1-F_{n_0m-d}(t)) \ud t \nonumber\\
\leq & \ 2 \int_0^\infty \exp\Bigg\{-\frac{\big(\frac{2(k-1)}{\alpha}\big)^{\frac{2}{n_0m-d}}-1}{2} t\Bigg\} f_{n_0m-d}(t) \ud t \nonumber\\
= & \ 2 \E \Bigg[\exp\Bigg\{-\frac{\big(\frac{2(k-1)}{\alpha}\big)^{\frac{2}{n_0m-d}}-1}{2} \xi_1 \Bigg\} \Bigg] \nonumber\\
= & \ 2 \Big[1+\big(\frac{2(k-1)}{\alpha}\big)^{\frac{2}{n_0m-d}}-1\Big]^{-(n_0m-d)/2}\nonumber\\
=& \alpha/(k-1), \label{eq:pr_Z_h_2}
\end{align}
where the second equality is due to the moment generating function of the chi-square random variables with $(n_0m-d)$ degrees of freedom.
Combining \eqref{eq:f_x_h} and \eqref{eq:pr_Z_h_2}, we can conclude that $f(\bx,h^*)\geq1-\alpha$ for any $\bx\in\Theta$, which completes the proof.
\Halmos

\vspace{5pt}
\hypertarget{EC.7.4}{
\subsection*{EC.7.4. \hspace{5pt} Proof of Lemma \ref{lem-hhet-bound}}
}

The lower bound can be proved using the same argument for proving the lower bound in Lemma \ref{lem-h-bound}.
Let
\[\hhet^* \coloneqq \left\{2(n_0-1)\Big[\big(\frac{2m(k-1)}{\alpha}\big)^{\frac{2}{n_0-1}}-1\Big] \times \max_{\bx\in\Theta} \bx^\intercal (\cX^\intercal \cX)^{-1} \bx \right\}^{1/2}, \]
and let
$$f(\bx,\hhet) \coloneqq \int_0^\infty \left[\int_0^\infty \Phi \left( \frac{\hhet}{\sqrt{(n_0-1) (t^{-1}+s^{-1})\bx^\intercal (\cX^\intercal \cX)^{-1} \bx}} \right) \gamma_{(1)}(s)\ud s \right]^{k-1} \gamma_{(1)}(t) \ud t.$$
It then suffices to show that $f(\bx,\hhet^*)\geq1-\alpha$ for any $\bx\in\Theta$, in order to prove the upper bound.

Let $Z_1,\ldots,Z_{k-1}$ be $(k-1)$ independent standard normal random variables.
Let $\xi_1,\ldots,\xi_k$ be $k$ independent random variables, each of which is the smallest order statistic of $m$ chi-square random variables with $(n_0-1)$ degrees of freedom.
Moreover, assume that $Z_i$ is independent of $\xi_{i'}$, for $1\leq i \leq k-1$, $1\leq i' \leq k$.
With the same argument leading to \eqref{eq:f_x_h}, we have
\begin{equation}\label{eq:f_x_hhet}
f(\bx,\hhet^*) \geq 1 - (k-1) \times \pr \left(Z_1 > \frac{\hhet^*}{\sqrt{(n_0-1) (\xi_k^{-1}+\xi_1^{-1})\bx^\intercal (\cX^\intercal \cX)^{-1} \bx}} \right).
\end{equation}
With the same argument leading to \eqref{eq:pr_Z_h}, we have
\begin{equation*} \label{eq:pr_Z_hhet}
\pr \left(Z_1 > \frac{\hhet^*}{\sqrt{(n_0-1) (\xi_k^{-1}+\xi_1^{-1})\bx^\intercal (\cX^\intercal \cX)^{-1} \bx}} \right) \leq \E \Bigg[\exp\Bigg\{-\frac{\big(\frac{2m(k-1)}{\alpha}\big)^{\frac{2}{n_0-1}}-1}{2} \xi_{(1)}\Bigg\} \Bigg],
\end{equation*}
where $\xi_{(1)} \coloneqq \min\{\xi_k,\xi_1\}$ is the smallest order statistic of $2m$ independent chi-square random variables with $(n_0-1)$ degrees of freedom, and its pdf is $2m f_{n_0-1}(t)(1-F_{n_0-1}(t))^{2m-1}$.
Then, with the same argument leading to \eqref{eq:pr_Z_h_2}, we have
\begin{align}
\pr \left(Z_1 > \frac{\hhet^*}{\sqrt{(n_0-1) (\xi_k^{-1}+\xi_1^{-1})\bx^\intercal (\cX^\intercal \cX)^{-1} \bx}} \right)
&\leq 2m \int_0^\infty \exp\Bigg\{-\frac{\big(\frac{2m(k-1)}{\alpha}\big)^{\frac{2}{n_0-1}}-1}{2} t\Bigg\} f_{n_0-1}(t) \ud t \nonumber\\
&= 2m \E \Bigg[\exp\Bigg\{-\frac{\big(\frac{2m(k-1)}{\alpha}\big)^{\frac{2}{n_0-1}}-1}{2} \xi_0 \Bigg\} \Bigg] \nonumber\\
&= \alpha/(k-1), \label{eq:pr_Z_hhet_2}
\end{align}
where $\xi_0$ is a chi-square random variables with $(n_0-1)$ degrees of freedom.
Combining \eqref{eq:f_x_hhet} and \eqref{eq:pr_Z_hhet_2}, we can conclude that $f(\bx,\hhet^*)\geq1-\alpha$ for any $\bx\in\Theta$, which completes the proof.\Halmos

\pdfbookmark[2]{Proof of Theorem \ref{thm-GSC}}{link-e6}
\hypertarget{EC.8}{
\section*{EC.8. \hspace{5pt} Proof of Theorem \ref{thm-GSC}}
}

Theorem \ref{thm-GSC} can be viewed as a corollary of the following Theorem \ref{thm-GSC2}.
Therefore we only provide the proof of Theorem \ref{thm-GSC2} and remark that Theorem \ref{thm-GSC} holds immediately.

\begin{theorem} \phantomsection \label{thm-GSC2}
Let $N_{ij}$ denote the number of samples of alternative $i$ taken at design point $\bx_j$, and $\widehat{Y}_{ij}$ denote their means, for $i=1,\ldots,k$, $j=1,\ldots,m$.
Let $\widehat{\bY}_i = (\widehat{Y}_{i1}, \ldots, \widehat{Y}_{im})^\intercal$ and $\widehat{\BFbeta}_i = (\cX^\intercal \cX)^{-1} \cX^\intercal \widehat{\bY}_i$ for $i=1,\ldots,k$.
Under Assumption \ref{ass-model} or \ref{ass-model2}, the GSC defined in \eqref{eq:GSC_equality} is the LFC for a selection procedure of the R\&S-C problem with the IZ formulation and a fixed design,
if all the following properties hold:
\begin{enumerate}[label=(\roman*)]
\item
The selected alternative is   $\widehat{i^*}(\bx) = \argmax\limits_{1\leq i\leq k} \left\{  \bx^\intercal \widehat{\BFbeta}_i \right\} $.
\item
Conditionally on $\{N_{ij}:1\leq i\leq k, 1\leq j\leq m\}$, $\widehat{Y}_{ij} \sim \cN \left(\bx_j^\intercal \BFbeta_i, \sigma_i^2(\bx_j)/N_{ij} \right)$ for all $i=1,\ldots,k$, $j=1,\ldots,m$, and $\widehat{Y}_{ij}$ is independent of $\widehat{Y}_{i'j'}$ if $(i,j) \neq (i',j')$.
\item
$N_{ij}$ is independent of the  configuration of the means, for all $i=1,\ldots,k$, $j=1,\ldots,m$.
\end{enumerate}

\end{theorem}

\proof{Proof.}

Suppose that $\BFbeta=(\BFbeta_i:1\leq i \leq k)$ follows the GSC.
Then, $i^*(\bx) \equiv 1$ and by Property (i), conditionally on $\bX=\bx$,
\begin{align}
\PCS(\bx;\BFbeta) &=  \pr \left(\bx^\intercal \widehat{\BFbeta}_1 - \bx^\intercal \widehat{\BFbeta}_i  > 0,\; \forall i= 2,\ldots,k \right) \nonumber\\
&= \E \left[ \pr \left(\bx^\intercal \widehat{\BFbeta}_1 - \bx^\intercal \widehat{\BFbeta}_i  > 0,\;\forall i= 2,\ldots,k\Big|
 N_{ij}, 1\leq i\leq k, 1\leq j \leq m \right\} \right], \label{eq:PCS_GSC}
\end{align}
where the expectation is taken with respect to the $N_{ij}$'s and we write $\PCS(\bx;\BFbeta)$ to stress its dependence on $\BFbeta$ since we will consider a different configuration of the means later.

By Property (ii), conditionally on $\bX=\bx$ and $\{N_{ij}: 1 \leq i \leq k, 1\leq j\leq m\}$,  $\bx^\intercal \widehat{\BFbeta}_i$ is independent of $\bx^\intercal \widehat{\BFbeta}_{i'}$ for $i\neq i'$;
moreover,
\[\bx^\intercal \widehat{\BFbeta}_i \big| \{N_{i,j}: 1 \leq i \leq k, 1\leq j\leq m\} \; \sim \; \cN\left(\bx^\intercal \BFbeta_i, \tilde\sigma^2(\bx,\Sigma_i) \right),\]
where $\tilde\sigma^2(\bx,\Sigma_i)\coloneqq \bx^\intercal (\cX^\intercal \cX)^{-1} \cX^\intercal \Sigma_i \cX (\cX^\intercal \cX)^{-1} \bx$ and $\Sigma_i \coloneqq \mathrm{Diag}(\sigma_i^2(\bx_1)/N_{i1}, \ldots, \sigma_i^2(\bx_m)/N_{im})$.
In particular, $\tilde\sigma^2(\bx,\Sigma_i)$ does not depend on $\BFbeta$ by Property (iii).
Hence, if we let $\phi(\cdot; \mu,\sigma^2)$ denote the pdf of $\cN(\mu,\sigma^2)$, it follows from \eqref{eq:PCS_GSC} that
\begin{align} \label{eq-case-GSC}
\PCS(\bx;\BFbeta) = \E \left[ \int_{-\infty}^{+\infty} \prod_{i=2}^k \Phi \left( \frac{t - \bx^\intercal \BFbeta_i}{\tilde\sigma(\bx,\Sigma_i)} \right)
\phi (t; \bx^\intercal \BFbeta_1, \tilde\sigma^2(\bx,\Sigma_1))\,\ud t  \right].
\end{align}

We now consider a different configuration of the means, $\BFbeta^\dagger=(\BFbeta^\dagger_i:1\leq i\leq k)$.
We will show below that $\PCS(\bx;\BFbeta^\dagger) \geq \PCS(\bx;\BFbeta) $ for all $\bx\in\Theta$.
For each $i=1,\ldots,k$, we define sets $\Theta_i^{(1)}$ and $\Theta_i^{(2)}$ as follows,
\begin{align*}
&\Theta_i^{(1)} = \{\bx \in \Theta : \bx^\intercal \BFbeta_i^\dag - \bx^\intercal \BFbeta_j^\dag \geq \delta \mbox{ for all $j\neq i$} \}, \\
&\Theta_i^{(2)} = \{\bx \in \Theta : \bx^\intercal \BFbeta_i^\dag - \bx^\intercal \BFbeta_j^\dag \geq 0 \mbox{ for all $j\neq i$, and } \bx^\intercal \BFbeta_i^\dag - \bx^\intercal \BFbeta_j^\dag < \delta \mbox{ for some $j\neq i$} \}.
\end{align*}
Clearly,  $\{\Theta_i^{(1)}, \Theta_i^{(2)}:i=1,\ldots,k\}$ are mutually exclusive and $\Theta = \bigcup_{i=1}^k \left(\Theta_i^{(1)} \bigcup \Theta_i^{(2)}\right)$.
We next conduct our analysis for each $\Theta_i^{(1)}$ and $\Theta_i^{(2)}$, respectively.

\begin{itemize}
\item
\textbf{Case 1: $\Theta_1^{(1)} \neq \emptyset$.}
For any $\bx \in \Theta_1^{(1)}$, $\bx^\intercal \BFbeta_1^\dagger - \bx^\intercal \BFbeta_i^\dagger \geq \delta$ for each $i=2,\ldots,k$.
By the same analysis that leads to \eqref{eq-case-GSC}, we can show that for any $\bx \in \Theta_1^{(1)}$,
\begin{align*}
&\PCS(\bx;\BFbeta^\dagger)\\
={}& \E \left[ \int_{-\infty}^{+\infty} \prod_{i=2}^k \Phi \left( \frac{t - \bx^\intercal \BFbeta_i^\dagger}{\tilde\sigma(\bx,\Sigma_i)} \right)\phi (t; \bx^\intercal \BFbeta_1^\dagger, \tilde\sigma^2(\bx,\Sigma_1))\,\ud t \right]\\
={}&\E \left[ \int_{-\infty}^{+\infty} \prod_{i=2}^k \Phi \left( \frac{t +(\bx^\intercal \BFbeta_1^\dagger - \bx^\intercal \BFbeta_1) - \bx^\intercal \BFbeta_i^\dagger}{\tilde\sigma(\bx,\Sigma_i)} \right)\phi (t+(\bx^\intercal \BFbeta_1^\dagger - \bx^\intercal \BFbeta_1); \bx^\intercal \BFbeta_1^\dagger, \tilde\sigma^2(\bx,\Sigma_1))\,\ud t \right]\\
={}&\E \left[ \int_{-\infty}^{+\infty} \prod_{i=2}^k \Phi \left( \frac{t -(\bx^\intercal \BFbeta_1 - \bx^\intercal \BFbeta_1^\dagger + \bx^\intercal \BFbeta_i^\dagger) }{\tilde\sigma(\bx,\Sigma_i)} \right)\phi (t; \bx^\intercal \BFbeta_1, \tilde\sigma^2(\bx,\Sigma_1))\,\ud t \right].
\end{align*}
Due to \eqref{eq:GSC_equality} and the fact that $\bx^\intercal \BFbeta_1^\dagger - \bx^\intercal \BFbeta_i^\dagger \geq \delta$ for each $i=2,\ldots,k$,
$(\bx^\intercal \BFbeta_1 - \bx^\intercal \BFbeta_1^\dagger + \bx^\intercal \BFbeta_i^\dagger) \leq \bx^\intercal \BFbeta_i$, for $i=2,\ldots,k$.
Since $\Phi(\cdot)$ is an increasing function, it is straightforward to see that $\PCS(\bx;\BFbeta^\dagger) \geq \PCS(\bx;\BFbeta)$ for any $\bx\in\Theta_1^{(1)}$.

\item
\textbf{Case 2: $\Theta_1^{(2)} \neq \emptyset$.}
Fix an arbitrary $\bx \in \Theta_1^{(2)}$, let $\Omega(\bx) \coloneqq \{i=2,\ldots,k : \bx^\intercal \BFbeta_1^\dagger - \bx^\intercal \BFbeta_i^\dag  \geq \delta\}$.
Then, $\Omega(\bx) \subset \{2,\ldots,k\}$ by the definition of $\Theta_1^{(2)}$. If $\Omega(\bx) = \emptyset$, then each alternative $i$, $i=2,\ldots,k$, is in the IZ, and thus $\PCS(x,\BFbeta^\dagger) = 1$.
Otherwise,  $(\bx^\intercal \BFbeta_1 - \bx^\intercal \BFbeta_1^\dagger + \bx^\intercal \BFbeta_i^\dagger) \leq \bx^\intercal \BFbeta_i$ for each $i\in\Omega(\bx)$.
Hence,
\begin{align*}
\PCS(\bx; \BFbeta^\dagger) &\geq {\pr} \left(\bx^\intercal \widehat{\BFbeta}_1^\dagger - \bx^\intercal \widehat{\BFbeta}_i^\dag  > 0, \; \forall i\in  \Omega(\bx)  \right)\\
&= \E \left[ \int_{-\infty}^{+\infty} \prod_{i \in \Omega(\bx)} \Phi \left( \frac{t - \bx^\intercal \BFbeta_i^\dag}{\tilde\sigma(\bx,\Sigma_i)} \right)\phi (t; \bx^\intercal \BFbeta_1^\dagger, \tilde\sigma^2(\bx,\Sigma_1))\,\ud t   \right]\\
&= \E \left[ \int_{-\infty}^{+\infty} \prod_{i \in \Omega(\bx)} \Phi \left( \frac{t - (\bx^\intercal \BFbeta_1 - \bx^\intercal \BFbeta_1^\dagger + \bx^\intercal \BFbeta_i^\dagger)}{\tilde\sigma(\bx,\Sigma_i)} \right)\phi (t; \bx^\intercal \BFbeta_1, \tilde\sigma^2(\bx,\Sigma_1))\,\ud t   \right]\\
&\geq \E \left[ \int_{-\infty}^{+\infty} \prod_{i \in \Omega(\bx)} \Phi \left( \frac{t - \bx^\intercal \BFbeta_i}{\tilde\sigma(\bx,\Sigma_i)} \right) \phi (t; \bx^\intercal \BFbeta_1, \tilde\sigma^2(\bx,\Sigma_1))\,\ud t  \right]\\
&\geq \PCS(\bx;\BFbeta),
\end{align*}
where the last inequality holds because $0\leq \Phi(\cdot) \leq 1$ and $|\Omega(\bx)| < k-1 $.

\item
\textbf{Other Cases.}
For each $i=2,\ldots,k$, if  $\Theta_i^{(1)} \neq \emptyset$, then we can simply swap the indexes of alternative 1 and alternative $i$, and follow the same analysis as in Case 1.
Likewise, for each $i=2,\ldots,k$, if $\Theta_i^{(2)} \neq \emptyset$, we can follow the analysis  in Case 2.

\end{itemize}

Therefore, we conclude that $\PCS(\bx;\BFbeta^\dagger) \geq \PCS(\bx;\BFbeta)$ for any $\bx\in \Theta $.
So $\E \left[ \PCS(\bX;\BFbeta^\dagger) \right] \geq \E \left[ \PCS(\bX;\BFbeta) \right]$.
Moreover, the foregoing analysis also shows that, the equality may hold only if random vector $\bX$ is degenerate to a constant vector.
Thus, the GSC is the LFC.
\Halmos
\endproof

\begin{remark}
Obviously, both Procedure TS and TS$^+$ possess those properties specified in Theorem \ref{thm-GSC2}, so Theorem \ref{thm-GSC} in \S \ref{sec-LFC} holds immediately as a corollary of Theorem \ref{thm-GSC2}.
\end{remark}

\pdfbookmark[2]{Proof of Theorem \ref{thm-design}}{link-e7}
\hypertarget{EC.9}{
\section*{EC.9. \hspace{5pt} Proof of Theorem \ref{thm-design}}
}

\proof{Proof of Theorem \ref{thm-design}.}
It suffices to show that the extreme design yields the minimal value of the solution $h$ to \eqref{eq-geth} among all symmetric designs.
We first notice that by \eqref{eq-geth}, the design matrix $\cX$ takes effect on the total sample size of Procedure TS only through the form $\cX^\intercal \cX$.
In the sequel, two design matrices $\cX$ and $\tilde\cX$ are said to be \emph{equivalent} if $\cX^\intercal \cX = \tilde\cX^\intercal \tilde\cX$.
For instance, swapping any two rows of $\cX$ leads to an equivalent design matrix since it does not change $\cX^\intercal \cX$.

Let $\cX_*$ denote the design matrix corresponding to the extreme design $\cS^1=\cdots =\cS^b = \cS^0$, and $\cX_\dag$ denote a nonequivalent design matrix corresponding to a symmetric design.
The key of the proof is to show that
\begin{equation}\label{eq-goal}
\bx^\intercal (\cX_*^\intercal \cX_*)^{-1} \bx \leq \bx^\intercal (\cX_\dag^\intercal \cX_\dag)^{-1} \bx, \quad \bx \in \Theta,
\end{equation}
where the equality holds if and only if $\bx = \left(1, \frac{l_2+u_2}{2},\ldots,\frac{l_d+u_d}{2} \right)^\intercal$, which is the center of $\Theta$.

To see this, let $h_*$ and $h_\dag$ denote the solution $h$ of \eqref{eq-geth} for $\cX_*$ and $\cX_\dag$, respectively.
Notice that the double integral on the left-hand side of \eqref{eq-geth} is strictly increasing in $h$ whereas strictly decreasing in $\bx^\intercal (\cX^\intercal \cX)^{-1} \bx$.
Hence, if \eqref{eq-goal} holds, then $h_* \leq h_\dag$, where the equality holds if and only if the random vector $\bX\equiv \left(1, \frac{l_2+u_2}{2},\ldots,\frac{l_d+u_d}{2} \right)^\intercal$.

Now we prove \eqref{eq-goal}.
For ease of presentation, we first consider a design matrix that corresponds to a general symmetric design.
Since the first element of the covariates is always 1, the $b(2^{d-1}) \times d$ design matrix $\cX$ is
\begin{equation*}
\cX =
\left(
  \begin{array}{c}
    (\ba_1^1)^\intercal \\
    \vdots \\
    (\ba_{2^{d-1}}^1)^\intercal \\
    \vdots \\
    (\ba_1^b)^\intercal \\
    \vdots \\
    (\ba_{2^{d-1}}^b)^\intercal \\
  \end{array}
\right)
=
\left(
  \begin{array}{cccc}
    1 &a_{1,2}^1 & \cdots & a_{1,d}^1 \\
    \vdots & \vdots & \vdots & \vdots \\
    1 & a_{2^{d-1},2}^1 & \cdots & a_{2^{d-1},d}^1 \\
    \vdots & \vdots & \vdots & \vdots \\
    1 & a_{1,2}^b & \cdots & a_{1,d}^b \\
    \vdots & \vdots & \vdots & \vdots \\
    1 & a_{2^{d-1},2}^b & \cdots & a_{2^{d-1},d}^b \\
  \end{array}
\right)
\triangleq
\left(
  \begin{array}{cccc}
    \bone & \bv_2 & \cdots & \bv_d \\
  \end{array}
\right),
\end{equation*}
where $\bone$ denotes the $b(2^{d-1}) \times 1$ vector of ones.
We further set  $\cZ \coloneqq (\bv_2, \ldots, \bv_d)$.
Then $\cX = (\bone, \cZ)$, and
\begin{equation}\label{eq-matrix}
\cX^\intercal \cX =
\left(
  \begin{array}{cc}
    \bone^\intercal \bone & \bone^\intercal \cZ \\
    \cZ^\intercal \bone & \cZ^\intercal \cZ \\
  \end{array}
\right).
\end{equation}
Notice that $m = b(2^{d-1}) = \bone^\intercal \bone$.
Then, for any $\bx = (1, \bz^\intercal)^\intercal$, where $\bz \in \mathbb{R}^{d-1}$, standard matrix calculation \cite[\S 0.8.5]{horn1990matrix_ec} yields that
$$
\bx^\intercal (\cX^\intercal \cX)^{-1} \bx = \left( \bz - \cZ^\intercal \bone m^{-1} \right)^\intercal \cA(\cZ)^{-1} \left( \bz - \cZ^\intercal \bone m^{-1} \right) + m^{-1},
$$
where $\cA(\cZ) \coloneqq \cZ^\intercal \cZ - \cZ^\intercal \bone m^{-1} \bone^\intercal \cZ $ is the Schur complement of the block $\bone^\intercal \bone$ of $\cX^\intercal \cX$ in \eqref{eq-matrix} and it is nonsingular because $\cX^\intercal \cX$ is nonsingular.
The symmetry of design points implies that
$
\bv_w^\intercal \bone m^{-1} = \frac{l_w+u_w}{2},
$
for $w=2,\ldots,d$. So, by letting $\bs \coloneqq \left(\frac{l_2+u_2}{2},\ldots,\frac{l_d+u_d}{2} \right)^\intercal$, we have $\cZ^\intercal \bone m^{-1} = \bs$ and
\begin{equation}\label{eq-matrix-simpl}
\bx^\intercal (\cX^\intercal \cX)^{-1} \bx = \left( \bz - \bs \right)^\intercal \cA(\cZ)^{-1} \left( \bz - \bs \right) + m^{-1}.
\end{equation}
Hence, if $\bx = \left(1, \frac{l_2+u_2}{2},\ldots,\frac{l_d+u_d}{2} \right)^\intercal$, then $\bz=\bs$ and thus $\bx^\intercal (\cX^\intercal \cX)^{-1} \bx = m^{-1}$. Since both $\cX_*$ and $\cX_\dag$ are symmetric designs, $\bx^\intercal (\cX_*^\intercal \cX_*)^{-1} \bx = \bx^\intercal (\cX_\dag^\intercal \cX_\dag)^{-1} \bx$ if $\bx = \left(1, \frac{l_2+u_2}{2},\ldots,\frac{l_d+u_d}{2} \right)^\intercal$.

It remains to prove the strict inequality in \eqref{eq-goal} for $\bz \neq \bs$.
Let $\cX_* = (\bone, \cZ_*)$ and $\cX_\dag = (\bone, \cZ_\dag)$.
Due to \eqref{eq-matrix-simpl}, it suffices to show that $\cA(\cZ_{\dag})^{-1} - \cA(\cZ_{*})^{-1}$ is positive definite.
This is equivalent to showing that $\cA(\cZ_{*}) - \cA(\cZ_{\dag})$ is positive definite \cite[Corollary 7.7.4]{horn1990matrix_ec}, i.e., for any nonzero $\bz \in \mathbb{R}^{d-1}$,
\begin{equation}\label{eq-goal-sub}
\bz^\intercal \cA(\cZ_{*})\bz > \bz^\intercal \cA(\cZ_{\dag})\bz.
\end{equation}

Let $\cI$ denote the $b(2^{d-1}) \times b(2^{d-1})$ identity matrix. Then,
\[
\cA(\cZ) = \cZ^\intercal \left(\cI - \bone m^{-1} \bone^\intercal \right) \cZ
= \cZ^\intercal \left(\cI - \bone m^{-1} \bone^\intercal \right) \left(\cI - \bone m^{-1} \bone^\intercal \right) \cZ
= \left(\cZ - \bone \bs^\intercal\right)^\intercal \left(\cZ - \bone \bs^\intercal\right),
\]
since $\cZ^\intercal \bone m^{-1} = \bs$.
Denote $\bz \coloneqq (z_2,\ldots,z_d)^\intercal$ and $\bs \coloneqq (s_2,\ldots,s_d)^\intercal$.
Then
\[
\bz^\intercal \cA(\cZ)\bz = \left[ \left(\cZ - \bone \bs^\intercal \right) \bz \right]^\intercal \left[ \left(\cZ - \bone \bs^\intercal \right) \bz \right]
= \sum_{w=2}^d \sum_{q=2}^d z_w z_q (\bv_w-\bone s_w)^\intercal (\bv_q-\bone s_q).
\]
Thanks to the symmetry of the design points, it is easy to verify that
\begin{align*}
(\bv_w-\bone s_w)^\intercal (\bv_q-\bone s_q) =
\begin{cases}
0, & \text{if $w\neq q$},\\[0.4em]
2^{d-1} \sum_{j=1}^b (\rho_w^j)^2, & \text{if $w = q$},\\
\end{cases}
\end{align*}
where $\rho_w^j \coloneqq \big| a_{1,w}^j - s_w \big| = \cdots = \big| a_{2^{d-1},w}^j - s_w \big|$ is the common distance of $\ba_1^j,\ldots,\ba_{2^{d-1}}^j$ to the center of $\Theta$ along coordinate $x_w$,
for $w=2,\ldots,d$ and $j=1,\ldots,b$.
Hence,
\begin{align*}
\bz^\intercal \cA(\cZ)\bz = 2^{d-1} \sum_{w=2}^d z_w^2 \sum_{j=1}^b (\rho_w^j)^2 .
\end{align*}
Since $\rho_w^j \in \left(0,\frac{u_w-l_w}{2} \right]$,
obviously, $\{\rho_w^j = \frac{u_w-l_w}{2}|w=2,\ldots,d,\, j=1,\ldots,b\}$ maximizes $\bz^\intercal \cA(\cZ)\bz$.
Because $\bz$ is nonzero, i.e., at least one $z_w$ is not zero, the solution is unique in terms of $\rho_w^j$.
Notice that this solution exactly means that $\cS^1=\cdots =\cS^b = \cS^0$, i.e., the extreme design.
Hence, \eqref{eq-goal-sub} is proved and the proof of \eqref{eq-goal} is completed.
\Halmos
\endproof

\pdfbookmark[2]{Proof of Theorem \ref{thm-D-G-opt}}{link-e8}
\hypertarget{EC.10}{
\section*{EC.10. \hspace{5pt} Proof of Theorem \ref{thm-D-G-opt}}
}

Before the proof, we first introduce formally the D-optimality and the G-optimality of experimental designs in the linear regression setting.
Consider the linear regression model
\begin{equation*}
Y(\bx) = \bx^\intercal \BFbeta + \epsilon,
\end{equation*}
where $\BFbeta, \bx \in \mathbb{R}^{d}$ and $\epsilon$ is random error with mean 0 and variance $\sigma^2$.
Assuming the design region is $\Theta$, we choose $m$ design points with $\bx_i \in \Theta$, $i=1,\ldots,m$.
Let $\bY = (Y(\bx_1), \ldots, Y(\bx_m))^\intercal$, $\cX = (\bx_1, \ldots, \bx_m)^\intercal$, and $\Upsilon =  \left\{\cX : \textnormal{rank}\left(\cX^\intercal\cX\right) = d, \bx_i \in \Theta, i=1,\ldots, m\right\}$.
It is known that  if $\cX \in \Upsilon$, the OLS estimator of $\BFbeta$ is $\widehat{\BFbeta} = (\cX^\intercal \cX)^{-1} \cX^\intercal \bY$.
Moreover, $\Var(\widehat{\BFbeta}) = \sigma^2 (\cX^\intercal \cX)^{-1}$ and $\Var(\bx^\intercal \widehat{\BFbeta}) = \sigma^2 \bx^\intercal (\cX^\intercal \cX)^{-1} \bx$. The D-optimality and the G-optimality are related to the two variances, respectively.

A design $\cX_*$ is said to be \emph{D-optimal} if
\begin{equation*}
\cX_* = \argmax_{\cX \in \Upsilon} \det \left( \cX^\intercal \cX \right).
\end{equation*}
The D-optimal design aims to minimize the volume of confidence ellipsoid for $\BFbeta$ given a fixed confidence level under the assumption that the errors are normally distributed.

A design $\cX_*$ is said to be \emph{G-optimal} if
\begin{equation*}
\cX_* = \argmin_{\cX \in \Upsilon} \left\{ \max_{\bx \in \Theta} \bx^\intercal (\cX^\intercal \cX)^{-1} \bx \right\}.
\end{equation*}
The G-optimal design aims to minimize the maximum variance of the fitted response over the design region.

Theorem \ref{thm-D-G-opt} is an application of the general equivalence theory; see, e.g., \citet[Chapter 3]{silvey1980optimal_ec} for a careful discussion on this subject. Some concepts need to be introduced before the proof.

The first concept is the \emph{continuous design}, also called approximate design.
Suppose that we relax the constraint that the number of samples at each design point must be an integer, that is, we can allocate any portion of a given total sample size $m$  to any point in $\Theta$.
Formally speaking, the allocation can be described by a probability distribution $\psi$ on $\Theta$, which can be either continuous or discrete.
Let $\bX$ be a random vector with distribution $\psi$, and define $\cM(\psi) \coloneqq \E_\psi(\bX \bX^\intercal)$.
For example, if $\cX$ is an \emph{exact} design which contains \emph{distinct} points $\bx_1,\ldots,\bx_n$, having $m_1,\ldots,m_n$ samples, respectively, where $m_1 + \cdots + m_n=m$, then the distribution $\psi$ for sample allocation is defined by $\pr(\bX=\bx_i)= m_i / m$, $i=1,\ldots,n$, and  thus $\cM(\psi) = m^{-1} \cX^\intercal \cX$. However, a continuous design may not be an exact design due to the integrality constraint.

By allowing continuous designs, the D-optimal design is extended to be a distribution
\[\psi_* = \argmax_{\psi \in \Psi} \det \left( \cM (\psi) \right),\]
where $\Psi$ denotes the set of all $\psi$ such that $\cM (\psi)$ is nonsingular.
Notice that $\Psi$ is also the set of all $\psi$ such that $\cM (\psi)$ is positive definite.
Likewise, the G-optimal design can be extended as follows
\[\psi_* = \argmin_{\psi \in \Psi} \left\{ \max_{\bx \in \Theta} \bx^\intercal (\cM(\psi))^{-1} \bx \right\}.\]

More generally, consider a function $f$ of positive definite matrices.
A continuous design $\psi_*$ is said to be $f$-optimal if
\[\psi_* = \argmax_{\psi \in \Psi} f\left( \cM (\psi) \right).\]
For instance, the D-optimal design and the G-optimal design can be obtained by setting $f(\cM) = \log \det \left( \cM \right)$ and $f(\cM ) = - \max\limits_{\bx \in \Theta} \bx^\intercal (\cM )^{-1} \bx $, respectively.
It is easy to verify that both functions are concave in $\cM$.
The use of concavity will become clear in Lemma \ref{lem-optimal} below.

At last, we introduce two kinds of derivatives. The \emph{G\^{a}teaux derivative} of $f$ at $\cM_1$ in the direction of $\cM_2$ is defined as
$$G_f(\cM_1,\cM_2) \coloneqq \lim_{\varepsilon \to  0^+} \frac{1}{\varepsilon} \left[ f(\cM_1 + \varepsilon \cM_2) - f(\cM_1) \right].$$
We say $f$ is \emph{differentiable} at $\cM_1$ if $G_f(\cM_1,\cM_2)$ is well defined.
The \emph{Fr\'{e}chet derivative} of $f$ at $\cM_1$ in the direction of $\cM_2$ is defined as
$$F_f(\cM_1,\cM_2) \coloneqq \lim_{\varepsilon \to  0^+} \frac{1}{\varepsilon} \left[ f\left\{(1-\varepsilon) \cM_1 + \varepsilon \cM_2 \right\} - f(\cM_1) \right]
= G_f(\cM_1,\cM_2-\cM_1).$$

We state Theorem 3.7 of \cite{silvey1980optimal_ec} as Lemma \ref{lem-optimal} and will apply it to prove Theorem \ref{thm-D-G-opt}.

\begin{lemma} \phantomsection \label{lem-optimal}
If $f$ is a concave function of positive definite matrices and is differentiable at $\cM(\psi_*)$, then $\psi_*$ if $f$-optimal if and only if $F_f(\cM(\psi_*),\bx \bx^\intercal)\leq 0$ for all $\bx \in \Theta$.
\end{lemma}

Now we are ready to prove Theorem \ref{thm-D-G-opt}.

\proof{Proof of Theorem \ref{thm-D-G-opt}.}
Consider the continuous design $\psi_0$ that assigns probability $1/(2^{d-1})$ at each corner point of $\Theta$.
Since $m=b(2^{d-1})$, $\psi_0$ is indeed the exact extreme design defined in \S \ref{subsec-optimal}.
Hence, it suffices to prove that $\psi_0$ is both D-optimal and G-optimal in the continuous case.

We first prove the D-optimality of $\psi_0$. Let $\cX_0$ denote the design matrix corresponding to $\psi_0$. Then, $\cX_0$ is a $2^{d-1} \times d$ matrix with each row corresponding one of the $2^{d-1}$ distinct corners of $\Theta$. For example, if $d=3$,
$$
\cX_0 =
\left(
  \begin{array}{ccc}
    1 & l_2 & l_3 \\
    1 & l_2 & u_3 \\
    1 & u_2 & l_3 \\
    1 & u_2 & u_3 \\
  \end{array}
\right).
$$
It is easy to see that $\cM(\psi_0) = \frac{1}{2^{d-1}} \cX_0^\intercal \cX_0$. In the sequel, we verify the conditions of Lemma \ref{lem-optimal} for $f = \log \det $ to prove the D-optimality.

The concavity of $f$ is trivial; see, e.g., Theorem 7.6.6 of \cite{horn1990matrix_ec}. For the differentiability, notice that for any positive definite matrix $\cM_1$,
$$
\log \det(\cM_1 + \varepsilon \cM_2) - \log \det(\cM_1) = \log \det (\cI + \varepsilon \cM_1^{-1} \cM_2) = \sum_{w=1}^d \log (1+\varepsilon \lambda_w),
$$
where $\cI$ is the $d \times d$ identity matrix and $\lambda_1, \ldots, \lambda_d$ are the eigenvalues of $\cM_1^{-1} \cM_2$ which are all real \cite[Corollary 7.6.2]{horn1990matrix_ec}. Hence,
\[
G_f(\cM_1,\cM_2) = \lim_{\varepsilon \to  0^+} \frac{1}{\varepsilon} \left[  \log \det(\cM_1 + \varepsilon \cM_2) - \log \det(\cM_1) \right]
= \sum_{w=1}^d \lambda_w
= \tr(\cM_1^{-1} \cM_2),
\]
is well defined, so $f$ is differentiable at $\cM_1$.

Moreover,
$$F_f(\cM_1,\cM_2) = G_f(\cM_1,\cM_2-\cM_1) = \tr(\cM_1^{-1} \cM_2 - \cI) = \tr(\cM_1^{-1} \cM_2) - d.$$
Hence, for any $\bx \in \Theta$,
$$F_f(\cM(\psi_0),\bx \bx^\intercal) = \tr(\cM(\psi_0)^{-1} \bx \bx^\intercal) - d = \tr(\bx^\intercal \cM(\psi_0)^{-1} \bx) - d =  \bx^\intercal \cM(\psi_0)^{-1} \bx - d.$$
Since $\cM(\psi_0)^{-1}$ is positive definite, $F_f(\cM(\psi_0),\bx \bx^\intercal)$ is convex in $\bx$, thereby achieving its maximum only if $\bx$ is one of the corners of $\Theta$, i.e., $\bx \in \cS^0$.
Therefore, to verify $F_f(\cM(\psi_0),\bx \bx^\intercal)\leq 0$, it suffices to show  that
\begin{equation}\label{eq-D-goal}
\bx^\intercal \cM(\psi_0)^{-1} \bx = d,\quad \bx \in \cS^0.
\end{equation}

We denote $\cX_0 \coloneqq (\bone, \cZ_0) \triangleq (\bone, \bv_2, \ldots, \bv_d)$, where $\bone$ is the $2^{d-1} \times 1$ vector of ones.
Following the standard matrix calculations similar to those in the proof of Theorem \ref{thm-design}, we can have that, for any $\bx = (1, \bz^\intercal)^\intercal$, where $\bz \coloneqq \left(z_2,\ldots,z_d \right)^\intercal \in \mathbb{R}^{d-1}$,
\begin{equation}\label{eq-D-goal-a}
\bx^\intercal \cM(\psi_0)^{-1} \bx = 2^{d-1} \bx^\intercal (\cX_0^\intercal \cX_0)^{-1} \bx
 = 2^{d-1} \left[ \left( \bz - \bs \right)^\intercal \left\{ \left(\cZ_0 - \bone \bs^\intercal\right)^\intercal \left(\cZ_0 - \bone \bs^\intercal\right) \right\}^{-1} \left( \bz - \bs \right) + 1/\left(2^{d-1} \right) \right],
\end{equation}
where $\bs = \left(s_2,\ldots,s_d \right)^\intercal \coloneqq \left(\frac{l_2+u_2}{2},\ldots,\frac{l_d+u_d}{2} \right)^\intercal $.
Notice that $ \left(\cZ_0 - \bone \bs^\intercal\right)^\intercal \left(\cZ_0 - \bone \bs^\intercal\right) $ is a $(d-1) \times (d-1)$ matrix
whose $(w-1,q-1)$-th element is $(\bv_w-\bone s_w)^\intercal (\bv_q-\bone s_q)$, for $w,q=2,\ldots,d$, and that
\begin{align*}
(\bv_w-\bone s_w)^\intercal (\bv_q-\bone s_q) =
\begin{cases}
0, & \text{if $w\neq q$},\\[0.4em]
2^{d-1} \left(\frac{u_w-l_w}{2}\right)^2, & \text{if $w = q$}.\\
\end{cases}
\end{align*}
Hence,
$$ \left(\cZ_0 - \bone \bs^\intercal\right)^\intercal \left(\cZ_0 - \bone \bs^\intercal\right) = 2^{d-1} \mathrm{Diag}\left\{ \left(\frac{u_2-l_2}{2}\right)^2, \ldots, \left(\frac{u_d-l_d}{2}\right)^2 \right\},$$
and thus
$$ \left\{ \left(\cZ_0 - \bone \bs^\intercal\right)^\intercal \left(\cZ_0 - \bone \bs^\intercal\right) \right\}^{-1} = \frac{1}{2^{d-1}} \mathrm{Diag}\left\{ \left(\frac{2}{u_2-l_2}\right)^2, \ldots, \left(\frac{2}{u_d-l_d}\right)^2 \right\}.$$
Moreover, for any $\bx \in \cS^0$, $\bz\in \{l_2,u_2\} \times \cdots \times \{l_d,u_d\}$.
Hence, $(z_w-s_w)^2 = \left(\frac{u_w-l_w}{2}\right)^2$, for $w=2,\ldots,d$.
Then,
\begin{align} \label{eq-D-goal-b}
\left( \bz - \bs \right)^\intercal \left\{ \left(\cZ_0 - \bone \bs^\intercal\right)^\intercal \left(\cZ_0 - \bone \bs^\intercal\right) \right\}^{-1} \left( \bz - \bs \right)
= \frac{1}{2^{d-1}} \sum_{w=2}^d \left(\frac{2}{u_w-l_w}\right)^2 (z_w-s_w)^2 = \frac{d-1}{2^{d-1}}.
\end{align}
Then, \eqref{eq-D-goal}  follows immediately from \eqref{eq-D-goal-a} and \eqref{eq-D-goal-b}, proving the D-optimality by Lemma \ref{lem-optimal}.

The G-optimality of $\psi_0$ can be proved similarly, by taking $f(\cM(\psi)) = - \max\limits_{\bx \in \Theta} \bx^\intercal (\cM(\psi))^{-1} \bx $.
Or, we can conclude this immediately by applying the known equivalence between the D-optimality and the G-optimality for continuous designs established in \citet{kiefer1960equivalence_ec}; see also \citet[\S 3.11]{silvey1980optimal_ec}.
\Halmos
\endproof

\pdfbookmark[2]{References}{link-ec-ref}

\end{document}